%% file: main_ode.tex
\documentclass{article}
\usepackage[preprint]{iclr2027_conference}
\usepackage{times}

\input{header.tex}

\input{def}

\title{An ODE-based Sampler for Discrete Diffusion Models}

\author{%
\parbox{\textwidth}{\centering
    \vspace*{1ex}
    \textbf{Samson Gourevitch}$^{*,1}$ \qquad
    \textbf{Yazid Janati}$^{*,2,3}$\\[0.3ex]
    \textbf{Eric Moulines}$^{3,4}$ \qquad
    \textbf{Eric P. Xing}$^{2,3}$ \qquad
    \textbf{Alain Durmus}$^{\dagger,1}$ \qquad
    \textbf{Giacomo Zanella}$^{\dagger,5}$\\[0.9ex]
    {\normalfont\small
        $^1$ CMAP, Ecole polytechnique \quad
        $^2$ Institute of Foundation Models \quad
        $^3$ MBZUAI\\
        $^4$ EPITA, LRE \quad
        $^5$ Bocconi University, Department of Decision Sciences and BIDSA
    }
}
}

\begin{document}
\setlength{\parskip}{4pt}
\setlength{\abovedisplayskip}{4pt}
\setlength{\belowdisplayskip}{4pt}
\setlength{\abovedisplayshortskip}{2pt}
\setlength{\belowdisplayshortskip}{2pt}
\setlength{\textfloatsep}{10pt plus 2pt minus 2pt}
\setlength{\floatsep}{8pt plus 2pt minus 2pt}
\setlength{\intextsep}{8pt plus 2pt minus 2pt}
\captionsetup{skip=4pt}
\maketitle

\begin{abstract}
    The marginals of a continuous diffusion model can be reproduced by a deterministic ordinary differential equation (ODE), known as the probability flow ODE, which provides a foundation for accelerated sampling and consistency distillation. In this paper we address the question whether an analogous formulation exists for discrete diffusion models and answer it affirmatively by introducing exponential residual flows (ERFs). Our framework augments the discrete state with a continuous residual variable for each token and position. These residuals evolve deterministically according to piecewise-defined ODEs whose vector fields depend on the current discrete state and the forward and backward rate matrices of the prescribed diffusion model. The discrete state changes whenever a residual reaches zero. We prove that this construction preserves the marginals of the original diffusion. We derive a practical discretization of our sampler and specialize it to masked and uniform diffusion. The deterministic trajectories of ERFs support consistency distillation and yield empirical improvements over previous approaches across a variety of benchmarks including code and text generation tasks.
\end{abstract}

\blfootnote{* Equal contribution}
\blfootnote{$\dagger$ Equal senior contribution}
\blfootnote{Corresponding authors: \texttt{samson.gourevitch@polytechnique.edu},\\
\texttt{yazid.janati@mbzuai.ac.ae}}

\input{body/introduction}
\input{body/background}
\input{body/deterministic_sampler}
\input{body/distillation}
\input{body/related_work.tex}
\input{body/experiments}
\input{body/conclusion}

\setlength{\parskip}{.5pc}
\setlength{\abovedisplayskip}{6pt}
\setlength{\belowdisplayskip}{6pt}
\setlength{\abovedisplayshortskip}{4pt}
\setlength{\belowdisplayshortskip}{4pt}
\setlength{\textfloatsep}{20pt plus 2pt minus 4pt}
\setlength{\floatsep}{12pt plus 2pt minus 2pt}
\setlength{\intextsep}{12pt plus 2pt minus 2pt}
\captionsetup{skip=10pt}
\clearpage

\subsection*{AI use statement}
A generative AI assistant was used to support literature searches, discuss mathematical expressions and experimental results, and assist with manuscript drafting, editing, and \LaTeX{} preparation. The authors take responsibility for all claims, citations, derivations, reported results, and the final content of the paper.

\subsection*{Acknowledgments}
We thank Patrick Pynadath and Jawad Chemaou for their useful comments on the presentation of this paper.

AD received funding from the Fondation de l'École polytechnique through its Servir la science campaign. AD is supported by the France 2030 program under reference ANR-25-PEIA-0001 (THEOREM project). AD is funded by the European Union (ERC-2022-SYG-OCEAN-101071601). AD is supported by Hi! Paris and the Agence Nationale de la Recherche (Grant 11-LABX-0047). This work received government funding administered by the National Research Agency (ANR) under the France 2030 program \emph{Hi! PARIS}, grant number ANR-23-IACL-0005.

\newpage
\bibliographystyle{iclr2027_conference}
\bibliography{bibliography}

\appendix
\section*{Appendix contents}
\startcontents[appendix]
\printcontents[appendix]{}{1}[3]{}

\input{appendix/practical_discretization}
\input{appendix/bridge_simplex}
\input{appendix/ctmc_integrators}
\input{appendix/distillation}
\input{appendix/experimental_details}
\input{appendix/experiments}
\input{appendix/diffusiongemma}
\input{appendix/exponential_residual_flow_transcription}
\end{document}

%% file: header.tex
\usepackage{ifthen}
\usepackage{xargs}
\usepackage{amsmath}
\usepackage{bm}
\usepackage{amssymb}
\usepackage{dsfont}
\usepackage{mathtools}
\usepackage{amsthm}
\usepackage{booktabs}
\usepackage{colortbl}
\usepackage{comment}
\usepackage{stmaryrd}
\usepackage[colorlinks=true,linkcolor=blue,citecolor=blue,urlcolor=blue]{hyperref}
\usepackage{todonotes}
\usepackage{wrapfig}

\newcommand{\finishwrappedtable}{%
    \par
    \ifnum\value{WF@wrappedlines}>1
        \addtocounter{WF@wrappedlines}{-1}%
        \vspace*{\value{WF@wrappedlines}\baselineskip}%
    \fi
    \WFclear
}

\usepackage{caption}
\usepackage{subcaption}
\usepackage{float}
\usepackage{placeins}
\usepackage{upgreek}
\usepackage{graphicx}
\usepackage{algorithm}
\usepackage{algpseudocode}
\usepackage{titletoc}
\makeatletter
\renewcommand*{\theHALG@line}{\thealgorithm.\arabic{ALG@line}}
\newcommand{\fixedspaceparagraph}[1]{%
  \@startsection{paragraph}{4}{\z@}{1.5ex}{-1em}{\normalsize\bfseries}{#1}%
}
\makeatother
\usepackage{enumitem}
\definecolor{crimson}{RGB}{220,20,60}
\definecolor{erflineblue}{HTML}{F0F7FF}
\definecolor{tablegroup}{HTML}{E8E8FF}
\definecolor{tableours}{HTML}{FFDFC2}

\newcommand\blfootnote[1]{%
  \begingroup
  \renewcommand\thefootnote{}\footnote{#1}%
  \addtocounter{footnote}{-1}%
  \endgroup
}
\usepackage[framemethod=TikZ]{mdframed}
\mdfdefinestyle{propFrame}{%
    linecolor=black!8!white,
    outerlinewidth=.6pt,
    roundcorner=5pt,
    innertopmargin=7pt,
    innerbottommargin=7pt,
    innerrightmargin=8pt,
    innerleftmargin=8pt,
    backgroundcolor=black!3!white
}

\mdfsetup{
  skipabove=.2cm,
  skipbelow=0pt
}

\usepackage[capitalize,noabbrev]{cleveref}

\theoremstyle{plain}
\newtheorem{theorem}{Theorem}[section]
\newtheorem{proposition}{Proposition}
\newtheorem{lemma}{Lemma}
\newtheorem{corollary}{Corollary}
\theoremstyle{definition}
\newtheorem{definition}{Definition}
\newtheorem{assumption}{Assumption}
\theoremstyle{plain}

\crefname{appendix}{appendix}{appendices}
\Crefname{appendix}{Appendix}{Appendices}
\AddToHook{cmd/appendix/after}{%
  \crefalias{section}{appendix}%
  \crefalias{subsection}{subappendix}%
  \crefalias{subsubsection}{subsubappendix}%
}
\crefname{algorithm}{algorithm}{algorithms}
\Crefname{algorithm}{Algorithm}{Algorithms}
\crefname{assumption}{assumption}{assumptions}
\Crefname{assumption}{Assumption}{Assumptions}
\algrenewcommand\algorithmicrequire{\textbf{Require:}}
\algrenewcommand\algorithmicensure{\textbf{Ensure:}}
\algnewcommand\algorithmicoutput{\textbf{Output:}}
\algnewcommand\Output{\item[\algorithmicoutput]}

\newcounter{hypA}

%% file: def.tex
\newlist{renumerate}{enumerate}{1}
\setlist[renumerate]{label=\textup{(\roman*)},nosep,leftmargin=*}
\newlist{benumerate}{enumerate}{1}
\setlist[benumerate]{label=$\bullet$,nosep,leftmargin=*}
\def\vocab{K}
\def\loo{\mathrm{loo}}

\def\len{L}

\def\msv{\mathsf{V}}

\def\msx{\mathsf{X}}

\def\rset{\mathbb{R}}

\def\rmd{\mathrm{d}}

\def\normpdf{\mathrm{N}}
\def\eqsp{\,}

\def\one{\mathbf{1}}

\def\pE{\mathbb{E}}
\def\softmax{\mathrm{softmax}}
\def\pP{\mathbb{P}}

\def\indic{\mathds{1}}

\def\eqdef{\vcentcolon=}

\def\simiid{\overset{\mathrm{i.i.d.}}{\sim}}

\def\law{\text{Law}}

\def\last{n}
\newcommand{\tk}[1]{{t_{#1}}}
\newcommand{\sk}[1]{{s_{#1}}}

\newcommand{\sres}[3]{\ensuremath{#1_{\scriptscriptstyle\mathrm{#2},\,#3}}}
\DeclareMathOperator*{\argmax}{argmax}
\DeclareMathOperator*{\argmin}{argmin}

\def\mask{\mathsf{m}}

\newcommand{\dotp}[2]{\langle #1, #2 \rangle}

\def\txts{\textstyle}
\def\rhs{r.h.s.}

\def\ie{\textit{i.e.}}

\newcommand{\intset}[2]{\llbracket #1, #2 \rrbracket}

\newcommand{\kldivergence}[2]{\mathrm{KL}(#1\|#2)}

 \def\rset{{\mathbb{R}}}

 \def\rmd{\mathrm{d}}
 \def\eqsp{\;}
 \def\eqdef{\coloneqq}

\newcommandx{\Vnorm}[2][1=V]{\| #2 \|_{#1}}
\newcommandx{\VnormEq}[2][1=V]{\left\| #2 \right\|_{#1}}
\newcommandx{\norm}[2][1=]{\ifthenelse{\equal{#1}{}}{\left\Vert #2 \right\Vert}{\left\Vert #2 \right\Vert^{#1}}}
\newcommandx{\normLigne}[2][1=]{\ifthenelse{\equal{#1}{}}{\Vert #2 \Vert}{\Vert #2\Vert^{#1}}}

\newcommand{\defEns}[1]{\left\lbrace #1 \right\rbrace }

\newcommand{\ps}[2]{\left\langle#1,#2 \right\rangle}

\newcommand{\tbar}{\mathrel{\raisebox{0.15ex}{$\scriptscriptstyle\mid$}}}

\newcommandx\vartarg[4][4=]{
    \ifthenelse{\equal{#2}{}}{
        \ifthenelse{\equal{#3}{}}{
            \varpi^{#4} _{\!#1}
        }{
            \varpi^{#4} _{\!#1}(#3)
            }
    }{
        \ifthenelse{\equal{#3}{}}{
            \varpi^{#4} _{\!#1}(\cdot|#2)
        }{
            \varpi^{#4} _{\!#1}(#3|#2)
        }
    }}

\newcommandx\targ[4][4=]{
    \ifthenelse{\equal{#2}{}}{
        \ifthenelse{\equal{#3}{}}{
            \pi^{#4} _{\!#1}
        }{
            \pi^{#4} _{\!#1}(#3)
            }
    }{
        \ifthenelse{\equal{#3}{}}{
            \pi^{#4} _{\!#1}(\cdot|#2)
        }{
            \pi^{#4} _{\!#1}(#3|#2)
        }
    }}

\newcommandx\interp[4][4=\interpscale]{
    \ifthenelse{\equal{#2}{}}{
        \ifthenelse{\equal{#3}{}}{
            \pi^{#4} _{#1}
        }{
            \pi^{#4} _{#1}(#3)
            }
    }{
        \ifthenelse{\equal{#3}{}}{
            \pi^{#4} _{#1}(\cdot|#2)
        }{
            \pi^{#4} _{#1}(#3|#2)
        }
    }}

\newcommandx\dpdata[4][4=]{
\ifthenelse{\equal{#2}{}}{
    \ifthenelse{\equal{#3}{}}{
        p^{d, #4} _{#1}
    }{
        p^{d, #4} _{#1}(#3)
    }
}{
    \ifthenelse{\equal{#3}{}}{
        p^{d, #4} _{#1}(\cdot|#2)
    }{
        p^{d, #4} _{#1}(#3|#2)
    }
}}

\newcommandx\hpdata[4][4=]{
\ifthenelse{\equal{#2}{}}{
    \ifthenelse{\equal{#3}{}}{
        \hat{p}^{#4} _{#1}
    }{
        \hat{p}^{#4} _{#1}(#3)
    }
}{
    \ifthenelse{\equal{#3}{}}{
        \hat{p}^{#4} _{#1}(\cdot|#2)
    }{
        \hat{p}^{#4} _{#1}(#3|#2)
    }
}}
\newcommandx\pdata[4][4=]{
    \ifthenelse{\equal{#2}{}}{
        \ifthenelse{\equal{#3}{}}{
            p^{#4} _{#1}
        }{
            p^{#4} _{#1}(#3)
        }
    }{
        \ifthenelse{\equal{#3}{}}{
            p^{#4} _{#1}(\cdot|#2 )
        }{
            p^{#4} _{#1}(#3|#2)
        }
    }}

\newcommandx\barpdata[4][4=]{
    \ifthenelse{\equal{#2}{}}{
        \ifthenelse{\equal{#3}{}}{
            \bar{p}^{#4} _{#1}
        }{
            \bar{p}^{#4} _{#1}(#3)
        }
    }{
        \ifthenelse{\equal{#3}{}}{
            \bar{p}^{#4} _{#1}(\cdot|#2 )
        }{
            \bar{p}^{#4} _{#1}(#3|#2)
        }
    }}

\newcommandx\pnoise[4][4=]{
\ifthenelse{\equal{#2}{}}{
    \ifthenelse{\equal{#3}{}}{
        \upsilon^{#4} _{#1}
    }{
        \upsilon^{#4} _{#1}(#3)
    }
}{
    \ifthenelse{\equal{#3}{}}{
        \upsilon^{#4} _{#1}(\cdot|#2 )
    }{
        \upsilon^{#4} _{#1}(#3|#2)
    }
}}

\newcommandx\ipdata[4][4=]{
\ifthenelse{\equal{#2}{}}{
    \ifthenelse{\equal{#3}{}}{
        \bar{p}^{#4} _{#1}
    }{
        \bar{p}^{#4} _{#1}(#3)
    }
}{
    \ifthenelse{\equal{#3}{}}{
        \bar{p}^{#4} _{#1}(\cdot|#2 )
    }{
        \bar{p}^{#4} _{#1}(#3|#2)
    }
}}

\newcommandx\ratemat[4][4=]{
\ifthenelse{\equal{#2}{}}{
    \ifthenelse{\equal{#3}{}}{
        Q^{#4} _{#1}
    }{
        \lambda^{#4} _{#1}(#3)
    }
}{
    \ifthenelse{\equal{#3}{}}{
        Q^{#4} _{#1}(\cdot | #2)
    }{
        Q^{#4} _{#1}(#3 | #2)
    }
}}

\newcommandx\rratemat[4][4=]{
\ifthenelse{\equal{#2}{}}{
    \ifthenelse{\equal{#3}{}}{
        \bar Q^{#4} _{#1}
    }{
        \bar\lambda^{#4} _{#1}(#3)
    }
}{
    \ifthenelse{\equal{#3}{}}{
        \bar Q^{#4} _{#1}(\cdot | #2)
    }{
        \bar Q^{#4} _{#1}(#3 | #2)
    }
}}

\newcommandx\mdm[4][4=]{
\ifthenelse{\equal{#2}{}}{
    \ifthenelse{\equal{#3}{}}{
        p^{\mathsf{d}, #4} _{#1}
    }{
        p^{\mathsf{d}, #4} _{#1}(#3)
    }
}{
    \ifthenelse{\equal{#3}{}}{
        p^{\mathsf{d}, #4} _{#1}(\cdot|#2)
    }{
        p^{\mathsf{d}, #4} _{#1}(#3|#2)
    }
}}

\newcommandx\jpdata[4][4=]{
\ifthenelse{\equal{#2}{}}{
    \ifthenelse{\equal{#3}{}}{
        \tilde{p}^{#4} _{#1}
    }{
        \tilde{p}^{#4} _{#1}(#3)
    }
}{
    \ifthenelse{\equal{#3}{}}{
        \tilde{p}^{#4} _{#1}(\cdot|#2)
    }{
        \tilde{p}^{#4} _{#1}(#3|#2)
    }
}}

\newcommandx\cpdata[4][4=]{
    \ifthenelse{\equal{#2}{}}{
        \ifthenelse{\equal{#3}{}}{
            \pi^{#4} _{#1}
        }{
            \pi^{#4} _{#1}(#3)
        }
    }{
        \ifthenelse{\equal{#3}{}}{
            \pi^{#4} _{#1}(\cdot|#2)
        }{
            \pi^{#4} _{#1}(#3|#2)
        }
    }}

  \newcommandx\potg[4][4=]{
    \ifthenelse{\equal{#2}{}}{
        \ifthenelse{\equal{#3}{}}{
            g^{#4} _{#1}
        }{
            g^{#4} _{#1}(#3)
        }
    }{
        \ifthenelse{\equal{#3}{}}{
            g^{#4} _{#1}(\cdot|#2)
        }{
            g^{#4} _{#1}(#3|#2)
        }
    }}

\newcommandx\jcpdata[4][4=]{
    \ifthenelse{\equal{#2}{}}{
        \ifthenelse{\equal{#3}{}}{
            \bar\pi^{#4} _{#1}
        }{
            \bar\pi^{#4} _{#1}(#3)
        }
    }{
        \ifthenelse{\equal{#3}{}}{
            \bar\pi^{#4} _{#1}(\cdot|#2)
        }{
            \bar\pi^{#4} _{#1}(#3|#2)
        }
    }}

\newcommandx\dtilt[4][4=]{
    \ifthenelse{\equal{#2}{}}{
        \ifthenelse{\equal{#3}{}}{
            \pi^{d, #4} _{#1}
        }{
            \pi^{d, #4} _{#1}(#3)
        }
    }{
        \ifthenelse{\equal{#3}{}}{
            \pi^{d, #4} _{#1}(\cdot|#2)
        }{
            \pi^{d, #4} _{#1}(#3|#2)
        }
    }}

\newcommandx\tilt[4][4=]{
    \ifthenelse{\equal{#2}{}}{
        \ifthenelse{\equal{#3}{}}{
            \pi^{#4} _{#1}
        }{
            \pi^{#4} _{#1}(#3)
        }
    }{
        \ifthenelse{\equal{#3}{}}{
            \pi^{#4} _{#1}(\cdot|#2)
        }{
            \pi^{#4} _{#1}(#3|#2)
        }
    }}

\newcommand\updata[3]{
    \ifthenelse{\equal{#2}{}}{
        \ifthenelse{\equal{#3}{}}{
            \bar{p}_{#1}
        }{
            \bar{p}_{#1}(#3)
        }
    }{
        \ifthenelse{\equal{#3}{}}{
            \bar{p}_{#1}(\cdot|#2)
        }{
            \bar{p}_{#1}(#3|#2)
        }
    }}

\newcommand\phat[3]{
    \ifthenelse{\equal{#2}{}}{
        \ifthenelse{\equal{#3}{}}{
            \hat{p}_{#1}
        }{
            \hat{p}_{#1}(#3)
        }
    }{
        \ifthenelse{\equal{#3}{}}{
            \hat{p}_{#1}(\cdot|#2)
        }{
            \hat{p}_{#1}(#3|#2)
        }
    }}

\newcommandx\dfw[4][4=]{
        \ifthenelse{\equal{#3}{}}{
            q^{d, #4} _{\smash{#1}}(\cdot|#2)
        }{
            q^{d, #4} _{\smash{#1}}(#3|#2)
        }
    }

\newcommandx\fw[4][4=]{
    \ifthenelse{\equal{#2}{}}{
        \ifthenelse{\equal{#3}{}}{
            q^{#4} _{#1}
        }{
            q^{#4} _{#1}(#3)
        }
        }{
        \ifthenelse{\equal{#3}{}}{
            q^{#4} _{\smash{#1}}(\cdot | #2 )
        }{
            q^{#4} _{\smash{#1}}(#3 | #2 )
        }
        }
    }

\newcommandx\barfw[4][4=]{
    \ifthenelse{\equal{#2}{}}{
        \ifthenelse{\equal{#3}{}}{
            \bar{q}^{#4} _{#1}
        }{
            \bar{q}^{#4} _{#1}(#3)
        }
        }{
        \ifthenelse{\equal{#3}{}}{
            \bar{q}^{#4} _{\smash{#1}}(\cdot | #2 )
        }{
            \bar{q}^{#4} _{\smash{#1}}(#3 | #2 )
        }
        }
    }
\newcommandx\fwp[3][3=]{
        \ifthenelse{\equal{#3}{}}{
            \bq_{\smash{#1}}(#2)
        }{
            \bq^{#3}_{\smash{#1}}(#2)
        }
    }

\newcommandx\sfwp[3][3=]{
    \ifthenelse{\equal{#3}{}}{
        \tilde\bq_{\smash{#1}}(#2)
    }{
        \tilde\bq^{#3}_{\smash{#1}}(#2)
    }
}
\newcommand{\simplex}{\Delta}
\newcommand{\Simplex}[1]{\simplex_{#1}}

\newcommandx\denoiser[5][4=, 5=0]{
    \ifthenelse{\equal{#2}{}}{
        \ifthenelse{\equal{#3}{}}{
            \ifthenelse{\equal{#1}{}}{
                \hat{\bx}^{#4}_{#5}
            }{
                \hat\bx^{#4}_{#5}(\cdot, #1)
            }
        }{
            \hat\bx^{#4} _{#5}(#3, #1)
        }
    }{
        \ifthenelse{\equal{#3}{}}{
            \hat\bx^{#4} _{#5}(\cdot, #1|#2)
        }{
            \hat\bx^{#4} _{#5}(#3, #1|#2)
        }
    }
}

\newcommandx\student[4][3=\phi, 4=]{
    \ifthenelse{\equal{#4}{}}{
        \ifthenelse{\equal{#1}{}}{
            \ifthenelse{\equal{#2}{}}{
                p_0^{#3}
            }{
                p_0^{#3}(\cdot\,|#2)
            }
        }{
            p_0^{#3}(\cdot\,|#2,#1)
        }
    }{
        \ifthenelse{\equal{#1}{}}{
            \ifthenelse{\equal{#2}{}}{
                p_0^{#3,#4}
            }{
                p_0^{#3,#4}(\cdot\,|#2)
            }
        }{
            p_0^{#3,#4}(\cdot\,|#2,#1)
        }
    }
}

\newcommandx\matrixdenoiser[5][4=, 5=0]{
    \ifthenelse{\equal{#2}{}}{
        \ifthenelse{\equal{#3}{}}{
            \hat\bx_{0\tbar #1}(\cdot)
        }{
            \hat\bx_{0\tbar #1}(#3)
        }
    }{
        \ifthenelse{\equal{#3}{}}{
            \hat\bx_{0\tbar #1}(\cdot|#2)
        }{
            \hat\bx_{0\tbar #1}(#3|#2)
        }
    }
}

\newcommandx\loodenoiser[5][4=, 5=0]{
    \ifthenelse{\equal{#2}{}}{
        \ifthenelse{\equal{#3}{}}{
            \hat\bx^{\loo}_{0\tbar #1}(\cdot)
        }{
            \hat\bx^{\loo}_{0\tbar #1}(#3)
        }
    }{
        \ifthenelse{\equal{#3}{}}{
            \hat\bx^{\loo}_{0\tbar #1}(\cdot|#2)
        }{
            \hat\bx^{\loo}_{0\tbar #1}(#3|#2)
        }
    }
}

    \newcommandx\hdenoiser[4][4=]{
    \ifthenelse{\equal{#2}{}}{
        \ifthenelse{\equal{#3}{}}{
            \hat{D}^{#4}_{#1}
        }{
            \hat{D}^{#4} _{#1}(#3)
        }
    }{
        \ifthenelse{\equal{#3}{}}{
            \hat{D}^{#4} _{#1}(\cdot|#2)
        }{
            \hat{D}^{#4} _{#1}(#3|#2)
        }
    }}

\newcommandx\epspred[4][4=]{
    \ifthenelse{\equal{#2}{}}{
        \ifthenelse{\equal{#3}{}}{
            \varepsilon^{#4}_{#1}
        }{
            \varepsilon^{#4} _{#1}(#3)
        }
    }{
        \ifthenelse{\equal{#3}{}}{
            \varepsilon^{#4} _{#1}(\cdot|#2)
        }{
            \varepsilon^{#4} _{#1}(#3|#2)
        }
    }}

\newcommand\clf[3]{
    \ifthenelse{\equal{#2}{}}{
        g_{#1}(#3|\cdot)
    }{
        g _{#1}(#3|#2)
    }
}

\newcommand\cfgdist[3]{
    \ifthenelse{\equal{#2}{}}{
        p^w _{#1}(#3|\cdot)
    }{
        p^w _{#1}(#3|#2)
    }
}

\newcommand\cscore[3]{
    \ifthenelse{\equal{#2}{}}{
        \ifthenelse{\equal{#3}{}}{
            s _{#1}
        }{
            s _{#1}(#3)
        }
    }{
        \ifthenelse{\equal{#3}{}}{
            s _{#1}(\cdot|#2)
        }{
            s _{#1}(#3|#2)
        }
    }}

 \def\bpi{\pi}

\def\interpscale{\eta}

\def\categorical{\mathrm{Cat}}

\def\bx{x}
\def\be{e}

\def\by{y}

\def\bq{\mathbf{q}}

\def\algoname{{\sc{Exponential Residual Flow}}}

\def\param{\theta}

\newcommandx{\hpredx}[3][2=0,3=\param]{\smash{m^{#3} _{#2|#1}}}
\newcommandx{\predx}[2][2=0]{\smash{m _{#2|#1}}}
\newcommandx{\prednoise}[2][2=\param]{\smash{\epsilon^{#2} _{#1}}}

\newcommandx\score[4][4=]{
\ifthenelse{\equal{#2}{}}{
    \ifthenelse{\equal{#3}{}}{
        s^{#4} _{#1}
    }{
        s^{#4} _{#1}(#3)
    }
}{
    \ifthenelse{\equal{#3}{}}{
        s^{#4} _{#1}(\cdot|#2)
    }{
        s^{#4} _{#1}(#3|#2)
    }
}}

\makeatletter
\@ifpackageloaded{todonotes}{}{%
    \usepackage[disable,textsize=tiny,textwidth=\marginparwidth]{todonotes}%
}
\makeatother
\makeatletter
\tikzset{
    notestyleraw/.append style={
        fill opacity=0.8,    
        text opacity=1       
    }
}
\makeatother
\newcommand{\alain}[1]{\todo[linecolor=cyan,backgroundcolor=cyan!25,bordercolor=cyan]{\tiny{\textbf{AD}: #1}}}

\newcommand{\coint}[1]{\left[#1\right)}
\newcommand{\ocint}[1]{\left(#1\right]}
\newcommand{\ooint}[1]{\left(#1\right)}
\newcommand{\ccint}[1]{\left[#1\right]}

\renewcommand{\det}{\mbox{det}}

\def\tx{\tilde{x}}

\newcommand\oh[1]{\tilde{#1}}

\def\unif{\mathrm{Unif}}

\ifdefined\erfslides

\fi

\newcommand{\representer}[1]{#1}
\newcommand{\PP}{\mathbb{P}}

%% file: body/introduction.tex
\section{Introduction}

Discrete diffusion models have emerged as a compelling alternative to autoregressive models. By allowing multiple positions to be updated in parallel, they offer a path toward faster decoding \citep{lou2024discrete,sahoo2024simple} and potentially more efficient steering and control \citep{schiff2024simple}. In continuous time, these models are described by forward and reverse continuous-time Markov chains \citep{campbell2022continuous,lou2024discrete}. Standard simulation of the reverse process is inherently stochastic: Gillespie's algorithm draws fresh randomness at each jump \citep{gillespie1977exact}, while ancestral sampling of discrete-time reverse chains likewise relies on stochastic transitions \citep{shi2024simplified,sahoo2024simple,hoogeboom2021argmax,schiff2024simple}.

This raises a natural question: can one design a discrete diffusion sampler whose only randomness is the initial condition, as in the probability-flow ODE for continuous diffusion models \citep{song2021score,song2021ddim}? A direct analogue on a finite state space is impossible because a deterministic map cannot split probability mass. Continuous auxiliary randomness is therefore required at initialization, particularly for masked diffusion, whose terminal distribution is a Dirac mass. Existing uses of auxiliary randomness serve different purposes. The projected Gaussian trajectories of \citet{sahoo2025diffusion} act as more of a reparameterization and do not define a deterministic flow driven solely by the current discrete state. Gumbel Distillation \citep{zhang2026gumbeldistillation} instead uses Gumbel noise to construct deterministic training targets from an autoregressive teacher for parallel decoding, rather than a marginal-preserving deterministic flow for a discrete diffusion process.

Our contributions are as follows:

\begin{enumerate}[wide,label=$\bullet$]
\item  Our central contribution is a new deterministic sampler for discrete diffusion models, which we call the \emph{exponential residual flow} (ERF). We augment the state with non-negative auxiliary variables, called residuals, initialized as independent exponential random variables, one for each token and position. Their evolution is then governed by an ODE determined by the forward and reverse rates of the discrete dynamics that governed jumps of the discrete state. A token change is triggered whenever a residual reaches zero. In contrast to Gillespie's first-reaction construction, where residual clocks are refreshed after each jump, ERF continues to evolve all remaining residuals deterministically. Consequently, once the initial augmented state has been sampled, no additional randomness is required.
We prove that these augmented dynamics preserve the marginal distribution of the original CTMC at every time. At each fixed time, the residuals remain independent standard exponential random variables and are independent of the discrete state. The resulting process generally differs from the reverse Markov chain and need not be Markovian.

\item To make this construction applicable to learned diffusion models, we introduce a novel numerical integration scheme that freezes the leave-one-out denoiser over each sampling interval while preserving the known time dependence of the transition rates. We derive explicit formulations for both masked and uniform diffusion models, and show that the resulting scheme is exact for factorized data distributions when the denoiser is exact.

\item We evaluate ERF against ancestral sampling under matched model-evaluation budgets and find comparable generation quality. The deterministic nature of ERF also makes it directly amenable to consistency distillation. We exploit this property to train few-step student models and evaluate the resulting speed--quality trade-offs on text generation, code generation, and Sudoku completion.
\end{enumerate}

%% file: body/background.tex
\paragraph{Notation.}
We consider integer-valued sequences in $\msx=\msv^\len$, where $\msv=\intset{1}{\vocab}$, and write $\Simplex{\vocab}=\{p\in\rset^\vocab:p^k\geq0,\ \sum_{k=1}^{\vocab}p^k=1\}$ for the probability simplex. For $\bx\in\msx$, $\bx^\ell$ denotes its $\ell$-th token and $\bx^{-\ell}\in\msv^{\len-1}$ the sequence with position $\ell$ removed. For masked diffusion, the mask token is $\mask\eqdef\vocab$.
For any vector $x=(x^1,\ldots,x^d)\in\msv^d$, we denote its one-hot matrix by $\oh{x}\in\{0,1\}^{d\times\vocab}$, with entries $\oh{x}^{\ell k}\eqdef\indic_{x^\ell=k}$. For a single token $x\in\msv$, $\oh{x} \in \rset^\vocab$ denotes its one-hot vector.
For $\pi\in\Simplex{\vocab}$, $\categorical(\pi)$ denotes the categorical distribution, and $\categorical(x;\pi)\eqdef \ps{\oh{x}}{\pi}$ is the probability of $x$ under $\categorical(\pi)$. We use the convention $\sup\,\emptyset=-\infty$.
\begin{figure}[t]
    \centering
    \includegraphics[width=\linewidth]{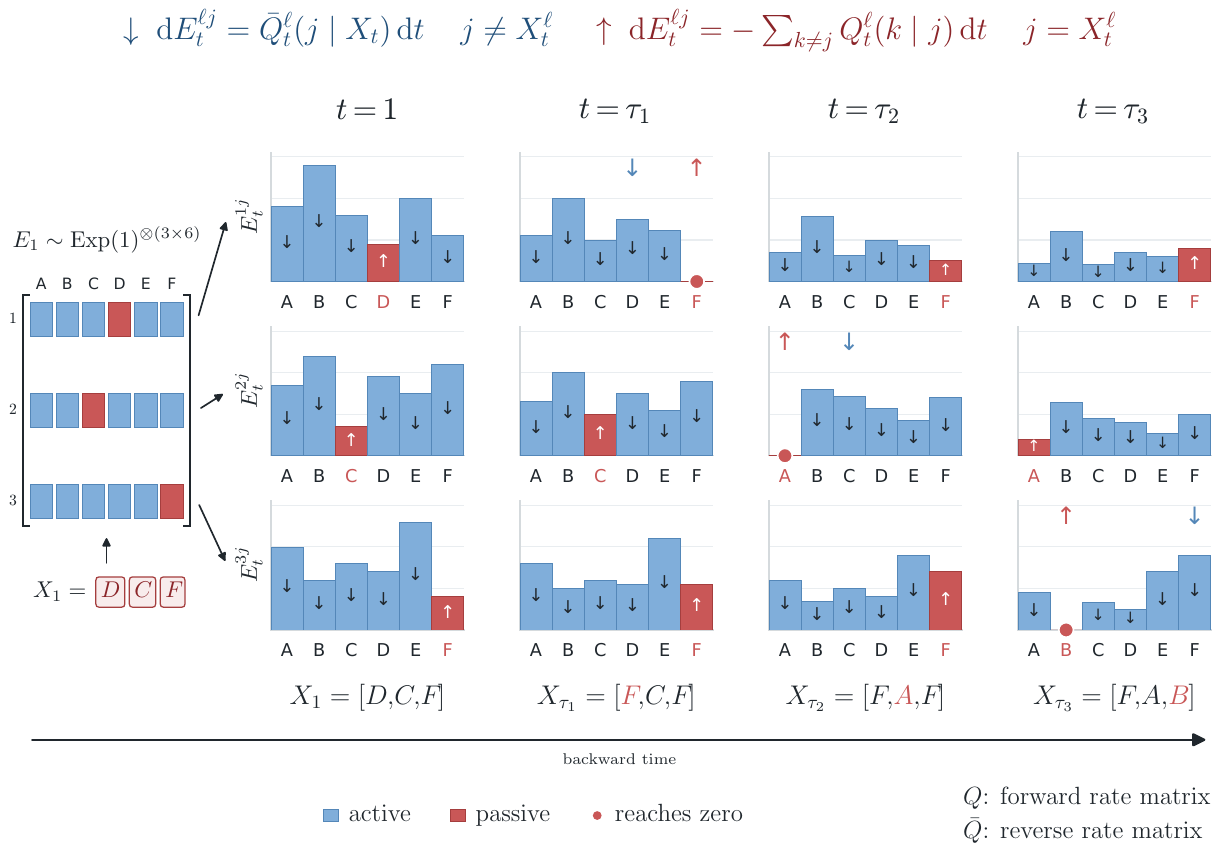}
    \captionsetup{font=small}
    \caption{\textbf{The Exponential Residual Flow algorithm.} Evolution of the joint state $(X_t,E_t)$, initialized independently with $X_1\sim \pdata{1}{}{}$ and $E_1\sim\operatorname{Exp}(1)^{\otimes (3\times 6)}$. Each histogram displays the residual values at one position, which increase (red) or decrease (blue) \emph{deterministically} as $t$ decreases. A residual reaching zero triggers a single token change in $X_t$.}
    \label{fig:erf-illustration}
\end{figure}

\section{Background on discrete diffusion models}
We first briefly introduce discrete diffusion models \citep{sahoo2024simple,shi2024simplified,austin2021structured,hoogeboom2021argmax,schiff2024simple}. This class of generative models constructs a stochastic process that transports a simple reference distribution,
$\smash{\pdata{1}{}{} \eqdef \prod_{\ell=1}^{\len}\categorical(\bpi^\ell)}$, with $\smash{\bpi\in(\Simplex{\vocab})^{\len}}$, to a target data distribution $\pdata{0}{}{}$. To this end, we use a forward process that progressively corrupts the data distribution, ultimately transforming it into $\pdata{1}{}{}$. The generative model is then obtained by learning an approximation of the corresponding reverse process.

 We consider as forward process a Continuous Time Markov Chain (CTMC) process $(Y_t)_{t \in [0,1]}$ started at $\pdata{0}{}{}$ with transitions
$\smash{\fw{t\tbar s}{\bx_s}{\bx_t}[] = \prod_{\ell=1}^L \fw{t\tbar s}{\bx_s^{\ell}}{\bx_t^{\ell}} [\ell]}$, where $x_s,x_t \in \msx$, $0 \leq s < t \leq 1$  and
\begin{equation}
    \label{eq:transition}
    \fw{t\tbar s}{\bx_s^{\ell}}{\bx_t^{\ell}}[\ell]
    =
    \categorical\!\left(\bx_t^{\ell};\alpha_{t\tbar s} \oh{\bx}_s^{\ell}+(1-\alpha_{t\tbar s})\bpi^{\ell}\right)\eqsp, \quad 0\leq s < t \leq 1 \eqsp,
\end{equation}
where $\tilde{x}_s$ denotes the one-hot encoding of $x_s$ and $(\alpha_t)_{t\in[0,1]}$ is a monotone noise schedule with $\alpha_0=1$, $\alpha_1 \approx 0$, and $\alpha_{t\tbar s}= \alpha_t/\alpha_s$.
The time-reversal of this process is obtained by reversing the direction of time and is defined as the process $(Y_{t_-})_{t \in [0,1]}$, backward in time\footnote{Equivalently, we could define the time-reversal as $Y_{1-t_-}$ forward in time with transitions $\pdata{1-s\tbar 1-t}{}{}$} with $Y_{t_-}=\lim_{h\downarrow0}Y_{t-h}$. For ease of notation, we denote $Y_{t_-}$ by $X_t$. Since $Y_{t_-} = Y_t$ almost surely for any $t$, the process $X_t$ has marginals $\pdata{t}{}{}$, and allows by construction to transport the source distribution $\pdata{1}{}{}$ to the target $\pdata{0}{}{}$. In addition, the process $(X_{t})_{t \in [0,1]}$ is itself a CTMC with transitions
\begin{equation}
    \label{eq:reverse}
    \txts \pdata{s\tbar t}{\bx_t}{\bx_s}
    =
    \sum_{\bx_0}
    \fw{s\tbar 0,t}{\bx_0,\bx_t}{\bx_s}\,
    \pdata{0\tbar t}{\bx_t}{\bx_0} \eqsp,
\end{equation}
with $\pdata{0 \tbar t}{\bx_t}{\cdot}$ the denoising posterior and $\fw{s\tbar 0, t}{}{}$ the conditional distribution of $Y_s$ given $(Y_0, Y_t)$, whose explicit expression is given in \Cref{sec:bridge-simplex}. During training, these reverse transitions are learned using a factorized model parameterized by $\param$: $\pdata{s\tbar t}{\bx_t}{\bx_s}[\param] \eqdef \prod_{\ell = 1} ^\len \pdata{s\tbar t}{\bx_t}{\bx^\ell _s}[\param, \ell]$. To obtain a practical model for inference, we then consider the following joint distribution:
\begin{equation}
    \label{eq:plugin-reverse}
     \txts \pdata{0:n}{}{\bx_{\tk{0:n}}}[\param] = \pdata{1}{}{\bx_\tk{\last}} \prod_{i = 1} ^\last \pdata{\sk{i}\tbar \tk{i}}{\bx_{\tk{i}}}{\bx_{\sk{i}}}[\param] \eqsp,
\end{equation}
for a choice of timesteps $0 = t_0 < \ldots < t_n = 1$, setting $s_i = t_{i - 1}$ for notational convenience. A common parameterization used in the literature \citep{shi2024simplified,sahoo2024simple,hoogeboom2021argmax,schiff2024simple} is $\pdata{s\tbar t}{\bx_t}{\bx_s}[\param] = \fw{s\tbar 0, t}{\denoiser{t}{}{\bx_t}[\param], \bx_t}{\bx_s}$, where $\smash{\fw{s\tbar 0,t}{}{}}$ denotes the extension of the bridge transition in \eqref{eq:bridge-prob} to simplex-valued inputs, and $\denoiser{}{}{}[\param]$ denotes a neural network with values in $(\Simplex{\vocab})^{\len}$. In what follows we assume, without any loss of generality, that it  approximates the leave-one-out denoiser, whose $\ell$-th component is $\smash{(t,\bx_t)\mapsto \pE[\oh{X}_0^\ell\mid X_t^{-\ell}=\bx_t^{-\ell}]}$; see \Cref{sec:bridge-simplex}.
\paragraph{Rate matrices and discrete score.} Another useful perspective on discrete diffusion is provided through rate matrices associated to the forward and backward processes \citep{campbell2022continuous,sun2023scorebased,lou2024discrete}. The rate matrix associated to the forward transitions \eqref{eq:transition} is given for any $\bx, \by \in \msx$ and $t\in\ccint{0,1}$ by
\begin{equation}
    \label{eq:forward-rate}
    \txts \ratemat{t}{\bx}{\by} = \sum_{\ell = 1}^\len \indic_{\by^{-\ell} = \bx^{-\ell}} \ratemat{t}{\bx^\ell}{\by^\ell}[\ell] \eqsp, \quad \text{where} \quad \ratemat{t}{x}{y}[\ell] = (-\dot\alpha_t / \alpha_t) \ps{\oh{y}}{\bpi^\ell} \quad (x\neq y)\eqsp.
\end{equation}
Here $x,y\in\msv$, and $\ratemat{t}{x}{x}[\ell] = - \sum_{y \neq x} \ratemat{t}{x}{y}[\ell]$ for all $\ell \in \intset{1}{\len}$. Let $(\pdata{t}{}{})_{t\in\ccint{0,1}}$ denote the family of marginals of $(Y_t)_{t \in [0,1]}$ with transition probabilities given by \eqref{eq:transition} and initial distribution $\pdata{0}{}{}$.  By definition, $(\pdata{t}{}{})_{t\in\ccint{0,1}}$ satisfies the Kolmogorov equation \citep[Theorem 1]{feller1940integro}: for $t\in\ooint{0,1}$ and $\bx,\by \in \msx$,
\begin{equation}
    \label{eq:kolmogorov}
    \txts \partial_t \pdata{t}{}{\by} = \sum_{\bx \in \msx} \ratemat{t}{\bx}{\by} \pdata{t}{}{\bx} \eqsp.
\end{equation}
Note that the forward transition $\fw{t\tbar s}{\bx_s}{}$ specified in \eqref{eq:transition} are obtained by solving \eqref{eq:kolmogorov} started from $\updelta_{\bx_s}$ at time $s$. The family of rate matrices associated to the backward process is $(\rratemat{t}{}{})_{t\in[0,1]}$ \footnote{Equivalently, the rate matrices associated with the forward in time convention correspond to $(\rratemat{1-t}{}{})_{t\in[0,1]}$}, where $\rratemat{t}{}{}$ satisfies for any $\by\neq\bx$, such that $\pdata{t}{}{\bx}>0$\footnote{The definition of $\rratemat{t}{}{}$ is arbitrary when $\pdata{t}{}{\bx}=0$.},
\begin{equation}
    \label{eq:reverse-rate}
    \rratemat{t}{\bx}{\by} \eqdef \frac{\pdata{t}{}{\by}}{\pdata{t}{}{\bx}} \ratemat{t}{\by}{\bx} = \sum_{\ell = 1}^\len \indic_{\by^{-\ell} = \bx^{-\ell}} \frac{\pdata{t}{\bx^{-\ell}}{\by^\ell}[\ell]}{\pdata{t}{\bx^{-\ell}}{\bx^\ell}[\ell]} \ratemat{t}{\by^\ell}{\bx^\ell}[\ell] \eqsp,
\end{equation}
with $\rratemat{t}{\bx}{\bx} \eqdef \bar\lambda_t(\bx) = - \sum_{\by\neq\bx} \rratemat{t}{\bx}{\by}$. Using the definition of the reverse transitions in \eqref{eq:reverse}, \eqref{eq:kolmogorov} and \eqref{eq:reverse-rate}, we obtain that the reverse transitions solve the following Kolmogorov equation: for any $x_t,y\in\msx$, $\pdata{t}{}{\bx_t} >0$,
\begin{equation}
    \label{eq:reverse-kolmogorov}
    \txts \partial_s \pdata{s\tbar t}{\bx_t}{\by}
    = -\sum_{\bx} \,\rratemat{s}{\bx}{\by} \pdata{s\tbar t}{\bx_t}{\bx} \eqsp,
    \qquad
    \pdata{t\tbar t}{\bx_t}{\by} = \indic_{\bx_t=\by} \eqsp.
\end{equation}


%% file: body/deterministic_sampler.tex
\section{Deterministic sampling via exponential residual flows}
\label{sec:exponential-residual-flow}

In this section, we construct a sampler whose trajectory is deterministic after a single random initialization and with one-time marginals agreeing with those of the reverse CTMC. The key idea is to augment the discrete state with the exponential variables used in Gillespie's first-reaction method and replace their resampling at jump times with a deterministic continuous evolution, resulting in the \algoname\ sampler described in \Cref{alg:gillespie-first-reaction}.

\subsection{Exponential Residual Flow}

The CTMC $(X_t)_{t\in[0,1]}$ is piecewise constant and jumps at random times. We start by presenting Gillespie's algorithm which allows exact simulation of this CTMC. It simulates the succesive jumps of $X_t$ using auxiliary exponential random variables that define random jump times. These variables introduce randomness in the trajectory of $X_t$, we show that it is possible to instead recycle the initial exponential variables according to an ODE. 

\paragraph{Gillespie's algorithm.}
\label{sec:gillespie-exponential-race}
Gillespie's global algorithm \citep{gillespie1977exact} constructs the piecewise constant trajectory of the reverse CTMC one jump at a time. Starting from $X_t=x$, it first determines the next backward jump time $\tau<t$, with $X_u=x$ for $u\in(\tau,t]$. To do so, it draws $E\sim\operatorname{Exp}(1)$ and solves $\smash{\int_\tau^t\bar\lambda_u(x)\eqsp\rmd u=E}$. At time $\tau$, the trajectory jumps by changing a single token: the algorithm selects a position $\ell\in[\len]$ and a new token $j\neq x^\ell$ with probability $\rratemat{\tau}{x}{j}[\ell]/\bar\lambda_\tau(x)$. It sets $X_\tau^\ell=j$, leaves all other positions unchanged, and repeats from time $\tau$ and state $X_\tau$; see \Cref{app:ctmc-gillespie} for the formal construction.
\begin{algorithm}[t]
    \caption{Gillespie next-reaction / ERF algorithm.}
    \label{alg:gillespie-first-reaction}
    \label{alg:erf-sampler}
    \begin{algorithmic}[1]
        \Require $0\leq s\leq t \leq 1$, reverse rates $(\rratemat{u}{}{})_{u\in\ccint{s,t}}$, forward exit rates $(\lambda_u) _{u \in \ccint{s, t}}$
        \Output $(X_s, E_s)$ at time $s$.
        \State \textbf{Initialize:} $(X_t,E_t)\sim p_t\otimes \mathrm{Exp}^{\otimes (L\times K)}$, \,\, $X_s \gets X_t \eqsp, \, \, \tau^\star \gets t$, \,\, $E_s \gets E_t$ \label{line:gillespie-erf-initialization}
        \While{$\tau^\star > s$}
            \State $\tau^{\ell j}\gets\sup \big\{u\in[s,\tau^\star]:\int_u^{\tau^\star}\rratemat{r}{X_s}{j}[\ell]\eqsp\rmd r\geq E_s^{\ell j}\big\}$ for $j\neq X_s^\ell$ \label{line:candidate-jumps}\label{line:erf-integrated-reverse-rate} \Comment{\textcolor{crimson}{in parallel}}
            \State $(\ell^\star,j^\star)\gets\argmax_{(\ell,j):j\neq X_s^\ell}\,\,\tau^{\ell j}$  \label{line:next-event-selection}
            \State $\tau^\star_{\mathrm{new}}\gets\max\{s,\tau^{\ell^\star j^\star}\}$
            \State $E_s^{\ell j} \gets E_s^{\ell j} - \int_{\tau^{\star}_{\mathrm{new}}}^{\tau^{\star}} \rratemat{r}{X_s}{j}[\ell]\eqsp\rmd r$, \quad \, $j\neq X_s^\ell$ \Comment{\textcolor{crimson}{update active residuals}} \label{line:erf-active-residual-update}
            \If{$X_s^\ell=j$} \Comment{\textcolor{crimson}{update passive residuals}} \label{}
                \State \textbf{either} $E_s^{\ell j} \sim \mathrm{Exp}(1)$ \Comment{\textcolor{crimson}{Gillespie}} \label{line:gillespie-passive-residual-refresh}
                \State \tikz[baseline=(erfline.base)]\node[fill=erflineblue,rounded corners=2pt,inner xsep=2pt,inner ysep=0.4pt,outer sep=0pt] (erfline) {\textbf{or} $E_s^{\ell j} \gets E_s^{\ell j} + \int_{\tau^{\star}_{\mathrm{new}}}^{\tau^{\star}} \lambda_r^{\ell}(j)\eqsp\rmd r$}; \Comment{\textcolor{crimson}{ERF (ours)}} \label{line:erf-passive-residual-update}
            \EndIf
            \If{$\tau^{\ell^\star j^\star}\geq s$}
                \State $X_s^{\ell^\star} \gets j^\star$ \label{line:gillespie-erf-token-update}
            \EndIf
            \State $\tau^{\star} \gets \tau^{\star}_{\mathrm{new}}$

        \EndWhile
        \State \Return $X_s, E_s$
    \end{algorithmic}
\end{algorithm}

Note that this construction has two sources of randomness: the exponential draw determining the jump time and the categorical draw determining the token change. Our goal is to eliminate randomness beyond initialization. We now describe Gillespie's \emph{first-reaction} algorithm, which removes the need for the categorical draw \citep{gillespie1977exact,anderson2007modified}. Given $X_t=x$, draw independent variables $E^{\ell j}\sim\operatorname{Exp}(1)$ for every position $\ell\in[\len]$ and token $j\neq x^\ell$, and determine the candidate jump times from
\begin{equation}
    \label{eq:first-reaction-candidate-time}
    \int_{\tau^{\ell j}}^t
    \rratemat{u}{x}{j}[\ell]\eqsp\rmd u
    = E^{\ell j}.
\end{equation}
The next jump encountered backward from $t$ is the largest among these jump times, so we select $\smash{(\ell^\star,j^\star)\in\argmax_{(\ell,j):j\neq x^\ell}\tau^{\ell j}}$ and set $\smash{\tau=\tau^{\ell^\star j^\star}}$. The trajectory remains equal to $x$ on $(\tau,t]$; at time $\tau$, the algorithm sets $\smash{X_\tau^{\ell^\star}=j^\star}$, leaves all other positions unchanged, and repeats from time $\tau$ and state $X_\tau$. This construction simulates the reverse CTMC without categorical sampling, at the expense of drawing $\len(\vocab-1)$ i.i.d. standard exponential variables at each jump.

\fixedspaceparagraph{Next-reaction method.} The next-reaction method reuses the exponential variables associated with token changes that were not selected \citep{gibson2000efficient,anderson2007modified,durmus2021piecewisedeterministic}. To see how, we can express the calculation of the next jump as a continuous evolution. Given $X_t=x$, start from $E_t^{\ell j}=E^{\ell j}$ and evolve all variables with $j\neq x^\ell$ simultaneously according to $$\rmd E_u^{\ell j}/\rmd u=\rratemat{u}{x}{j}[\ell] \eqsp,$$ 
with time evolving backwards. We call these evolving variables \emph{residuals}. The first residual to reach zero determines the jump time $\tau$; its indices $(\ell^\star,j^\star)$ determine the position and new token. This gives the same jump as solving \eqref{eq:first-reaction-candidate-time} for every pair and selecting the largest candidate time in line~\ref{line:next-event-selection}. Throughout this evolution, the discrete state stays equal to $x$ until time $\tau$, when line~\ref{line:gillespie-erf-token-update} sets $X_\tau^{\ell^\star}=j^\star$, leaving all other positions unchanged. By memorylessness, the residuals for the other token changes are conditionally i.i.d. standard exponential variables given the discrete trajectory, so they can be reused for the newt jump, only the residual that reached zero needs to be resampled.

We use a variant of this construction, given in \Cref{alg:gillespie-first-reaction}, that maintains one residual for every position--token pair, initialized in line~\ref{line:gillespie-erf-initialization}. Residuals with $j\neq x^\ell$ are \emph{active} and are updated in line~\ref{line:erf-active-residual-update}. Those with $j=x^\ell$ are \emph{passive} and are resampled in line~\ref{line:gillespie-passive-residual-refresh}. This variant requires $\len$ i.i.d. standard exponential draws at each jump.

\input{body/sec3.1v2}


\subsection{Integrating the exponential residual flow}
\label{sec:integrating-exponential-residual-flow}

We now describe the practical implementation of \Cref{alg:erf-sampler}. For $\by\neq\bx$, the reverse rates in \eqref{eq:reverse-rate} are commonly approximated as
\begin{equation}
    \label{eq:score-parameterized-reverse-rate}
    \txts \rratemat{t}{\bx}{\by}[\theta]
    \eqdef
    \sum_{\ell=1}^{\len}
    \indic_{\by^{-\ell}=\bx^{-\ell}}
    \ps{\oh{\by}^\ell}{\score{t}{}{\bx}[\theta]^\ell}
    \ratemat{t}{\by^\ell}{\bx^\ell}[\ell]
    \eqsp,
\end{equation}
where the score approximation $\score{t}{}{\bx}[\theta]^\ell$ estimates the density ratios on the \rhs\ in \eqref{eq:reverse-rate} \citep{lou2024discrete}. We must then estimate the integrals of these approximate reverse rates in lines~\ref{line:erf-integrated-reverse-rate}-\ref{line:erf-active-residual-update} and those of the associated forward exit rates in line~\ref{line:erf-passive-residual-update}. However, exact integration may be computationally challenging or infeasible.

To address this problem, most samplers introduce discretization strategies for these integration steps and consider a grid $0=\tk{0}<\cdots<\tk{\last}=1$. A first option is to adapt the $\tau$-leaping algorithm by defining an approximate process successively on $\ccint{s_i,t_i}$: given $X_{t_i}$, $(X_{u})_{u\in\ccint{s_i,t_i}}$ is the CTMC with constant reverse rate $\rratemat{u}{\bx}{\by; \tk{i}}[\param] = \rratemat{\tk{i}}{X_\tk{i}}{\by}[\param]$; more details are provided in \Cref{app:ctmc-euler-tau}. Another option, given the form of the reverse rate and its approximation \eqref{eq:score-parameterized-reverse-rate}, is to freeze only the score approximation at time $t_i$, while retaining the forward rates $\ratemat{u}{\bx^\ell}{\by^\ell}[\ell]$ in \eqref{eq:forward-rate}, whose time integrals are available in closed form, similarly to \citep{conforti2025non}. We develop a different alternative here.
First, note that the score is related to the leave-one-out denoiser by
\begin{equation}
    \label{eq:score_loo}
    \langle \oh{\by}^\ell, \score{t}{}{\bx_t}[\param]^\ell \rangle = \fw{t\tbar 0}{\denoiser{t}{}{\bx_t}[\param]^\ell}{\by^\ell} \big/ \fw{t\tbar 0}{\denoiser{t}{}{\bx_t}[\param]^\ell}{\bx^\ell _t} \eqsp. 
\end{equation}
This relation suggests to freeze only the leave-one-out denoiser at time $t_i$ since if we plug \eqref{eq:score_loo} in \eqref{eq:score-parameterized-reverse-rate}, the integration with respect to time of the resulting approximate reverse rate can be made explicit for most models of interest. This choice has the advantage that it yields an exact sampler for a single token or a factorized distribution when the leave-one-out predictor is optimal contrary to other freezing strategies; see \Cref{sec:loo-freezing-exactness}.

\paragraph{Practical implementation.} Given $\smash{(\hat X_{\tk{i}},\hat E_{\tk{i}})}$, a single discretization step produces $\smash{(\hat X_{\sk{i}},\hat E_{\sk{i}})}$ by applying \Cref{alg:erf-sampler} on $[\sk{i},\tk{i}]$ with the following approximate reverse rate matrix:
\begin{mdframed}[style=propFrame,nobreak=true,innertopmargin=2pt,innerbottommargin=2pt]
\setlength{\abovedisplayskip}{2pt}
\setlength{\belowdisplayskip}{2pt}
\setlength{\abovedisplayshortskip}{0.75pt}
\setlength{\belowdisplayshortskip}{1pt}
\begin{equation}
    \label{eq:loo-frozen-reverse-rate}
     \rratemat{r}{\bx}{\by; \textcolor{crimson}{\tk{i}}}[\theta]
    \eqdef
    \sum_{\ell=1}^{\len}
    \indic_{\by^{-\ell}=\bx^{-\ell}}
    \frac{\fw{r\tbar0}{\denoiser{\textcolor{crimson}{\tk{i}}}{}{\hat{X}_{\textcolor{crimson}{\tk{i}}}}[\theta]^\ell}{\by^\ell}[\ell]}
         {\fw{r\tbar0}{\denoiser{\textcolor{crimson}{\tk{i}}}{}{\hat{X}_{\textcolor{crimson}{\tk{i}}}}[\theta]^\ell}{\bx^\ell}[\ell]}
    \ratemat{r}{\by^\ell}{\bx^\ell}[\ell]
    \eqsp, \qquad r\in[\sk{i},\tk{i}].
\end{equation}
\end{mdframed}
For the forward exit rates, we keep the original rates $(\lambda^\ell _r)_{r\in[\sk{i}, \tk{i}]}$. In general, discretizing ERF requires the forward rates obtained by time-reversing the frozen reverse process on $[\sk{i}, \tk{i}]$, and using those of the original forward process, \emph{e.g.} \eqref{eq:forward-rate}, can lead to infinitely many jumps. For leave-one-out freezing, however, \eqref{eq:loo-frozen-reverse-rate} is the exact time reversal, on $[\sk{i}, \tk{i}]$, of the original forward process started at the factorized distribution given by $\denoiser{{\tk{i}}}{}{\hat{X}_{{\tk{i}}}}[\theta]$. Each interval is therefore an exact ERF step for this distribution and has finitely many jumps; see \Cref{app:ctmc-partial-freezing}.

\paragraph{Specialization to MDM and UDM.}
\label{sec:erf-mdm-udm}

Having specified the partially frozen reverse rate matrix in \eqref{eq:loo-frozen-reverse-rate}, we now specify the calculation of jump times in line 3-4 of \Cref{alg:erf-sampler}  to MDM and UDM. For a current state $X_r$ at time $r$ for $r \in \ccint{s_i,t_i}$ and each active residual $E_r^{\ell j}$, the candidate time is obtained by solving $\smash{\int_u^r \rratemat{v}{X_r}{j;\tk{i}}[\param,\ell]\eqsp\rmd v=E_r^{\ell j}}$ for $u$ (line~\ref{line:erf-integrated-reverse-rate} of \Cref{alg:erf-sampler}). For MDM this is straightforward since when $X_r^\ell = \mask$,
\begin{equation}
    \int_u^r \rratemat{v}{z}{j;\tk{i}}[\param,\ell]\eqsp\rmd v
    =
        \denoiser{\tk{i}}{}{X_{\tk{i}}}[\param]^{\ell j}
        \log\left(\frac{1-\alpha_r}{1-\alpha_u}\right)
\end{equation}
and its candidate time follows by solving for $u$. Unmasked positions cannot jump so their candidate jump times are not computed. For UDM, however, the reverse rate integral is not explicitly invertible in $u$, and each candidate time must therefore be approximated by bisection. Since UDM can produce many jumps within a single interval, repeatedly solving these equations can be computationally costly. To control this cost, we split each interval into $k$ subintervals and allow at most one jump per position in each, advancing all residuals to the subinterval boundary after that jump; see \Cref{app:ctmc-partial-freezing}. To reduce \emph{both} the number of jumps and the cost of computing their times, we exploit the fact that different rate matrices can generate the same marginals $(\pdata{t}{}{})_{t\in[0,1]}$ \citep{campbell2024generativeflows,campbell2022continuous,gat2024discreteflowmatching} and use the net-flux reverse rates \citep{zhang2023discreteprobabilityflow}, defined for $x \neq y$ as 
\begin{equation}
    \label{eq:net-flux-reverse-rate}
    \rratemat{t}{\bx}{\by}[\mathrm{NF}]
    \!\eqdef\!
    \frac{\left[
        \pdata{t}{}{\bx}\rratemat{t}{\bx}{\by}
        -\pdata{t}{}{\by}\rratemat{t}{\by}{\bx}
    \right]_+}{\pdata{t}{}{\bx}} \!=\! -\sum_{\ell = 1}^\len \indic_{y^{-\ell} = x^{-\ell}} \frac{\dot\alpha_t}{\vocab \alpha_t} \bigg[\frac{\pdata{t}{x^{-\ell} _t}{y^\ell}}{\pdata{t}{x^{-\ell} _t}{x^\ell _t}} - 1 \bigg]_{+}
\end{equation}
where $a_+=\max\{a,0\}$ and $\rratemat{t}{}{}$ is the canonical UDM rate matrix. Surprisingly, the UDM net-flux forward and backward rates are tractable and their time integrals are also analytically invertible which eliminates the need for bisection search for the jump times and greatly increases computational speed, as we show in the experiments. The resulting updates are given in \Cref{app:practical-discretization}, with the net-flux construction in \Cref{app:net-flux}.

\paragraph{Parallel sampling.} \Cref{alg:erf-sampler} processes jumps sequentially across positions because a jump may change the reverse rates. However, for MDM, UDM, and Uniform Net Flux with leave-one-out freezing, the frozen reverse rate at position $\ell$, $\rratemat{t}{z}{y}[\ell]$, depends on the current state $z$ only through $z^\ell$; see \Cref{app:ctmc-partial-freezing,app:practical-discretization}. This means that changes at other positions than $\ell$ do not affect the reverse rate at position $\ell$ and the associated residuals. Hence, within $[\sk{i},\tk{i}]$, the ERF dynamics can be seen as $\len$ independent single token CTMCs and different positions can be processed in parallel, with jumps at each position processed sequentially; see \Cref{alg:parallel-erf}.

%% file: body/sec3.1v2.tex
\fixedspaceparagraph{Exponential Residual Flow.}
\label{sec:exponential-residual-flow-construction}

The exponential residual flow (ERF) eliminates this remaining randomness by keeping the evolution of the active residuals unchanged and replacing the resampling of the $\len$ passive residuals at each jump by deterministic evolution. In \Cref{alg:gillespie-first-reaction}, this amounts to retaining the active update in line~\ref{line:erf-active-residual-update} and replacing the sampling step in line~\ref{line:gillespie-passive-residual-refresh} by the deterministic update in line~\ref{line:erf-passive-residual-update}. We initialize $(X_1,E_1)\sim p_1\otimes\operatorname{Exp}(1)^{\otimes(\len\times\vocab)}$. The entire subsequent trajectory is deterministic given $(X_1,E_1)$.
More precisely, we keep track of the residuals $E_{t}^{\ell j}$ that correspond to the current tokens at each position, i.e. $j = x_t^{\ell}$, and we increase them at a certain rate, while the other exponentials are decreased at a different rate and trigger a jump when they reach zero. The rates are chosen to preserve the marginals of the CTMC.

We can deduce from the algorithm a process $(X_t,E_t)_{t\in\ccint{0,1}}$ defined over $\ccint{0,1}$. The algorithm defines the process at successive jump times $(\tau_k)_{k\in\geq 0}$. Then, for any $k$, on $\ocint{\tau_{k+1}, \tau_k}$, $X_t$ is kept constant equal to $X_{\tau_k}$ and $E_t$ solves the system of ODEs:
\begin{equation}
    \label{eq:exponential-residual-flow}
    \frac{\rmd}{\rmd t}E_t^{\ell j}
    =\begin{cases}
        \rratemat{t}{X_t}{j}[\ell], & j\neq X_t^\ell,\\
        -\lambda_t^\ell(X_t^\ell), & j=X_t^\ell.
    \end{cases}
\end{equation}
We call the process $(X_t,E_t)_{t\in\ccint{0,1}}$ the exponential residual flow (ERF) process. A simple trajectory with three successive jumps is illustrated in \Cref{fig:erf-illustration}.

We show in our next result that the resulting process $(X_t)_{t \geq 0}$ admits the same time marginals as the original CTMC. For completeness, a more formal construction of this process alongside Gillespie's first-reaction form is given in \Cref{sec:erf_sampler}.

\begin{mdframed}[style=propFrame,nobreak=true]
\begin{theorem}[Correctness of the exponential residual flow]
\label{thm:exponential-residual-flow}
Under standard assumptions stated in \Cref{sec:erf-transcription} the ERF process is almost surely well defined with finitely many jumps, and, for every $t\in[0,1]$,
    \begin{equation}
        \mathcal{L}(X_t,E_t)
        =
        \pdata{t}{}{}\otimes\operatorname{Exp}(1)^{\otimes (L\times K)}
        \eqsp.
    \end{equation}
\end{theorem}
\end{mdframed}

To the best of our knowledge, \Cref{thm:exponential-residual-flow} yields a novel deterministic procedure for sampling the marginals of a CTMC which is deterministic given the initialization, similarly to the probability flow ODE in the continuous case \citep{song2021ddim}.

We motivate the choice of the velocity \eqref{eq:exponential-residual-flow} in \Cref{sec:erf-motivation} by deriving a continuity equation on the extended state space from the Kolmogorov equation \eqref{eq:kolmogorov} satisfied by $(p_t)_{t\in\ccint{0,1}}$. Its vector field is precisely the one in \eqref{eq:exponential-residual-flow}. This argument, however, does not by itself prove \Cref{thm:exponential-residual-flow}. A complete proof instead characterizes by induction the density of the process at successive jump times and decomposes the path space according to the number of jumps. We give this proof in \Cref{sec:erf-transcription}.

%% file: body/distillation.tex
\section{Application to distillation}
\label{sec:distillation}

As in the continuous diffusion case \citep{salimans2022progressive,song2023consistency}, one application of deterministic sampling is to distill a teacher diffusion model into a student model that performs better with few sampling steps. The absence of a direct analogue of the probability flow ODE in the discrete setting has often been used as an argument against trajectory-based distillation \citep{sahoo2025diffusion,deschenaux2024autoregressionfastllmsselfdistillation,zhang2023discreteprobabilityflow}. Our exponential residual flow supplies deterministic transport between augmented states $(X_t,E_t)$ and $(X_s,E_s)$, allowing student predictions to be matched along a teacher trajectory.

\paragraph{Consistency distillation.}

We rely on consistency distillation (CD) following \citet{song2023consistency}. From an augmented state, the student predicts the clean discrete state at time $0$ reached by the teacher trajectory, it does not need to predict the endpoint exponential residuals. We train its predictions from two points on the same trajectory to agree. Writing $\Psi^\theta_{s\tbar t}$ for the ERF transport of \Cref{alg:erf-sampler}, these points satisfy $(X_s,E_s)=\Psi^\theta_{s\tbar t}(X_t,E_t)$ for $s<t$.
In a previous approach, \citet{sahoo2025diffusion} couple $X_s$ and $X_t$ by using the same Gaussian noise in an underlying process. Their construction is deterministic conditional on this same noise, but the transition remains stochastic when conditioned only on $X_t$, which contradicts what is required in CD (see \Cref{sec:dcd-implementation}). In contrast, our approach deterministically produces $(X_s,E_s)$ from $(X_t,E_t)$.

Since the ERF output at time $0$ is uniquely determined by the augmented state at time $t$, a naive approach is to train a student to approximate this discrete-valued map directly. To make the learning problem more tractable, the output is relaxed at each position to an element of the probability simplex. More precisely, the student as a factorized distribution over discrete states:
$$
\txts \pdata{0\tbar t}{\bx_t, e_t}{\bx_0}[\phi] \eqdef \prod_{\ell = 1}^\len \pdata{0\tbar t}{\bx_t, e_t}{\bx^\ell _0}[\phi, \ell] \eqsp.
$$
For one current parameter $\phi$ and target parameter $\bar\phi$ which is held fixed during optimization, we consider the loss function: 
\begin{equation}
  \label{eq:consistency_distillation_loss}
  \mathcal L_{\mathrm{CD}}(\phi;\bar\phi) = \pE \big[\kldivergence{\pdata{0\tbar s}{X_s, E_s}{}[\bar\phi]}{\pdata{0\tbar t}{X_t, E_t}{}[\phi]} \big] \eqsp,
\end{equation}
where $(X_t,E_t)\sim\pdata{t}{}{}\otimes\operatorname{Exp}(1)^{\otimes\len\times\vocab}$, $(X_s,E_s)=\Psi^\theta_{s\tbar t}(X_t,E_t)$, $t$ is sampled from $\unif(\delta,1)$, and $s=t-\delta$, where $\delta$ is a hyperparameter. The target parameter $\bar\phi$ is either updated by EMA after each step or frozen for a training round and refreshed between rounds. 
As noted in \citet{song2023consistency}, for the minimizer to be exact, we need to impose the boundary condition $\pdata{0\tbar 0}{\bx_0,\be_0}{}[\phi]=\updelta_{\bx_0}$ and give a non-zero probability to $s=0$ so that at optimality, the student assigns all its mass to the clean endpoint of the ERF trajectory. Marginalizing over the augmented states $(X_t,E_t)$ at time $t$, these endpoints have law $\pdata{0}{}{}$.
In practice, imposing the boundary condition did not improve the performance so we don't use it, this condition's role is to remove trivial constant solutions from the minimizer, but initializing the network with the teacher appears to be enough to reach non-trivial solutions. We also consider a biased ERF sampler to obtain $(X_s,E_s)$ from $(X_t,E_t)$. Instead of using the pretrained teacher denoiser, we use the frozen student when computing the ERF rates. Finally, for UDM, we convert the student output to a denoiser type prediction using \eqref{eq:udm-loo-denoiser-conversion} before evaluating the loss. This is because the teacher was trained to model a leave-one-out denoiser while the quantity we want the student to learn is closer to a denoiser type prediction so this conversion removes a slight mismatch at least towards the start of training. The fixed-teacher and joint-refresh procedures are given in \Cref{alg:erf-consistency-distillation,alg:erf-consistency-distillation-joint-refresh}.

\paragraph{Sampling.} We consider the \emph{refresh} and \emph{bridge} transitions for the discrete state which are respectively
$\smash{\sum_{x_0}\fw{s\tbar0}{x_0}{x_s}\, \pdata{0\tbar t}{x_t,e_t}{x_0}[\phi]}$
and
$\smash{\sum_{x_0}\fw{s\tbar0,t}{x_0,x_t}{x_s}\, \pdata{0\tbar t}{x_t,e_t}{x_0}[\phi]}$,
with fresh exponential variables drawn independently after each transition. Refresh is exact when the teacher transports the prescribed augmented marginals and the student learns its endpoint map. Bridge is not exact in general, but performs well empirically. Both procedures are given in \Cref{alg:erf-consistency-sampling}. 

%% file: body/related_work.tex
\section{Related Work}

\paragraph{CTMC sampling and discretization.}
\citet{campbell2022continuous} formulated discrete diffusion in continuous time, connecting their sampling to exact CTMC simulation \citep{gillespie1977exact} and $\tau$-leaping \citep{gillespie2001approximate}. The modified next-reaction method \citep{anderson2007modified} preserves integrated-rate clocks across jumps but resamples the firing channel's exponential threshold. ERF instead becomes deterministic after a single random initialization, recycling residuals while preserving the exact one-time CTMC marginals, though generally with a different trajectory law. In practice, we freeze only the leave-one-out denoiser over each step while retaining the known time and state dependence of the rates, extending Euler/$\tau$-leaping-style discretizations.

\paragraph{Deterministic sampling of discrete diffusion models.}
\citet{zhang2023discreteprobabilityflow} introduce net-flux rates \eqref{eq:net-flux-reverse-rate} to reduce jumps in a discrete analogue of the probability-flow ODE; see \Cref{app:net-flux}, but the resulting process remains stochastic. \citet{eyring2026ddno} obtain deterministic trajectories by reusing Gumbel noise across discretization steps, potentially drifting from the target distribution. ERF instead gives deterministic transport after initialization while exactly preserving one-time marginals.

\paragraph{Distillation of discrete diffusion models.}
Distillation reduces sampling steps through distribution- or trajectory-based objectives. D-MMD \citep{hoogeboom2026beyond} uses discrete moment matching, while IDLM \citep{li2026idlm} uses a path space KL objective related to distribution matching \citep{yin2023one}. Both train auxiliary denoisers on student samples. Trajectory-based methods build on progressive and consistency distillation \citep{salimans2022progressive,song2023consistency}: SDTT \citep{deschenaux2024autoregressionfastllmsselfdistillation} aggregates teacher probabilities across stochastic unmasking steps, while DCD \citep{sahoo2025diffusion} matches predictions along projected Gaussian trajectories with shared clean states and noise. These trajectories remain stochastic, leading to heuristic or biased objectives. Conditioning on exponential residuals instead makes the teacher trajectory deterministic given $(X_t,E_t)$, enabling consistency distillation in the discrete setting.

%% file: body/experiments.tex
\section{Experiments}
\label{sec:experiments}
\begin{figure}[t]
    \centering
    \includegraphics[width=0.9\linewidth]{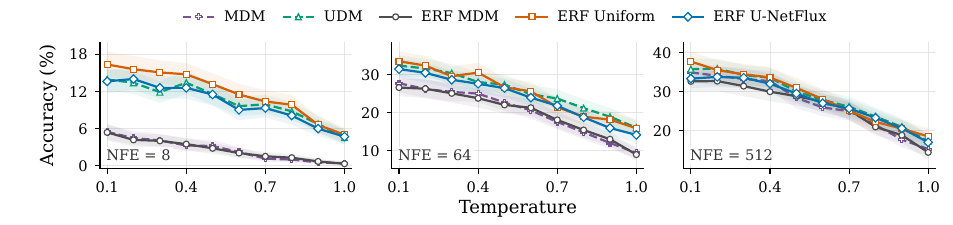}
    \vspace{-.01cm}
    \smallskip
    \includegraphics[width=0.9\linewidth]{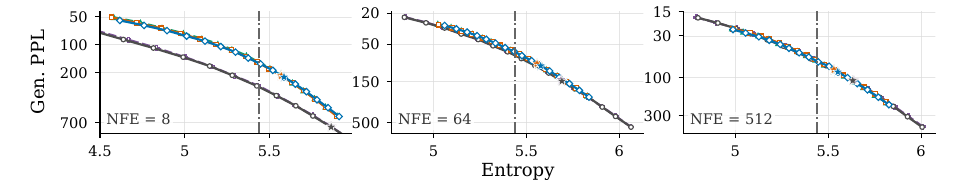}
    \captionsetup{font=small}
    \caption{Sampling quality at matched NFE on TinyGSM (top) and OpenWebText (bottom). Dashed curves denote ancestral samplers; solid curves denote ERF. TinyGSM bands are $95\%$ bootstrap intervals over questions. OWT curves trace the generative perplexity versus 1-gram entropy frontier by sweeping the temperature in $[0.8,1.1]$; $\star$ refers to temperature $1$, and the vertical dash-dotted line marks the OWT dataset entropy.}
    \label{fig:erf-sampler-comparison}
    \label{fig:sampler-tinygsm}
    \label{fig:sampler-owt}
\end{figure}
Our experiments first study the empirical consequences of ERF’s new joint law, showing that the resulting sampler achieves generation quality comparable to standard ancestral samplers at matched NFE. We then show that ERF consistency distillation (ERC) enables competitive generation with fewer NFEs than the teacher model.
 
 \paragraph{General setting.} We consider masked and uniform diffusion models trained on OpenWebText (OWT; \citealp{Gokaslan2019OpenWeb}) and TinyGSM \citep{liu2023tinygsm}, using DDiT backbones with $170$M and $168$M parameters, respectively. OWT generation is unconditional with sequences of length $1{,}024$; we measure GPT-2 Large generative perplexity \citep{radford2019language} together with 1-gram entropy. TinyGSM models generate Python solutions on a $512$-token canvas with fixed question tokens, and we evaluate answer accuracy on GSM8K \citep{cobbe2021trainingverifiers}, using all $1{,}319$ test questions for the sampling comparison and the same first $1{,}024$ questions for the main distillation comparison. For sampling, we also evaluate the released instruction-tuned DiffusionGemma model, with $26$B total and approximately $4$B active parameters \citep{gemma4diffusion2026blog}, on MMLU-Pro \citep{wang2024mmlu}.

\subsection{Evaluation of the ERF sampler}
\label{sec:erf-sampler-evaluation}
\begin{figure}[t]
    \centering
    \includegraphics[width=0.9\linewidth]{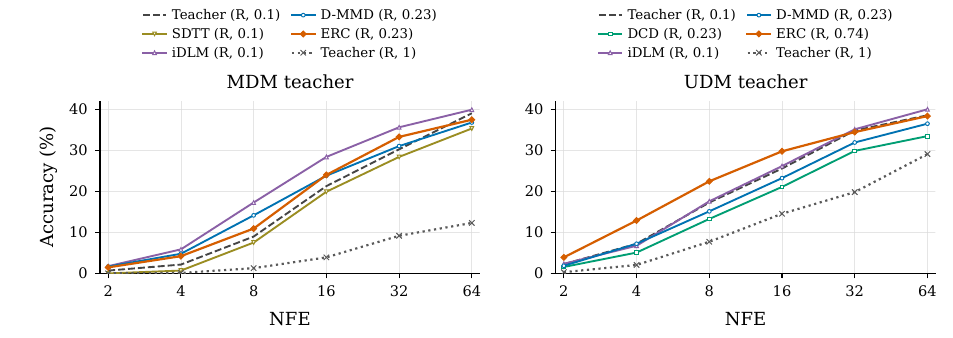}
    \captionsetup{font=small}
    \caption{TinyGSM distillation at $2$--$64$ NFEs. Each curve fixes its sampler and temperature to maximize mean accuracy across NFEs on the same $1{,}024$ questions. ERC selects $\delta=1/64$ for MDM and $1/8$ for UDM. $R$ refers to refresh sampling.}
    \label{fig:distillation-tinygsm}
\end{figure}
\paragraph{Sampling quality.}
Within each diffusion family, we compare ERF and ancestral sampling using the same pretrained checkpoint at matched NFE budgets. We use a uniform time grid and compute the ERF Uniform jump times in line~\ref{line:erf-integrated-reverse-rate} of \Cref{alg:erf-sampler} using $40$ bisection iterations per solve. Further evaluations appear in \Cref{sec:sampling-sweeps-appendix}. The results are given in \Cref{fig:erf-sampler-comparison}. ERF achieves sampling quality comparable to the ancestral MDM and UDM samplers across the evaluated datasets and NFE budgets. On TinyGSM, the samplers exhibit similar accuracy and temperature dependence; on OWT, their entropy--perplexity curves across temperatures largely overlap within each diffusion family.
\par\noindent
\begin{minipage}{\linewidth}
\setlength{\intextsep}{6pt}
\begin{wraptable}{r}{0.46\linewidth}
    \vspace{-.4cm}
    \captionsetup{font=scriptsize,skip=2pt}
    \caption{DiffusionGemma on $140$ MMLU-Pro questions and three paired seeds. Brackets give $95\%$ question-cluster bootstrap intervals.}
    \label{tab:diffusiongemma-mmlu-pro}
    \centering
    \scriptsize
    \setlength{\tabcolsep}{2pt}
    \begin{tabular*}{\linewidth}{@{\extracolsep{\fill}}lcc@{}}
        \toprule
        Sampler & Accuracy (\%) & s/question \\
        \midrule
        Ancestral & $57.4\ [50.5,64.3]$ & $59.4$ \\
        ERF Uniform & $56.7\ [49.8,63.6]$ & $61.1$ \\
        ERF U-NetFlux & $57.1\ [50.2,64.0]$ & $60.4$ \\
        \bottomrule
    \end{tabular*}
\end{wraptable}
The DiffusionGemma results support the same conclusion at a larger model scale (\Cref{tab:diffusiongemma-mmlu-pro}). At $48$ NFEs and temperature $1$, ERF Uniform and U-NetFlux are within $0.7$ percentage points of the UDM ancestral sampler on MMLU-Pro questions. See \Cref{app:diffusiongemma} for experiment details. 
\end{minipage}
\par\noindent
\begin{minipage}[t]{0.65\linewidth}
    \vspace{0pt}
    \raggedright
    \paragraph{ERF Uniform vs U-NetFlux.}
    The U-NetFlux version is the more efficient ERF sampler for uniform diffusion because we can solve explicitly for the crossing time in line~\ref{line:erf-integrated-reverse-rate} without bisection; see \Cref{app:net-flux}. Furthermore, it makes substantially fewer jumps and thus requires fewer solves. At $8$ NFEs, trajectories average approximately one jump per generated position, compared with seven for Uniform, yielding speedups of $2.83\times$ on TinyGSM and $2.00\times$ on OWT (\Cref{tab:sampler-efficiency}). This gain comes while maintaining similar sampling quality in \Cref{fig:erf-sampler-comparison}. We also evaluate ERF Uniform with a cap on jumps within the interval $[\sk{i}, \tk{i}]$; \Cref{fig:erf-uniform-boundedness,fig:uniform-substeps-owt-full} show its effect on sampling quality.
\end{minipage}\hfill
\begin{minipage}[t]{0.32\linewidth}
    \vspace{0pt}
    \captionsetup{type=table,font=scriptsize,skip=2pt}
    \caption{Runtime and jump averages at $8$ NFEs over three batches; times are seconds per sequence and jumps are per generated position.}
    \label{tab:sampler-efficiency}
    \centering
    \scriptsize
    \setlength{\tabcolsep}{2pt}
    \begin{tabular*}{\linewidth}{@{\extracolsep{\fill}}lrr@{}}
        \toprule
        Sampler & Time (s) & Jumps \\
        \midrule
        \multicolumn{3}{l}{TinyGSM} \\
        Uniform & $0.368$ & $6.90$ \\
        \textbf{U-NetFlux} & $\mathbf{0.130}$ & $\mathbf{1.02}$ \\
        \midrule
        \multicolumn{3}{l}{OWT} \\
        Uniform & $0.647$ & $7.11$ \\
        \textbf{U-NetFlux} & $\mathbf{0.323}$ & $\mathbf{1.04}$ \\
        \bottomrule
    \end{tabular*}
\end{minipage}\par
\subsection{Distillation}
\label{sec:distillation-experiments}

\paragraph{Setting and baselines.}
We compare ERC with SDTT \citep{deschenaux2024autoregressionfastllmsselfdistillation} for masked diffusion, DCD \citep{sahoo2025diffusion} for uniform diffusion, and D-MMD \citep{hoogeboom2026beyond} and iDLM \citep{li2026idlm} for both families. All results presented are from our own re-implementations and training runs, with all the students initialized from the corresponding pretrained teacher. Within each dataset and diffusion family, the methods share the \emph{same} teacher checkpoint. For both MDM and UDM, the teacher models are trained using the standard ELBOs; \citet[Equation 11]{sahoo2024simple} and \citet[Equation 19]{schiff2024simple}. In particular, for UDM our pre-trained model is a LOO denoiser. All the implementation details are given in \Cref{sec:dcd-implementation,sec:idlm-implementation}.

\begin{figure}[t]
    \centering
    \includegraphics[width=0.9\linewidth]{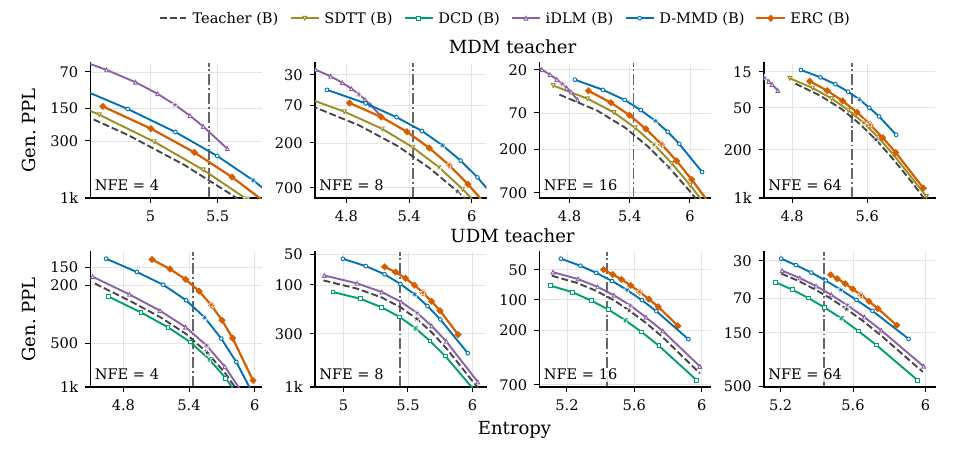}
    \captionsetup{font=small}
    \caption{OWT distillation with bridge sampling at $4$, $8$, $16$, and $64$ NFEs, with students trained for $100$ GPU-hours. Each point uses $2{,}048$ samples, with temperature applied to denoiser probabilities over $[0.8,1.2]$; $\star$ marks temperature $1$, and the vertical dash-dotted line marks the OWT dataset entropy. B refers to bridge sampling.}
    \label{fig:distillation-owt}
\end{figure}

\paragraph{Training.}
ERC conditions the student on Gumbel variables $-\log E_t$ through a learned projection; UDM ERC uses net-flux ERF rates.
On both datasets we run all the methods with the same GPU-hour budget (100 for OWT and 50 for TinyGSM) and batch size $128$. ERC uses four equal training rounds and the joint target-and-teacher refresh in \Cref{alg:erf-consistency-distillation-joint-refresh}. We sweep $\delta\in\{1/8,1/16,1/32,1/64\}$ and select one value per dataset and diffusion family using the evaluation criteria below. ERC uses forward KL on OWT and reverse KL on TinyGSM. For UDM, we apply the LOO-to-denoiser conversion formula to the student output before computing the loss \eqref{eq:udm-loo-denoiser-conversion}. This gave improved results on all our experiments. \Cref{sec:distillation-evaluation-details} gives the training and evaluation details. We disable the top-$p$ truncation used by D-MMD \citep{hoogeboom2026beyond} and fix the shared teacher’s temperature to $1$, so that all methods distill the same teacher distribution.

\paragraph{Sampling protocol.}
We evaluate all methods and the teacher with both \emph{refresh} and \emph{bridge} sampling at matched NFE, sweeping a common temperature grid to account for differences in logit sharpness across objectives. On TinyGSM, we select one sampler, temperature placement and temperature per method, together with one ERC gap $\delta$, and hold these choices fixed along each curve. On OWT, the main figure compares generative perplexity--entropy curves \citep{pynadath2026generativefrontiersevaluationmatters}. In the TinyGSM legend, B and R denote bridge and refresh sampling; the number gives the selected inference temperature. The OWT legend marks every method with B to indicate bridge sampling.

\paragraph{Results.} On TinyGSM, UDM ERC doubles the accuracy of the strongest competing baseline at $2$ and $4$ NFEs and also records higher accuracy at $8$ and $16$; by $32$ NFEs its accuracy is close to the teacher's. In this evaluation, it is the only method with accuracy above the teacher at some NFE budgets when fully sweeping temperature and samplers (even for the teacher). For MDM, ERC approximately matches the second-best baseline, see \Cref{fig:distillation-tinygsm}. On OWT we see the same trend, UDM ERC has lower perplexity at matched entropy than the other baselines on the full frontier, while MDM ERC approximately matches the second-best baseline, see \Cref{fig:distillation-owt}. On Sudoku, ERC gives modest but consistent gains over the other methods in many evaluated settings. The complete sampler comparisons and Sudoku results appear in \Cref{sec:distillation-sampler-ablations,sec:sudoku-distillation-results}. Across these comparisons, the largest ERC gains occur with uniform diffusion, while masked-diffusion results remain competetitive with the considered baselines.

\paragraph{Ablations.} We vary how the target and ERF transport (providing $X_s$ starting from $X_t$ in \eqref{eq:consistency_distillation_loss}) are updated, the training gap $\delta$, the inference sampler, and whether ERC receives Gumbel inputs. For UDM, we also compare two placements of sampling temperature. The protocols and complete grids appear in \Cref{sec:distillation-policy-ablations,sec:distillation-gap-ablations,sec:distillation-sampler-ablations,sec:distillation-gumbel-ablations}.

%% file: body/conclusion.tex
\section{Conclusion}

In this work, we show that the prescribed marginals of a discrete diffusion can be transported by deterministic dynamics once the token state is augmented with continuous residuals. This construction generalizes Gillespie's algorithm by recycling the exponential variables drawn at every jump according to an appropriate ODE. This gives a pathwise target for consistency distillation. In practice, leave-one-out freezing retains the analytically known time dependence of the rates and is exact for factorized data. Across the evaluated tasks, ERF achieves sampling quality comparable to ancestral updates at matched model evaluations, and its consistency objective yields its clearest few-step gains for uniform diffusion.

%% file: appendix/practical_discretization.tex
\section{Rates and jump times used in the discretization}
\label{app:practical-discretization}

Fix $[s,t]=[\sk{i},\tk{i}]$ and condition on $X_{\tk{i}}=x$. The leave-one-out denoiser $\denoiser{\tk{i}}{}{x}[\param]$ is cached throughout the step. Let $z$ be the current state at time $r\in[s,t]$, and let $u\in[s,r]$ be a candidate jump time. An active residual $E_r^{\ell j}$ crosses zero when the integrated reverse rate from $u$ to $r$ equals $E_r^{\ell j}$; passive residuals increase by the integrated forward exit rate. More general freezing schemes and the derivations are given in \Cref{app:ctmc-partial-freezing,app:net-flux}.

For MDM, only $z^\ell=\mask$ can jump, and then only to $j\neq\mask$. The integrals needed for active and passive residuals are
\begin{align*}
    \int_u^r\rratemat{v}{z}{j;\tk{i}}[\param,\ell]\,\rmd v
    &=\denoiser{\tk{i}}{}{x}[\param]^{\ell j}\log\!\left(\frac{1-\alpha_r}{1-\alpha_u}\right)
    &&(z^\ell=\mask),\\
    \int_u^r\lambda_v^\ell(z^\ell)\,\rmd v
    &=\indic_{z^\ell\neq\mask}\log\!\left(\frac{\alpha_u}{\alpha_r}\right).
\end{align*}
If $\denoiser{\tk{i}}{}{x}[\param]^{\ell j}>0$ and $E_r^{\ell j}\leq\denoiser{\tk{i}}{}{x}[\param]^{\ell j}\log((1-\alpha_r)/(1-\alpha_s))$, the candidate time for a masked position is
\begin{align*}
    T^{\ell j}
    &=\alpha^{-1}\!\left(1-(1-\alpha_r)\exp\!\left\{-\frac{E_r^{\ell j}}{\denoiser{\tk{i}}{}{x}[\param]^{\ell j}}\right\}\right),
\end{align*}
and $T^{\ell j}=-\infty$ otherwise. In particular, no further active jump is possible at an unmasked position. Net flux does not change the MDM rates.

For UDM with net flux and $j\neq z^\ell$, the active and passive integrals are
\begin{align*}
    \int_u^r\rratemat{v}{z}{j;\tk{i}}[\mathrm{NF},\param,\ell]\,\rmd v
    &=\frac{[\denoiser{\tk{i}}{}{x}[\param]^{\ell j}-\denoiser{\tk{i}}{}{x}[\param]^{\ell z^\ell}]_+}
            {\vocab\denoiser{\tk{i}}{}{x}[\param]^{\ell z^\ell}-1}\\
    &\quad\times\log\!\left(\frac{1+(\vocab\denoiser{\tk{i}}{}{x}[\param]^{\ell z^\ell}-1)\alpha_u}
                       {1+(\vocab\denoiser{\tk{i}}{}{x}[\param]^{\ell z^\ell}-1)\alpha_r}\right),\\
    \int_u^r b_v^{\mathrm{NF},\ell}(z)\,\rmd v
    &=\frac{\sum_{j\in\msv\setminus\{z^\ell\}}[\denoiser{\tk{i}}{}{x}[\param]^{\ell z^\ell}-\denoiser{\tk{i}}{}{x}[\param]^{\ell j}]_+}
            {\vocab\denoiser{\tk{i}}{}{x}[\param]^{\ell z^\ell}-1}\\
    &\quad\times\log\!\left(\frac{1+(\vocab\denoiser{\tk{i}}{}{x}[\param]^{\ell z^\ell}-1)\alpha_u}
                       {1+(\vocab\denoiser{\tk{i}}{}{x}[\param]^{\ell z^\ell}-1)\alpha_r}\right).
\end{align*}
Here $b_v^{\mathrm{NF},\ell}(z)$ is the forward net-flux exit rate at position $\ell$. When $\vocab\denoiser{\tk{i}}{}{x}[\param]^{\ell z^\ell}=1$, each logarithmic quotient divided by $\vocab\denoiser{\tk{i}}{}{x}[\param]^{\ell z^\ell}-1$ is understood as $\alpha_u-\alpha_r$. If $\denoiser{\tk{i}}{}{x}[\param]^{\ell j}>\denoiser{\tk{i}}{}{x}[\param]^{\ell z^\ell}$ and the active integral at $u=s$ is at least $E_r^{\ell j}$, its candidate time is
\begin{align*}
    T^{\ell j}
    &=\alpha^{-1}\!\left(
        \frac{\bigl(1+(\vocab\denoiser{\tk{i}}{}{x}[\param]^{\ell z^\ell}-1)\alpha_r\bigr)
        \exp\!\left\{\frac{(\vocab\denoiser{\tk{i}}{}{x}[\param]^{\ell z^\ell}-1)E_r^{\ell j}}{\denoiser{\tk{i}}{}{x}[\param]^{\ell j}-\denoiser{\tk{i}}{}{x}[\param]^{\ell z^\ell}}\right\}-1}
        {\vocab\denoiser{\tk{i}}{}{x}[\param]^{\ell z^\ell}-1}\right).
\end{align*}
For $\vocab\denoiser{\tk{i}}{}{x}[\param]^{\ell z^\ell}=1$, this becomes $T^{\ell j}=\alpha^{-1}(\alpha_r+E_r^{\ell j}/(\denoiser{\tk{i}}{}{x}[\param]^{\ell j}-\denoiser{\tk{i}}{}{x}[\param]^{\ell z^\ell}))$. When the stated conditions for a candidate jump fail, set $T^{\ell j}=-\infty$. The largest candidate time gives the next jump. After updating $z$, the residual integrals are recomputed with the same cached denoiser.

\subsection{Parallel ERF sampling}
\label{app:parallel-erf}

\Cref{alg:parallel-erf} applies the leave-one-out-frozen rates above independently at each position on each grid interval. In the algorithm, $\rratemat{r}{X_{\sk{i}}}{j;\tk{i}}[\ell]$ denotes the chosen frozen reverse rate, and $b_r^\ell(X_{\sk{i}})$ denotes its forward exit rate: $\lambda_r^\ell(X_{\sk{i}}^\ell)$ for MDM and UDM, or $b_r^{\mathrm{NF},\ell}(X_{\sk{i}})$ for Uniform Net Flux. We take the supremum of an empty set to be $-\infty$. The jump-time variables inside the parallel loop are local to each position; the cached denoiser is shared across positions.

\begin{algorithm}[t]
    \caption{Parallel ERF sampling with leave-one-out-frozen rates}
    \label{alg:parallel-erf}
    \begin{algorithmic}[1]
        \Require Grid $0=\tk{0}<\cdots<\tk{\last}=1$ and leave-one-out-frozen reverse and forward rates
        \State Draw $X_1\sim\pdata{1}{}{}$ and, independently, $E_1^{\ell j}\simiid\operatorname{Exp}(1)$ for all $(\ell,j)$
        \For{$i=\last,\ldots,1$}
            \State Cache $\denoiser{\tk{i}}{}{X_{\tk{i}}}[\param]$ and define the frozen rates on $[\sk{i},\tk{i}]$
            \State $(X_{\sk{i}},E_{\sk{i}})\gets(X_{\tk{i}},E_{\tk{i}})$
            \ForAll{$\ell\in\intset{1}{\len}$ \textbf{in parallel}}
                \State $\tau^\star\gets\tk{i}$
                \While{$\tau^\star>\sk{i}$}
                    \State $\tau^{\ell j}\gets\sup\defEns{u\in[\sk{i},\tau^\star]:\int_u^{\tau^\star}\rratemat{r}{X_{\sk{i}}}{j;\tk{i}}[\ell]\,\rmd r\geq E_{\sk{i}}^{\ell j}}$ for $j\neq X_{\sk{i}}^\ell$
                    \State $j^\star\gets\argmax_{j\neq X_{\sk{i}}^\ell}\tau^{\ell j}$
                    \State $\tau^\star_{\mathrm{new}}\gets\max\{\sk{i},\tau^{\ell j^\star}\}$
                    \State $E_{\sk{i}}^{\ell j}\gets E_{\sk{i}}^{\ell j}-\int_{\tau^\star_{\mathrm{new}}}^{\tau^\star}\rratemat{r}{X_{\sk{i}}}{j;\tk{i}}[\ell]\,\rmd r$ for $j\neq X_{\sk{i}}^\ell$
                    \State $E_{\sk{i}}^{\ell j}\gets E_{\sk{i}}^{\ell j}+\int_{\tau^\star_{\mathrm{new}}}^{\tau^\star}b_r^\ell(X_{\sk{i}})\,\rmd r$ for $j=X_{\sk{i}}^\ell$
                    \If{$\tau^{\ell j^\star}\geq\sk{i}$}
                        \State $X_{\sk{i}}^\ell\gets j^\star$
                    \EndIf
                    \State $\tau^\star\gets\tau^\star_{\mathrm{new}}$
                \EndWhile
            \EndFor
        \EndFor
        \State \Return $X_0,E_0$
    \end{algorithmic}
\end{algorithm}

%% file: appendix/bridge_simplex.tex
\section{Bridge and simplex extension}
\label{sec:bridge-simplex}
For $0\leq s<t\leq1$ with $\alpha_s>0$, conditional independence across positions gives $\smash{\fw{s\tbar0,t}{x_0,x_t}{x_s}=\prod_{\ell=1}^{\len}\fw{s\tbar0,t}{x_0^\ell,x_t^\ell}{x_s^\ell}[\ell]}$. For $x_0,x_t\in\msx$ and a position $\ell$ such that $\fw{t\tbar0}{x_0^\ell}{x_t^\ell}[\ell]>0$, Bayes' rule yields
\begin{equation}
    \label{eq:bridge-prob}
    \fw{s\tbar0,t}{x_0^\ell,x_t^\ell}{\cdot}[\ell]
    =
    \categorical\!\left(\cdot;
    \frac{
        \big[\alpha_{t\tbar s}\tx_t^{\ell}+(1-\alpha_{t\tbar s})\dotp{\tx_t^{\ell}}{\bpi^\ell}\one_{\vocab}\big]
        \odot\big[\alpha_s \tx^{\ell}_0+(1-\alpha_s)\bpi^\ell\big]
    }{
        \alpha_t\dotp{\tx^{\ell}_t}{\tx^{\ell}_0}+(1-\alpha_t)\dotp{\tx^{\ell}_t}{\bpi^\ell}
    }\right)\eqsp,
\end{equation}
where $\tx_0^\ell=\oh{x}_0^\ell$ and $\tx_t^\ell=\oh{x}_t^\ell$.
For $\nu\in\Simplex{\vocab}$, we extend this expression by replacing $x_0^\ell$ with $\nu$, equivalently setting $\fw{s\tbar0,t}{\nu,x_t^\ell}{x_s^\ell}[\ell]=\fw{t\tbar s}{x_s^\ell}{x_t^\ell}[\ell]\fw{s\tbar0}{\nu}{x_s^\ell}[\ell]/\fw{t\tbar0}{\nu}{x_t^\ell}[\ell]$, where $\fw{r\tbar0}{\nu}{\cdot}[\ell]=\categorical(\cdot;\alpha_r\nu+(1-\alpha_r)\bpi^\ell)$. The numerator in \eqref{eq:bridge-prob} has nonnegative entries whose sum is the denominator, since $\alpha_{t\tbar s}\alpha_s=\alpha_t$. Hence the extension is a probability vector whenever the denominator is positive. Completing zero-denominator rows by any probability vector and taking the positionwise product yields a stochastic reverse transition matrix on $\msx$; for masked diffusion, we use $\categorical(\cdot;\oh{x}_t^\ell)$ at unmasked positions.
Following \citet[Proposition~1]{gourevitch2026udm}, define $\loodenoiser{t}{}{x_t}^\ell\eqdef\pE[\oh{X}_0^\ell\mid X_t^{-\ell}=x_t^{-\ell}]$. Conditioning first on $X_t^{-\ell}$ and then on $X_t^\ell$ gives, whenever $\pdata{t}{}{x_t}>0$, $\pdata{s\tbar t}{x_t}{x_s^\ell}[\ell]=\fw{s\tbar0,t}{\loodenoiser{t}{}{x_t}^\ell,x_t^\ell}{x_s^\ell}[\ell]$. Thus substituting the leave-one-out denoiser into the simplex extension recovers the exact positionwise reverse conditionals, and their product gives the factorized approximation used in \eqref{eq:plugin-reverse}.

%% file: appendix/ctmc_integrators.tex
\section{CTMC integration schemes}
\label{app:ctmc-integrators}

\subsection{Formal construction of the exact simulation algorithms}
\label{app:ctmc-gillespie}

\subsubsection{Gillespie's first reaction form}
The Gillespie's first reaction form \citep{gillespie2001approximate}, can be used to simulate $(X_u)_{u\in\ccint{s,t}}$ backward in time from $X_t\sim p_t$ using the reverse rate matrix $(\rratemat{u}{}{})_{u\in\ccint{s,t}}$. We define sequentially the jump times $(T_k^\star)_{k\le N}$ and states $(X_{T_k^\star})_{k\le N}$ up to a maximum (random) number of jumps $N_{\infty}$ in the following way. 

\textbf{Initialization.}
Set $T_0^\star=t$, $N_{\infty} = \infty$ and draw independently $X_{T_0^{\star}} \sim p_t$, and for every $(\ell,j) \in \intset{1}{\len}\times\intset{1}{\vocab}$, $E_{T^{\star}_0}^{\ell j}\sim\operatorname{Exp}(1)$. 

\textbf{Induction.} At stage $k\geq0$, given the current jump time $T_k^{\star}$  state $X_{T_k^{\star}}$ and exponential clocks $E_{T_k^{\star}}^{\ell j}$  for every $(\ell,j) \in \intset{1}{\len}\times\intset{1}{\vocab}$, define
\begin{equation}
    \label{eq:gillespie-first-reaction}
    T_{k+1}^{\ell j}
    =
    \sup\defEns{u\in[s,T_k^\star]:\ \int_u^{T_k^\star}\rratemat{r}{X_{T_k^\star}}{j}[\ell]\eqsp\rmd r\geq E_{T_k^{\star}}^{\ell j}}
    \eqsp,
    \qquad
    (\ell_{k+1}^\star,j_{k+1}^\star)
    \in
    \argmax_{(\ell,j)} T_{k+1}^{\ell j}
    \eqsp.
\end{equation}
We use here the convention $\sup \emptyset = -\infty$. 
\begin{enumerate}[wide,label=$\bullet$]
    \item If $T_{k+1}^{\ell_{k+1}^\star j_{k+1}^\star}=-\infty$, the procedure stops, so we set $N_{\infty}=k$ and $X_s=X_{T_k^\star}$.
    \item Otherwise, the $(k+1)$-st jump occurs at $T_{k+1}^\star=T_{k+1}^{\ell_{k+1}^\star j_{k+1}^\star}$. We set $X_{T_{k+1}^\star}^{\ell_{k+1}^\star}=j_{k+1}^\star$ and leave all other positions unchanged. For every $(\ell,j)$ such that $X_{T_k^\star}^{\ell}\neq j$, set
    \[
        E_{T_{k+1}^\star}^{\ell j}
        =E_{T_k^\star}^{\ell j}-\int_{T_{k+1}^\star}^{T_k^\star}\rratemat{r}{X_{T_k^\star}}{j}[\ell]\,\rmd r\geq 0.
    \]
    Finally, draw $E_{T_{k+1}^\star}^{\ell_{k+1}^\star j_{k+1}^\star}\sim\operatorname{Exp}(1)$ independently of the past.
\end{enumerate}

  The recursion continues until no candidate event occurs before $s$. Under appropriate conditions, (see the discussion in \cite{feinberg2013}), $N$ is almost surely finite and the algorithm produces a trajectory $(X_u)_{u\in\ccint{s,t}}$ whose law is that of the reverse CTMC conditioned on $X_t=x$.

\subsubsection{The ERF sampler}
\label{sec:erf_sampler}
The ERF augments $X_u$ with one nonnegative residual $E_u^{\ell j}$ for each position $\ell$ and token $j$. We construct $(X_u,E_u)_{u\in\ccint{s,t}}$ as follows.

\textbf{Initialization.} Set $T_0^\star=t$, draw $X_{T_0^\star}\sim p_t$, and draw $E_{T_0^\star}^{\ell j}\simiid\operatorname{Exp}(1)$ for all $(\ell,j)\in\intset{1}{\len}\times\intset{1}{\vocab}$, independently of $X_{T_0^\star}$.

\textbf{Induction.} At stage $k\geq0$, given $(T_k^\star,X_{T_k^\star},E_{T_k^\star})$, write $z=X_{T_k^\star}$. For each $j\neq z^\ell$, define the candidate time
\begin{equation}
    T_{k+1}^{\ell j}
    =\sup\defEns{u\in[s,T_k^\star]:\ \int_u^{T_k^\star}\rratemat{r}{z}{j}[\ell]\eqsp\rmd r\geq E_{T_k^\star}^{\ell j}}
    \eqsp,
    \qquad
    (\ell_{k+1}^\star,j_{k+1}^\star)\in\argmax_{(\ell,j):j\neq z^\ell}T_{k+1}^{\ell j}
    \eqsp.
\end{equation}
Set $T_{k+1}^\star=T_{k+1}^{\ell_{k+1}^\star j_{k+1}^\star}$.
\begin{enumerate}[wide,label=$\bullet$]
    \item If $T_{k+1}^\star=-\infty$, the procedure stops. Set $N=k$, keep $X_v=z$ for $v\in[s,T_k^\star]$, and update the residuals on this interval by \eqref{eq:erf-first-reaction-residual-update}.
    \item Otherwise, keep $X_v=z$ for $v\in(T_{k+1}^\star,T_k^\star]$ and, for every $v\in[T_{k+1}^\star,T_k^\star]$, evolve all residuals by
    \begin{equation}
        \label{eq:erf-first-reaction-residual-update}
        E_v^{\ell j}
        =E_{T_k^\star}^{\ell j}
        +\begin{cases}
            -\displaystyle\int_v^{T_k^\star}\rratemat{r}{z}{j}[\ell]\,\rmd r,
                & j\neq z^\ell,\\[0.4em]
            \displaystyle\int_v^{T_k^\star}\lambda_r^\ell(z^\ell)\,\rmd r,
                & j=z^\ell.
        \end{cases}
    \end{equation}
    Set $X_{T_{k+1}^\star}^{\ell_{k+1}^\star}=j_{k+1}^\star$ and $X_{T_{k+1}^\star}^m=z^m$ for $m\neq\ell_{k+1}^\star$.
\end{enumerate}

The recursion continues until no candidate event occurs before $s$. 

\subsubsection{Gillespie's global form}

The global form of Gillespie's algorithm \citep{gillespie1977exact} uses one exponential clock per jump. As above, it constructs $(X_u)_{u\in\ccint{s,t}}$ backward from $X_t\sim p_t$.

\textbf{Initialization.} Set $T_0^\star=t$ and draw $X_{T_0^\star}\sim p_t$.

\textbf{Induction.} At stage $k\geq0$, given $(T_k^\star,X_{T_k^\star})$, write $z=X_{T_k^\star}$, let $\bar\lambda_r(z)\eqdef\sum_{\ell=1}^{\len}\sum_{j:j\neq z^\ell}\rratemat{r}{z}{j}[\ell]$, draw $E_k\sim\operatorname{Exp}(1)$ independently of the past, and set
\begin{equation}
    \label{eq:gillespie-global-event-time}
    T_{k+1}^\star
    =
    \sup\defEns{u\in[s,T_k^\star]:\ \int_u^{T_k^\star}\bar\lambda_r(z)\eqsp\rmd r\geq E_k}
    \eqsp.
\end{equation}
\begin{enumerate}[wide,label=$\bullet$]
    \item If $T_{k+1}^\star=-\infty$, the procedure stops, so we set $N=k$ and $X_u=z$ for $u\in[s,T_k^\star]$.
    \item Otherwise, draw $(\ell_{k+1}^\star,j_{k+1}^\star)$ among the pairs with $j\neq z^\ell$ according to
    \[
        \pP\bigl((\ell_{k+1}^\star,j_{k+1}^\star)=(\ell,j)\mid T_{k+1}^\star=u, X_{T_k^\star}=z\bigr)
        =\frac{\rratemat{u}{z}{j}[\ell]}{\bar\lambda_u(z)}\eqsp.
    \]
    Set $X_v=z$ for $v\in(T_{k+1}^\star,T_k^\star]$, set $X_{T_{k+1}^\star}^{\ell_{k+1}^\star}=j_{k+1}^\star$, leave the other positions unchanged, and continue the recursion.
\end{enumerate}
Under the same non-explosion conditions as in the first-reaction construction, this defines the reverse CTMC on $[s,t]$.

\subsection{Gillespie discretization with partially frozen rates}
\label{app:gillespie-discretization}

Exact simulation with time-inhomogeneous Gillespie dynamics requires integrating the time-dependent transition rates and reevaluating their learned component after every one-position jump. To reduce the number of neural function evaluations, consider the grid $0=\tk{0}<\cdots<\tk{\last}=1$, with $\sk{i}\eqdef\tk{i-1}$, and the partially frozen reverse rate matrix $\rratemat{r}{\cdot}{\cdot;\tk{i}}$ introduced in \Cref{sec:integrating-exponential-residual-flow}. On each interval $[\sk{i},\tk{i}]$, we approximate it with the learned neural network $\rratemat{r}{\cdot}{\cdot;\tk{i}}[\param]$ and use a single forward pass of the model and may freeze the same component as for ERF: the full rate matrix, the score, the denoiser, or the leave-one-out denoiser. Note that this discretization is almost the same as the discretization of the ERF sampler but Gillespie discretization offers slightly more flexibility than ERF because it does not require the forward rates associated with the approximate frozen backward rates.
Applying the first-reaction construction to this approximate rate matrix gives \Cref{alg:gillespie-discretization}.

\begin{algorithm}[t]
    \caption{Gillespie discretization with partially frozen reverse rates}
    \label{alg:gillespie-discretization}
    \begin{algorithmic}[1]
        \Require Grid $0=\tk{0}<\cdots<\tk{\last}=1$ and a partial-freezing parameterization
        \State Draw $X_1\sim\pdata{1}{}{}$
        \For{$i=\last,\ldots,1$} \Comment{\textcolor{red}{Discretize time}}
            \State Set $r\gets\tk{i}$, $s\gets\sk{i}$, and $z\gets X_{\tk{i}}$
            \State Evaluate the learned component at $(\tk{i},z)$ and define $\rratemat{\cdot}{\cdot}{\cdot;\tk{i}}[\param]$ \Comment{\textcolor{red}{Approximate the rates}}
            \While{$r>s$}
                \State Draw $E^{\ell j}\simiid\operatorname{Exp}(1)$ for every $(\ell,j)$ such that $j\neq z^\ell$
                \State $T^{\ell j}\gets\sup\defEns{u\in[s,r]:\int_u^r\rratemat{v}{z}{j;\tk{i}}[\param,\ell]\eqsp\rmd v\geq E^{\ell j}}$ \Comment{\textcolor{red}{Use the approximate rates}}
                \State $(\widehat\ell,\widehat j)\gets\argmax_{(\ell,j)}T^{\ell j}$ and $u\gets T^{\widehat\ell\widehat j}$
                \If{$u<s$}
                    \State $r\gets s$
                \Else
                    \State $z^{\widehat\ell}\gets\widehat j$ and $r\gets u$
                \EndIf
            \EndWhile
            \State $X_{\sk{i}}\gets z$
        \EndFor
        \State \Return $X_0$
    \end{algorithmic}
\end{algorithm}

\subsection{Equivalence of leave-one-out-frozen Gillespie and ancestral sampling}
\label{app:frozen-loo-ancestral}

Fix $[s,t]=[\sk{i},\tk{i}]$, $X_t=x$, and $\widehat p_i^\ell\eqdef\denoiser{t}{}{x}[\param]^\ell$. The transition in \eqref{eq:plugin-reverse}, using the bridge extension \eqref{eq:bridge-prob} is
\begin{equation}
    \label{eq:frozen-loo-ancestral-kernel}
    \begin{aligned}
        \pdata{s\tbar t}{x}{w}[\param]
        &=\prod_{\ell=1}^{\len}
        \fw{s\tbar0,t}{\widehat p_i^\ell,x^\ell}{w^\ell}[\ell]\\
        &=\prod_{\ell=1}^{\len}
        \frac{
            \fw{s\tbar0}{\widehat p_i^\ell}{w^\ell}[\ell]
            \fw{t\tbar s}{w^\ell}{x^\ell}[\ell]
        }{
            \fw{t\tbar0}{\widehat p_i^\ell}{x^\ell}[\ell]
        }.
    \end{aligned}
\end{equation}

For $r\in[s,t]$ and $y\neq z^\ell$, the approximate rates that we get by freezing the leave-one-out denoiser are
\begin{align}
    \label{eq:reverse-rate-loo}
    \rratemat{r}{z}{y;\tk{i}}[\param,\ell,\mathrm{loo}]
    =
    \ratemat{r}{y}{z^\ell}[\ell]
    \frac{
        \fw{r\tbar0}{\widehat p_i^\ell}{y}[\ell]
    }{
        \fw{r\tbar0}{\widehat p_i^\ell}{z^\ell}[\ell]
    }.
\end{align}
Since $\fw{r\tbar0}{\widehat p_i^\ell}{y}[\ell]$ is exactly the marginal at time $r$ of a single-token CTMC initialized from $\categorical(\widehat p_i^\ell)$ and evolved with $\ratemat{r}{y}{z^\ell}[\ell]$, \eqref{eq:reverse-rate-loo} is exactly the time-reversed rates of this process. Therefore we know the transition from $x^\ell$ at time $t$ to $w^\ell$ at time $s$ in close form:
\begin{align*}
    \frac{
        \fw{s\tbar0}{\widehat p_i^\ell}{w^\ell}[\ell]
        \fw{t\tbar s}{w^\ell}{x^\ell}[\ell]
    }{
        \fw{t\tbar0}{\widehat p_i^\ell}{x^\ell}[\ell]
    }
    =\fw{s\tbar0,t}{\widehat p_i^\ell,x^\ell}{w^\ell}[\ell].
\end{align*}
The frozen rates act independently across positions. Thus on the full joint distribution, exact Gillespie sampling gives
\begin{align*}
    \prod_{\ell=1}^{\len}
    \fw{s\tbar0,t}{\widehat p_i^\ell,x^\ell}{w^\ell}[\ell]
    =\pdata{s\tbar t}{x}{w}[\param].
\end{align*}

Therefore, the proposed discretization of Gillespie's algorithm with leave-one-out frozen rates is equivalent to standard ancestral sampling used in the discrete-time setting.
The same conclusion does not hold in general when freezing the score or the ordinary denoising posterior.

\subsection{Euler and categorical tau-leaping}
\label{app:ctmc-euler-tau}

\paragraph{Token-wise Euler.} Token-wise Euler freezes the state and time dependence of the rate matrix, effectively drawing the positions conditionally independently. For each $\ell$, its categorical probabilities are
\begin{equation}
    \label{eq:token-euler-update}
    \pP(X_s^\ell=y\mid X_t=x)
    =
    \begin{cases}
        h\rratemat{t}{x}{y}[\param,\ell] & y\neq x^\ell\eqsp,\\
        1+h\rratemat{t}{x}{x^\ell}[\param,\ell] & y=x^\ell\eqsp.
    \end{cases}
\end{equation}
The step is valid whenever $h\sum_{y\neq x^\ell}\rratemat{t}{x}{y}[\param,\ell]\leq1$ for every position $\ell$.

\paragraph{Categorical $\tau$-leaping.} Categorical $\tau$-leaping uses the same frozen rates but solves Kolmogorov exactly for these frozen rate until the first jump. Draw independently
\begin{equation}
    \label{eq:categorical-tau-counts}
    N^{\ell y}_i\sim\operatorname{Poisson}\!\left(h\rratemat{t}{x}{y}[\param,\ell]\right)
    \eqsp,
    \qquad \ell\in\intset{1}{\len}\eqsp,
    \quad y\neq x^\ell\eqsp.
\end{equation}
If $\sum_{y\neq x^\ell}N_i^{\ell y}=1$, position $\ell$ is set to the unique $y$ for which $N_i^{\ell y}=1$, otherwise it remains equal to $x^\ell$. Equivalently, with $\bar\lambda_t^{\param,\ell}(x)\eqdef\sum_{y\neq x^\ell}\rratemat{t}{x}{y}[\param,\ell]$, the update probabilities are $\pP(X_s^\ell=y\mid X_t=x)=h\rratemat{t}{x}{y}[\param,\ell]\exp\{-h\bar\lambda_t^{\param,\ell}(x)\}$ for $y\neq x^\ell$, and the remaining mass is assigned to $x^\ell$.

\subsection{Explicit partially frozen updates for MDM and UDM}
\label{app:ctmc-partial-freezing}

Besides the score representation in \eqref{eq:score-parameterized-reverse-rate}, the density ratio in \eqref{eq:reverse-rate} admits the identities
\begin{align}
    \frac{\pdata{t}{x^{-\ell}}{w^\ell}[\ell]}{\pdata{t}{x^{-\ell}}{x^\ell}[\ell]}
    &=
    \sum_{x_0\in\msv}
    \frac{\fw{t\tbar0}{x_0}{w^\ell}[\ell]}{\fw{t\tbar0}{x_0}{x^\ell}[\ell]}
    \pdata{0\tbar t}{x}{x_0}[\ell]
    \eqsp,
    \label{eq:reverse-ratio-denoising-posterior}\\
    \frac{\pdata{t}{x^{-\ell}}{w^\ell}[\ell]}{\pdata{t}{x^{-\ell}}{x^\ell}[\ell]}
    &=
    \frac{\fw{t\tbar0}{\loodenoiser{t}{}{x}^\ell}{w^\ell}[\ell]}{\fw{t\tbar0}{\loodenoiser{t}{}{x}^\ell}{x^\ell}[\ell]}
    \eqsp.
    \label{eq:reverse-ratio-loo}
\end{align}

These identities suggest three freezing schemes, by freezing the score in \eqref{eq:score-parameterized-reverse-rate}, the denoiser in \eqref{eq:reverse-ratio-denoising-posterior} or the leave-one-out denoiser in \eqref{eq:reverse-ratio-loo}. We give their explicit forms below.

\paragraph{MDM.} Since $\bpi^\ell=\mask$, \eqref{eq:forward-rate} gives $\ratemat{v}{y}{z^\ell}[\ell]=(-\dot\alpha_v/\alpha_v)\indic_{z^\ell=\mask}$ for $y\neq z^\ell$. Hence \eqref{eq:reverse-rate} permits a reverse jump only from $z^\ell=\mask$ to $y\neq\mask$. Under score freezing, the rate and its integral are
\begin{align*}
    \rratemat{v}{z}{y;\tk{i}}[\param,\ell]
    &=\langle\oh{y},\score{\tk{i}}{}{X_{\tk{i}}}[\param]^\ell\rangle
    \left(-\frac{\dot\alpha_v}{\alpha_v}\right),
    &
    \int_u^r\rratemat{v}{z}{y;\tk{i}}[\param,\ell]\,\rmd v
    &=\langle\oh{y},\score{\tk{i}}{}{X_{\tk{i}}}[\param]^\ell\rangle
    \log\left(\frac{\alpha_u}{\alpha_r}\right).
\end{align*}
Under denoising-posterior freezing, they are
\begin{align*}
    \rratemat{v}{z}{y;\tk{i}}[\param,\ell]
    &=\pdata{0\tbar \tk{i}}{X_{\tk{i}}}{y}[\param,\ell]
    \left(-\frac{\dot\alpha_v}{1-\alpha_v}\right),
    &
    \int_u^r\rratemat{v}{z}{y;\tk{i}}[\param,\ell]\,\rmd v
    &=\pdata{0\tbar \tk{i}}{X_{\tk{i}}}{y}[\param,\ell]
    \log\left(\frac{1-\alpha_r}{1-\alpha_u}\right).
\end{align*}
For masked diffusion, leave-one-out-denoiser and denoiser coincide on non-masked tokens so the rate and its integral have the same form as under denoising-posterior freezing.

\paragraph{UDM.} Under score freezing, every $y\neq z^\ell$ has
\begin{align*}
    \rratemat{v}{z}{y;\tk{i}}[\param,\ell]
    &=\frac{\langle\oh{y},\score{\tk{i}}{}{X_{\tk{i}}}[\param]^\ell\rangle}{\vocab}
    \left(-\frac{\dot\alpha_v}{\alpha_v}\right),
    &
    \int_u^r\rratemat{v}{z}{y;\tk{i}}[\param,\ell]\,\rmd v
    &=\frac{\langle\oh{y},\score{\tk{i}}{}{X_{\tk{i}}}[\param]^\ell\rangle}{\vocab}
    \log\left(\frac{\alpha_u}{\alpha_r}\right).
\end{align*}
For denoising-posterior freezing, the rate is
\begin{align*}
    \rratemat{v}{z}{y;\tk{i}}[\param,\ell]
    &=
    \frac{1}{\vocab}\left(-\frac{\dot\alpha_v}{\alpha_v}\right)
    \left(
        1-\pdata{0\tbar \tk{i}}{X_{\tk{i}}}{y}[\param,\ell]
        -\pdata{0\tbar \tk{i}}{X_{\tk{i}}}{z^\ell}[\param,\ell]
    \right.
    \\
    &\hspace{7em}\left.
        +\pdata{0\tbar \tk{i}}{X_{\tk{i}}}{y}[\param,\ell]
        \frac{1+(\vocab-1)\alpha_v}{1-\alpha_v}
        +\pdata{0\tbar \tk{i}}{X_{\tk{i}}}{z^\ell}[\param,\ell]
        \frac{1-\alpha_v}{1+(\vocab-1)\alpha_v}
    \right),
\end{align*}
whose integral is
\begin{align*}
    \int_u^r\rratemat{v}{z}{y;\tk{i}}[\param,\ell]\,\rmd v
    &=
    \frac{1}{\vocab}\log\left(\frac{\alpha_u}{\alpha_r}\right)
    +\pdata{0\tbar \tk{i}}{X_{\tk{i}}}{y}[\param,\ell]
    \log\left(\frac{1-\alpha_r}{1-\alpha_u}\right)
    \\
    &\quad
    +\frac{\pdata{0\tbar \tk{i}}{X_{\tk{i}}}{z^\ell}[\param,\ell]}{\vocab-1}
    \log\left(
        \frac{1+(\vocab-1)\alpha_r}
        {1+(\vocab-1)\alpha_u}
    \right).
\end{align*}
Finally, under leave-one-out-denoiser freezing,
\begin{align*}
    \rratemat{v}{z}{y;\tk{i}}[\param,\ell]
    &=
    \frac{1}{\vocab}\left(-\frac{\dot\alpha_v}{\alpha_v}\right)
    \frac{
        \alpha_v\langle\oh{y},\denoiser{\tk{i}}{}{X_{\tk{i}}}[\param]^\ell\rangle
        +(1-\alpha_v)/\vocab
    }{
        \alpha_v\langle\oh{z}^\ell,\denoiser{\tk{i}}{}{X_{\tk{i}}}[\param]^\ell\rangle
        +(1-\alpha_v)/\vocab
    },
    \\
    \int_u^r\rratemat{v}{z}{y;\tk{i}}[\param,\ell]\,\rmd v
    &=
    \frac{1}{\vocab}\log\left(\frac{\alpha_u}{\alpha_r}\right)
    \\
    &\quad
    +\frac{
        \langle\oh{y},\denoiser{\tk{i}}{}{X_{\tk{i}}}[\param]^\ell\rangle
        -\langle\oh{z}^\ell,\denoiser{\tk{i}}{}{X_{\tk{i}}}[\param]^\ell\rangle
    }{
        \vocab\langle\oh{z}^\ell,\denoiser{\tk{i}}{}{X_{\tk{i}}}[\param]^\ell\rangle-1
    }
    \log\left(
        \frac{
            1+\bigl(\vocab\langle\oh{z}^\ell,\denoiser{\tk{i}}{}{X_{\tk{i}}}[\param]^\ell\rangle-1\bigr)\alpha_u
        }{
            1+\bigl(\vocab\langle\oh{z}^\ell,\denoiser{\tk{i}}{}{X_{\tk{i}}}[\param]^\ell\rangle-1\bigr)\alpha_r
        }
    \right).
\end{align*}
When $\vocab\langle\oh{z}^\ell,\denoiser{\tk{i}}{}{X_{\tk{i}}}[\param]^\ell\rangle-1=0$, the second term is understood by continuity as
\begin{align*}
    \left(
        \langle\oh{y},\denoiser{\tk{i}}{}{X_{\tk{i}}}[\param]^\ell\rangle
        -\langle\oh{z}^\ell,\denoiser{\tk{i}}{}{X_{\tk{i}}}[\param]^\ell\rangle
    \right)(\alpha_u-\alpha_r).
\end{align*}
After every UDM transition, $z^\ell$ is replaced by the selected token and the expressions are reevaluated for the new current token using the cached posterior or leave-one-out prediction before solving for the following event.

\subsubsection{From integrated rates to jump updates}

For each candidate transition $(\ell,y)$, the jump time $T^{\ell y}$ is the largest solution in $[s,r]$ of
\begin{equation}
    \label{eq:partially-frozen-first-reaction-time}
    \int_{T^{\ell y}}^r\rratemat{v}{z}{y;\tk{i}}[\param,\ell]\,\rmd v
    =E_r^{\ell y}\eqsp.
\end{equation}
If the integrated rate over $[s,r]$ is smaller than $E_r^{\ell y}$, set $T^{\ell y}=-\infty$. The MDM and UDM jump-time updates are given below.

\paragraph{Closed-form update for MDM.} All active rates share the same time-dependent factor, so we select the smallest normalized residual and evaluate $\alpha^{-1}$ once.

Throughout this paragraph, a quotient with zero denominator is understood as $+\infty$.

Under score freezing, select
\begin{align*}
    (\widehat\ell,\widehat y)
    \in
    \argmin_{\substack{\ell:z^\ell=\mask\\y\neq\mask}}
    \frac{E_r^{\ell y}}
    {\langle\oh{y},\score{\tk{i}}{}{X_{\tk{i}}}[\param]^\ell\rangle}.
\end{align*}
If the minimum exceeds $\log(\alpha_s/\alpha_r)$, no transition occurs before $s$. Otherwise, the selected transition occurs at the single time
\begin{align*}
    T^\star
    =
    \alpha^{-1}\!\left(
        \alpha_r
        \exp\!\left\{
            \frac{E_r^{\widehat\ell\widehat y}}
            {\langle\oh{\widehat y},\score{\tk{i}}{}{X_{\tk{i}}}[\param]^{\widehat\ell}\rangle}
        \right\}
    \right).
\end{align*}

For denoising-posterior freezing, the selection and inversion have the form in \Cref{app:practical-discretization}, with its cached denoiser probability replaced by $\pdata{0\tbar\tk{i}}{X_{\tk{i}}}{y}[\param,\ell]$. In MDM, the leave-one-out and standard denoisers coincide at masked positions.

\paragraph{Root finding for UDM.} Score-frozen UDM candidate times are also explicit. The candidate time for $(\ell,y)$ lies in $[s,r]$ when
\begin{align*}
    \frac{\langle\oh{y},\score{\tk{i}}{}{X_{\tk{i}}}[\param]^\ell\rangle}{\vocab}
    \log\left(\frac{\alpha_s}{\alpha_r}\right)
    \geq E_r^{\ell y}\eqsp,
\end{align*}
and, when its coefficient is positive,
\begin{align*}
    T^{\ell y}
    =
    \alpha^{-1}\!\left(
        \alpha_r
        \exp\!\left\{
            \frac{\vocab E_r^{\ell y}}
            {\langle\oh{y},\score{\tk{i}}{}{X_{\tk{i}}}[\param]^\ell\rangle}
        \right\}
    \right).
\end{align*}
For denoising-posterior and leave-one-out-denoiser freezing, \eqref{eq:partially-frozen-first-reaction-time} has no closed-form solution in general so we solve the equation with bisection. If the integrated rate over $[s,r]$ reaches $E_r^{\ell y}$, bisect $[s,r]$ repeatedly: at midpoint $m$, replace the lower endpoint by $m$ when $\int_m^r\rratemat{v}{z}{y;\tk{i}}[\param,\ell]\,\rmd v\geq E_r^{\ell y}$, and the upper endpoint by $m$ otherwise. The integral is continuous and nonincreasing in $m$, so the retained intervals converge to the largest solution.

\subsubsection{Exactness of leave-one-out freezing under factorization}
\label{sec:loo-freezing-exactness}

Suppose that the data distribution factorizes across positions, and write $p_0^\ell\eqdef\pE[\oh{X}_0^\ell]\in\Simplex{\vocab}$ for its $\ell$-th marginal. Since the forward process also acts independently across positions, its marginal remains factorized at every time $r$, with $\ell$-th marginal $\fw{r\tbar0}{p_0^\ell}{}[\ell]$. Independence then gives
\begin{align*}
    \loodenoiser{r}{}{z}^\ell
    =
    \pE[\oh{X}_0^\ell\mid X_r^{-\ell}=z^{-\ell}]
    =
    p_0^\ell
\end{align*}
for every $r$ and $z$. Thus, at optimum, the cached leave-one-out prediction is independent of both the endpoint $(t,x)$ and the evolving discrete state. If $z$ and $w$ differ only at position $\ell$, the leave-one-out-frozen rate consequently satisfies
\begin{equation}
    \label{eq:loo-freezing-factorized-exact}
    \rratemat{r}{z}{w;\tk{i}}[\param,\mathrm{loo}]
    =
    \ratemat{r}{w^\ell}{z^\ell}[\ell]
    \frac{\fw{r\tbar0}{p_0^\ell}{w^\ell}[\ell]}
         {\fw{r\tbar0}{p_0^\ell}{z^\ell}[\ell]}
    =
    \rratemat{r}{z}{w}.
\end{equation}
Hence the partially frozen generator is the exact reverse generator throughout every discretization interval, and its exact integration introduces no discretization error. The single-token case is the special case $\len=1$: the conditioning context is empty, so the leave-one-out prediction is simply the data distribution $p_0$.

The other potential frozen rate formulas, for score freezing and denoiser freezing, are given in \Cref{app:ctmc-partial-freezing}: they retain either $\score{\tk{i}}{}{x}[\param]^\ell$ or $\pdata{0\tbar\tk{i}}{x}{\cdot}[\param,\ell]$ as $r$ and the current state change. They do not have the same property of exactness when $\pdata{0}{}{}$ is factorized. For example, consider a one-position binary UDM with uniform reference distribution and $\pdata{0}{}{1}=\rho\in(0,1)\setminus\{1/2\}$. Since $\pdata{r}{}{1}=\alpha_r\rho+(1-\alpha_r)/2$, its exact score ratio and denoising posterior satisfy
\begin{align*}
    \langle\oh{2},\score{r}{}{1}^1\rangle
    &=\frac{\pdata{r}{}{2}}{\pdata{r}{}{1}},
    &
    \pdata{0\tbar r}{1}{1}[1]
    &=\frac{\rho(1+\alpha_r)}{2\pdata{r}{}{1}}.
\end{align*}
Both vary with $\alpha_r$, so substituting their cached values at $t$ into the rate formulas generally differs from the exact reverse rates at $r<t$. For MDM, the ordinary posterior at a masked position equals the leave-one-out posterior; under factorization both equal $p_0^\ell$, so denoiser freezing is also exact there.

\subsubsection{A counterexample for score freezing}
\label{sec:score-freezing-explosion}

We now show that freezing the score for the ERF procedure while keeping the original forward rates can lead to an infinite number of jumps in finite time.
Consider UDM with one position, two tokens, and the schedule $\alpha_t=\exp(-2t)$. The original forward rates are constant and satisfy
\begin{align*}
    \ratemat{t}{1}{2}
    =
    \ratemat{t}{2}{1}
    =1.
\end{align*}
Freeze the score at time $t_i=1/10$ and state $X_{t_i}=1$, with cached value $\score{t_i}{}{1}=(1,2)$. The score-frozen rates in UDM are then
\begin{align*}
    \rratemat{t}{1}{2;\tk{i}}[\param]=2,
    \qquad
    \rratemat{t}{2}{1;\tk{i}}[\param]=1.
\end{align*}
The ERF nevertheless retains the original forward exit rate, equal to one, as the passive speed in either token. These active and passive speeds are incompatible: if the frozen reverse rates were the time reversal of the original forward rates under a positive law $q_t$, the two directions would require simultaneously $q_t(2)/q_t(1)=2$ and $q_t(1)/q_t(2)=1$.

Let $(E_{t_i}^{11},E_{t_i}^{12})=(A,B)$ with $A$ and $B$ independent $\operatorname{Exp}(1)$ variables. Starting from $X_{t_i}=1$ and writing $S\eqdef A+B/2$, the first jump occurs after backward elapsed time $B/2$ and leaves the residual pair $(S,0)$ in token $2$. After another time $S$, the process returns to $1$ with residuals $(0,S)$. More generally, starting from $1$ with residuals $(0,R)$, a two-jump cycle takes time $R$ and returns to $1$ with residuals $(0,R/2)$. The successive cycles therefore have durations $S,S/2,S/4,\ldots$, so infinitely many jumps accumulate at
\begin{equation}
    \label{eq:score-frozen-erf-explosion-time}
    T_{\mathrm{acc}}
    =
    \frac{B}{2}+S+\sum_{n=0}^{\infty}\frac{S}{2^n}
    =3A+2B<\infty.
\end{equation}
For every $h>0$, the event $3A+2B<h$ has probability $1-3\exp(-h/3)+2\exp(-h/2)>0$. Hence the explosion occurs with positive probability on any interval of positive length, whereas Gillespie sampling with the same bounded frozen reverse rates is nonexplosive.

The issue is not with the score approximation, the cached score can be population-optimal. Indeed, choose
\begin{align*}
    \pdata{0}{}{1}
    =\frac{1}{2}-\frac{\exp(1/5)}{6},
    \qquad
    \pdata{0}{}{2}
    =\frac{1}{2}+\frac{\exp(1/5)}{6}.
\end{align*}
Then $\pdata{t_i}{}{1}=1/3$ and $\pdata{t_i}{}{2}=2/3$, so the exact score at $1$ is $(1,2)$. Conditional on $X_{t_i}=1$, the score-frozen ERF explodes before time zero on the event $3A+2B<t_i$, which has positive probability.

\subsubsection{Why the explosion cannot occur with leave-one-out freezing}
\label{sec:loo-freezing-nonexplosion}
We now show that the same phenomenon cannot happen with leave-one-out freezing. The key idea is to show that the leave-one-out-frozen reverse generator in \eqref{eq:loo-frozen-reverse-rate} is actually the exact time reversal of the original forward generator started from a different law at time 0. Therefore, \Cref{thm:erf-tr-main} applies and we cannot have an infinite number of jumps. Let
$\widehat p_i^\ell
    \eqdef
    \denoiser{\tk{i}}{}{X_{\tk{i}}}[\param]^\ell$, for $\ell\in\intset{1}{\len}$,
and define the product law
\begin{align*}
    \widehat p_{r\tbar i}(z)
    \eqdef
    \prod_{\ell=1}^{\len}
    \fw{r\tbar0}{\widehat p_i^\ell}{z^\ell}[\ell].
\end{align*}
Each factor is a marginal of the original forward process started at $p_0 = \prod_{\ell = 1}^{\len} \widehat p_i^\ell$, hence $\widehat p_{r\tbar i}$ is the marginal of the CTMC with the original forward rate started from $\widehat p_{i}$. If $z$ and $w$ differ only at position $\ell$, then \eqref{eq:loo-frozen-reverse-rate} becomes
\begin{align*}
    \rratemat{r}{z}{w;\tk{i}}[\param,\mathrm{loo}]
    =
    \frac{\widehat p_{r\tbar i}(w)}{\widehat p_{r\tbar i}(z)}
    \ratemat{r}{w}{z}.
\end{align*}
Thus the frozen reverse rates are the time reversal of the original forward rates under $\widehat p_{r\tbar i}$, and the forward exit rates $\lambda_r^\ell(z^\ell)$ are the passive speeds used by ERF.

Fix the cached predictions and an interval $[s,t]$ on which the forward and frozen reverse rates are finite and continuous and $\widehat p_{r\tbar i}(z)>0$ for every $r\in[s,t]$ and $z\in\msx$. Let $\mathcal B_i(z)$ be the set of $e\in(0,\infty)^{\len\vocab}$ for which the ERF started from $(X_t,E_t)=(z,e)$ makes infinitely many jumps on $[s,t]$. By \Cref{thm:exponential-residual-flow},
\begin{align*}
    0=\sum_{z\in\msx}\widehat p_{t\tbar i}(z)\,
    \operatorname{Exp}(1)^{\otimes\len\vocab}\bigl(\mathcal B_i(z)\bigr).
\end{align*}
Each $\mathcal B_i(z)$ has Lebesgue measure zero, since $\widehat p_{t\tbar i}(z)>0$ and the exponential density is positive on $(0,\infty)^{\len\vocab}$. In the actual run, the conditional law of $X_t$ given the cached predictions may differ from $\widehat p_{t\tbar i}$. The flow is still nonexplosive if the conditional law of $E_t$ given $X_t=z$ and those predictions has a density for every $z$: it then assigns zero probability to every $\mathcal B_i(z)$. Thus the set of exploding residuals in the score-freezing example, which has positive Lebesgue measure, cannot occur here.

\section{Net-flux rate matrix}
\label{app:net-flux}

\subsection{Net flux construction and relation to prior work}

The relevant decomposition is that of the probability flux induced by the reverse rates, rather than of the rates alone. Assume that $\pdata{t}{}{x}>0$ and, for $x\neq w$, define
\begin{align*}
    F_t(x,w)&\eqdef\pdata{t}{}{x}\rratemat{t}{x}{w},
    &
    S_t(x,w)&\eqdef\frac{F_t(x,w)+F_t(w,x)}{2},
    &
    J_t(x,w)&\eqdef F_t(x,w)-F_t(w,x).
\end{align*}
Here $S_t$ is the symmetric traffic along the unordered pair $\{x,w\}$ and $J_t$ is its antisymmetric current. Using $\rratemat{t}{x}{x}=-\sum_{w\neq x}\rratemat{t}{x}{w}$, the reverse Kolmogorov equation becomes
\begin{align*}
    \partial_t\pdata{t}{}{x}
    &=-\sum_{w}\pdata{t}{}{w}\rratemat{t}{w}{x}
      =\sum_{w\neq x}\bigl(F_t(x,w)-F_t(w,x)\bigr)
      =\sum_{w\neq x}J_t(x,w).
\end{align*}
Thus the marginal evolution depends only on $J_t$: the symmetric term $S_t$ contributes equally in both directions and cancels in every difference $F_t(x,w)-F_t(w,x)$. Keeping the same current while forbidding simultaneous flux in both directions uniquely gives $F_t^{\mathrm{NF}}(x,w)=[J_t(x,w)]_+$. Dividing this flux by the mass at its source yields
\begin{align*}
    \rratemat{t}{x}{w}[\mathrm{NF}]
    \eqdef
    \frac{[J_t(x,w)]_+}{\pdata{t}{}{x}}
    =\frac{\left[
        \pdata{t}{}{x}\rratemat{t}{x}{w}
        -\pdata{t}{}{w}\rratemat{t}{w}{x}
    \right]_+}{\pdata{t}{}{x}},
\end{align*}
with diagonal entries chosen so that each row sums to zero. Since $a_+-(-a)_+=a$, the net-flux rates preserve $J_t$ and hence the Kolmogorov equation, while $\rratemat{t}{x}{w}[\mathrm{NF}]\rratemat{t}{w}{x}[\mathrm{NF}]=0$. Moreover, any pair of nonnegative opposing fluxes with difference $J_t(x,w)$ has total flux at least $|J_t(x,w)|$, with equality for the net-flux construction. It is therefore the minimum-activity generator with the same pairwise current.

The time-reversal identity gives $\pdata{t}{}{w}\rratemat{t}{w}{x}=\pdata{t}{}{x}\ratemat{t}{x}{w}$, and consequently $\rratemat{t}{x}{w}[\mathrm{NF}]=[\rratemat{t}{x}{w}-\ratemat{t}{x}{w}]_+$.

A similar construction appears in \citet{campbell2024generativeflows}, however the two constructions are not equivalent and ours leads to an even less stochastic process. Let us write $\rratemat{t}{\cdot}{\cdot}[X_0]$ for the reverse generator conditioned on the clean endpoint. \citet{campbell2024generativeflows} applies net flux conditionally and then averages under the posterior of $X_0$, whereas ours first forms the unconditional generator and then applies net flux. Thus, for $x\neq w$, the two rates are respectively
\begin{align*}
    \pE\!\left[\rratemat{t}{x}{w}[\mathrm{NF},X_0]\mid X_t=x\right]
    \qquad\text{and}\qquad
    \rratemat{t}{x}{w}[\mathrm{NF}],
\end{align*}
and they are not equal in general because the positive-part operation in the net-flux projection does not commute with averaging. More precisely, its convexity gives
\begin{align*}
    \rratemat{t}{x}{w}[\mathrm{NF}]
    \leq
    \pE\!\left[\rratemat{t}{x}{w}[\mathrm{NF},X_0]\mid X_t=x\right].
\end{align*}
The inequality is strict when different clean endpoints induce opposing currents between $x$ and $w$. Hence our unconditional net-flux generator has no larger off-diagonal rate for any transition, and therefore no larger total jump intensity, while inducing the same marginal evolution. Interestingly, applying net flux once more after averaging the conditional construction recovers our generator.

\subsection{Forward net-flux rates and passive residuals}
\label{app:net-flux-forward-rates}

ERF requires the forward rates associated with the chosen reverse generator to evolve its passive residuals. For strictly positive marginals and exact canonical rates, time reversal gives, for $x\neq w$,
\begin{equation}
    \label{eq:net-flux-forward-rate}
    \ratemat{t}{x}{w}[\mathrm{NF}]
    \eqdef
    \frac{\pdata{t}{}{w}}{\pdata{t}{}{x}}
    \rratemat{t}{w}{x}[\mathrm{NF}]
    =\left[
        \ratemat{t}{x}{w}
        -\rratemat{t}{x}{w}
    \right]_+.
\end{equation}
Indeed, substituting \eqref{eq:net-flux-reverse-rate} gives $[\pdata{t}{}{w}\rratemat{t}{w}{x}-\pdata{t}{}{x}\rratemat{t}{x}{w}]_+/\pdata{t}{}{x}$, and the canonical time-reversal identity reduces its first term to $\ratemat{t}{x}{w}$. Diagonal entries are the negatives of the corresponding row sums. Thus both net-flux generators are obtained by subtracting the canonical forward and reverse rates in opposite orders and taking positive parts; no additional model or density evaluation is needed.

For position $\ell$, the passive speed must therefore be replaced by $b_t^{\mathrm{NF},\ell}(x)\eqdef\sum_{w:\,w^{-\ell}=x^{-\ell},\,w^\ell\neq x^\ell}\ratemat{t}{x}{w}[\mathrm{NF}]$. Their leave-one-out-frozen integrals and jump times are given in \Cref{app:practical-discretization}.

\subsection{Score and denoising-posterior freezing for UDM net flux}
\label{app:net-flux-other-freezing}

Fix $[s,t]=[\sk{i},\tk{i}]$, cache the learned output at $x=X_{\tk{i}}$, and let $z$ be the current configuration at time $r\in[s,t]$. For $y\neq z^\ell$, the original UDM forward rate is $-\dot\alpha_v/(\vocab\alpha_v)$. Applying the positive-part projection to the frozen canonical rates in \Cref{app:ctmc-partial-freezing} gives the following off-diagonal rates.

Under score freezing, the projected reverse rate and the complementary passive forward exit rate at position $\ell$ are
\begin{align*}
    \rratemat{v}{z}{y;\tk{i}}[\mathrm{NF},\param,\ell]
    &=\left(-\frac{\dot\alpha_v}{\vocab\alpha_v}\right)\left[
        \left\langle\oh{y},\score{\tk{i}}{}{x}[\param]^\ell\right\rangle-1
    \right]_+,\\
    b_v^{\mathrm{NF},\ell}(z)
    &=\left(-\frac{\dot\alpha_v}{\vocab\alpha_v}\right)
      \sum_{y\in\msv\setminus\{z^\ell\}}
      \left[1-\left\langle\oh{y},\score{\tk{i}}{}{x}[\param]^\ell\right\rangle\right]_+.
\end{align*}

Under denoising-posterior freezing, the corresponding rates are
\begin{align*}
    \rratemat{v}{z}{y;\tk{i}}[\mathrm{NF},\param,\ell]
    &=-\dot\alpha_v\left[
        \frac{\pdata{0\tbar\tk{i}}{x}{y}[\param,\ell]}{1-\alpha_v}
        -\frac{\pdata{0\tbar\tk{i}}{x}{z^\ell}[\param,\ell]}
              {1+(\vocab-1)\alpha_v}
    \right]_+,\\
    b_v^{\mathrm{NF},\ell}(z)
    &=-\dot\alpha_v
      \sum_{y\in\msv\setminus\{z^\ell\}}
      \left[
        \frac{\pdata{0\tbar\tk{i}}{x}{z^\ell}[\param,\ell]}
             {1+(\vocab-1)\alpha_v}
        -\frac{\pdata{0\tbar\tk{i}}{x}{y}[\param,\ell]}{1-\alpha_v}
      \right]_+.
\end{align*}
Each passive speed is the sum of the positive-part complements $[\ratemat{v}{z^\ell}{y}[\ell]-\rratemat{v}{z}{y;\tk{i}}[\param,\ell]]_+$ over $y\neq z^\ell$, as in \eqref{eq:net-flux-forward-rate}. The diagonal entries are the negatives of the corresponding row sums. The score-frozen rates integrate directly, while each expression inside the positive part of a posterior-frozen rate changes sign at most once as $\alpha_v$ varies, so its integral is obtained by splitting at that crossing.

\paragraph{Forward rates compatibility}. The new forward rates justify the ERF construction only if they evolve a positive distribution under which the frozen reverse rates are their time reversals. This need not hold for score or denoising-posterior freezing: in the binary score example of \Cref{sec:score-freezing-explosion}, the net-flux reverse rate from token $1$ to token $2$ is $[2-1]_+=1$, whereas the opposite forward rate is $[1-1]_+=0$. Any positive candidate distribution $q_v$ would therefore have to satisfy
\begin{align*}
    q_v(1)[2-1]_+ = q_v(2)[1-1]_+,
\end{align*}
which forces $q_v(1)=0$ so score freezing can create incompatibile forward and backward rates leading to an infinite number of jumps. For leave-one-out freezing, as in \Cref{sec:loo-freezing-nonexplosion}, we consider the distribution
\begin{align*}
    \widetilde p_{v\tbar i}(z)
    \eqdef
    \prod_{\ell=1}^{\len}
    \fw{v\tbar0}{\denoiser{\tk{i}}{}{x}[\param]^\ell}{z^\ell}[\ell],
\end{align*}
the product of the forward marginals initialized by the cached leave-one-out predictions. For discrete states $z,w$ differing at one position, the canonical time-reversal identity and its net-flux projection give
\begin{align*}
    \widetilde p_{v\tbar i}(z)\rratemat{v}{z}{w;\tk{i}}[\param,\mathrm{loo}]
    &=
    \widetilde p_{v\tbar i}(w)\ratemat{v}{w}{z},\\
    \widetilde p_{v\tbar i}(z)
    \left[\rratemat{v}{z}{w;\tk{i}}[\param,\mathrm{loo}]-\ratemat{v}{z}{w}\right]_+
    &=
    \left[\widetilde p_{v\tbar i}(w)\ratemat{v}{w}{z}
          -\widetilde p_{v\tbar i}(z)\ratemat{v}{z}{w}\right]_+\\
    &=
    \widetilde p_{v\tbar i}(w)
    \left[\ratemat{v}{w}{z}-\rratemat{v}{w}{z;\tk{i}}[\param,\mathrm{loo}]\right]_+.
\end{align*}
The second identity is the time-reversal balance for the new rates. The net-flux projection also preserves the difference between the two forward fluxes, so the new forward rates still evolve $\widetilde p_{v\tbar i}$. Hence the augmented-marginal argument applies when ERF is initialized from this product law, and the nonexplosion argument applies to other entry laws as in \Cref{sec:loo-freezing-nonexplosion}.

%% file: appendix/distillation.tex
\section{Additional material for distillation}
\label{app:distillation}

\subsection{Consistency distillation training}
\label{sec:erf-consistency-training-appendix}

The objective and its motivation are given in \Cref{sec:distillation}; the round-based training procedure is summarized below. Write $D$ for the selected forward or reverse KL divergence. In \Cref{alg:erf-consistency-distillation}, only the frozen target is refreshed between rounds and the pretrained teacher generates every trajectory. The pretrained teacher is the original diffusion model and is evaluated from the current tokens and time.

\begin{algorithm}[H]
    \caption{Exponential-residual consistency distillation with roundwise target refresh}
    \label{alg:erf-consistency-distillation}
    \begin{algorithmic}[1]
        \Require Pre-trained teacher parameters $\theta$, student parameters $\phi$, gap $\delta$, numbers of rounds $N$ and updates $M$, learning rate $\eta$, averaging decay $\rho$, and KL choice $D$
        \Ensure Distilled student parameters $\phi$
        \State $\phi^{\mathrm{EMA}}\gets\phi$
        \For{$n=1,\ldots,N$}
            \State $\bar\phi\gets\phi^{\mathrm{EMA}}$ \Comment{\textcolor{crimson}{Freeze the target for this round}}
            \For{$m=1,\ldots,M$}
                \State Sample $t\sim\unif(\delta,1)$ and set $s\gets t-\delta$
                \State Sample $X_0\sim\pdata{0}{}{}$ and $X_t\sim\fw{t\tbar0}{X_0}{}$
                \State Sample $E_t^{\ell j}\simiid\operatorname{Exp}(1)$ independently of $X_t$
                \State $(X_s,E_s)\gets\Psi^\theta_{s\tbar t}(X_t,E_t)$
                \State $\widehat{\mathcal L}\gets D\!\left(\pdata{0\tbar t}{X_t,E_t}{}[\phi],\operatorname{sg}\!\left[\pdata{0\tbar s}{X_s,E_s}{}[\bar\phi]\right]\right)$
                \State $\phi\gets\phi-\eta\nabla_\phi\widehat{\mathcal L}$
                \State $\phi^{\mathrm{EMA}}\gets\rho\phi^{\mathrm{EMA}}+(1-\rho)\phi$
            \EndFor
        \EndFor
        \State \Return $\phi$
    \end{algorithmic}
\end{algorithm}

The joint-refresh variant also replaces the model that supplies the ERF rollout predictions between rounds. The first round uses the pretrained diffusion model. In every later round, the rollout uses the same frozen EMA copy $\bar\phi$ as the consistency target. To advance from $t$ to $s$, we split the interval into numerical ERF substeps. At the beginning of a substep, say at time $u$, the frozen ERC network is evaluated with token input $X_u$, time input $u$, and residual input $-\log E_u$. For MDM, its output is the mean denoising distribution used in the masked ERF rates. For UDM, its output is the native leave-one-out distribution used directly in the UDM ERF rates. This output is frozen while one ERF substep advances the residuals and changes the tokens. The resulting $(X,E)$ becomes the network input at the next substep. Thus, the model is reevaluated at every numerical substep, but not after each token jump occurring within that substep. The complete procedure is summarized in \Cref{alg:erf-consistency-distillation-joint-refresh}.

\begin{algorithm}[H]
    \caption{Exponential-residual consistency distillation with roundwise target-and-teacher refresh}
    \label{alg:erf-consistency-distillation-joint-refresh}
    \begin{algorithmic}[1]
        \Require Pre-trained teacher parameters $\theta$, student parameters $\phi$, gap $\delta$, numbers of rounds $N$ and updates $M$, learning rate $\eta$, averaging decay $\rho$, and KL choice $D$
        \Ensure Distilled student parameters $\phi$
        \State $\phi^{\mathrm{EMA}}\gets\phi$
        \For{$n=1,\ldots,N$}
            \State $\bar\phi\gets\phi^{\mathrm{EMA}}$ \Comment{\textcolor{crimson}{Freeze the target for this round}}
            \For{$m=1,\ldots,M$}
                \State Sample $t\sim\unif(\delta,1)$ and set $s\gets t-\delta$
                \State Sample $X_0\sim\pdata{0}{}{}$ and $X_t\sim\fw{t\tbar0}{X_0}{}$
                \State Sample $E_t^{\ell j}\simiid\operatorname{Exp}(1)$ independently of $X_t$
                \State Choose the teacher grid $t=u_0>u_1>\cdots>u_J=s$
                \For{$q=0,\ldots,J-1$}
                    \If{$n=1$}
                        \State Evaluate the pretrained teacher with network inputs $(X_{u_q},u_q)$
                    \Else
                        \State Evaluate the frozen ERC model $\bar\phi$ with network inputs $(X_{u_q},u_q,-\log E_{u_q})$
                    \EndIf
                    \State Advance $(X_{u_q},E_{u_q})$ to $(X_{u_{q+1}},E_{u_{q+1}})$ by one ERF substep with this output frozen
                \EndFor
                \State $\widehat{\mathcal L}\gets D\!\left(\pdata{0\tbar t}{X_t,E_t}{}[\phi],\operatorname{sg}\!\left[\pdata{0\tbar s}{X_s,E_s}{}[\bar\phi]\right]\right)$
                \State $\phi\gets\phi-\eta\nabla_\phi\widehat{\mathcal L}$
                \State $\phi^{\mathrm{EMA}}\gets\rho\phi^{\mathrm{EMA}}+(1-\rho)\phi$
            \EndFor
        \EndFor
        \State \Return $\phi$
    \end{algorithmic}
\end{algorithm}

For UDM, the ERC network natively predicts a leave-one-out distribution. The two uses of this prediction must be distinguished. During a rollout, the pretrained teacher supplies its usual leave-one-out prediction, while a refreshed trajectory teacher supplies the residual-conditioned leave-one-out prediction of the frozen ERC model; in both cases, this leave-one-out distribution is used directly to construct the UDM ERF rates. In the consistency loss, both the current student prediction at $(X_t,E_t)$ and the frozen target prediction at $(X_s,E_s)$ are instead converted to mean denoising distributions before applying $D$. The conversion is pointwise in any additional residual conditioning, so both sides retain the same dependence on $E_u$. It is given by
\begin{equation}
    \label{eq:udm-loo-denoiser-conversion}
    \denoiser{u}{}{x}[\omega]^\ell
    =\softmax\!\left(
        \log\denoiser{u}{}{x}[\omega,\loo]^\ell
        +\log\!\left(1+\frac{\vocab\alpha_u}{1-\alpha_u}\right)\oh{x}^\ell
    \right).
\end{equation}

\subsection{Multi-step sampling}

\begin{algorithm}[H]
    \caption{Multi-step refresh or bridge sampling with a distilled student}
    \label{alg:erf-consistency-sampling}
    \begin{algorithmic}[1]
        \Require Student parameters $\phi$, decreasing grid $1=\tk{k}>\tk{k-1}>\cdots>\tk{1}>\tk{0}=0$ with $\sk{i}\eqdef\tk{i-1}$, and choice of refresh or bridge sampling
        \Output A clean state $\widehat X_0$
        \State Sample $X_{\tk{k}}\sim\pdata{1}{}{}$ and $E_{\tk{k}}^{\ell j}\simiid\operatorname{Exp}(1)$
        \For{$i=k,k-1,\ldots,1$}
            \State Sample $\widehat X_0\sim\pdata{0\tbar\tk{i}}{X_{\tk{i}},E_{\tk{i}}}{}[\phi]$
            \If{$i>1$}
                \State \textbf{either} sample $X_{\sk{i}}\sim\fw{\sk{i}\tbar0}{\widehat X_0}{}$ \Comment{\textcolor{crimson}{Refresh}}
                \State \textbf{or} sample $X_{\sk{i}}\sim\fw{\sk{i}\tbar0,\tk{i}}{\widehat X_0,X_{\tk{i}}}{}$ \Comment{\textcolor{crimson}{Bridge}}
                \State Sample a fresh $E_{\sk{i}}^{\ell j}\simiid\operatorname{Exp}(1)$
            \EndIf
        \EndFor
        \State \Return $\widehat X_0$
    \end{algorithmic}
\end{algorithm}

\subsection{Bridge sampling and the ERF coupling}
\label{app:bridge_sampling}

The bridge $\fw{s\tbar0,t}{y,x}{}$ is defined by the original forward diffusion; see \eqref{eq:bridge-prob}. Therefore, to obtain samples from the marginal $\pdata{s}{}{}$, it suffices to sample $X_0\sim\pdata{0\tbar t}{x}{}$ where $\pdata{0\tbar t}{x}{}$ is the original posterior, i.e., $\frac{\pdata{0}{}{y}\fw{t\tbar0}{y}{x}}{\pdata{t}{}{x}}$. However, the ERF process generally changes the coupling of $X_t$ and $X_0$. The coupling learned by the student is
\begin{equation}
    \label{eq:erc-bridge-compatibility}
    \pdata{0\tbar t}{x_t}{y}[\phi] \eqdef \pE_{E_t}\!\left[\pdata{0\tbar t}{x_t,E_t}{y}[\phi]\right].
\end{equation}
Consequently, the ideal endpoint distribution $\pdata{0\tbar t}{x_t,E_t}{}[\phi]$ need not give a sample from the forward posterior $\pdata{0\tbar t}{x}{}$ and the bridge generally does not produce $\pdata{s}{}{}$. Refresh sampling is always valid since it requires only $\widehat X_0\sim\pdata{0}{}{}$: then $\sum_y\pdata{0}{}{y}\fw{s\tbar0}{y}{z}=\pdata{s}{}{z}$, and fresh independent exponential residuals restore the augmented marginal.

For masked diffusion, however, each position jumps at most once, from $\mask$ to a token. A passive residual changed by ERF can therefore never affect a later jump. The ERF discrete path is therefore equivalent to Gillespie since the passive residuals are never re-used. MDM ERF thus has the original reverse-chain law, so its ideal endpoint student satisfies $\pdata{0\tbar t}{x_t}{y}[\phi]=\pdata{0\tbar t}{x_t}{y}$ and bridge sampling is exact in this case.

\subsection{D-MMD}
\label{sec:dmmd-implementation}

We implement D-MMD following \citet{hoogeboom2026beyond}. Let $\phi$, $\theta$, and $\psi$ denote the parameters of the student, frozen teacher, and auxiliary model. For masked diffusion, training uses the path
\begin{align*}
    X_0&\sim\pdata{0}{}{},
    &X_t&\sim\fw{t\tbar0}{X_0}{},
    &\widehat X_0^\ell&\sim\categorical\!\left(\denoiser{t}{}{X_t}[\phi]^\ell\right),
    &X_s&\sim\fw{s\tbar0,t}{\widehat X_0,X_t}{},
    \qquad s<t.
\end{align*}
All sampled states, including $\widehat X_0$ and $X_s$, are held fixed during differentiation, as in \citet[Algorithm~1]{hoogeboom2026beyond}. During the generator update, only the explicit student probabilities are differentiated; teacher and auxiliary predictions are fixed. During the auxiliary update, the student and teacher predictions are fixed and only the auxiliary parameters are differentiated. Thus the displayed expressions specify surrogate losses with these stop-gradient conventions, rather than derivatives through the parameter-dependent sampling law. Writing $w_s=(1-\alpha_s)^{-1}$, for masked diffusion we alternate the objectives
\begin{equation}
    \label{eq:dmmd-masked-loss}
    \begin{aligned}
        \mathcal L_{\mathrm{gen}}(\phi)
        &=-\pE\!\left[w_s\sum_{\ell:X_s^\ell=\mask}
        \left\langle \denoiser{t}{}{X_t}[\phi]^\ell,
        \log \denoiser{s}{}{X_s}[\theta]^\ell-\log \denoiser{s}{}{X_s}[\psi]^\ell\right\rangle\right],\\
        \mathcal L_{\mathrm{aux}}(\psi)
        &=-\pE\!\left[w_s\sum_{\ell:X_s^\ell=\mask}
        \left\langle \denoiser{t}{}{X_t}[\phi]^\ell+\denoiser{s}{}{X_s}[\theta]^\ell,
        \log \denoiser{s}{}{X_s}[\psi]^\ell\right\rangle\right].
    \end{aligned}
\end{equation}

For uniform diffusion, the student, teacher, and auxiliary model all use leave-one-out predictions. We sample $\widehat X_0^\ell\sim\categorical(\denoiser{t}{}{X_t}[\phi,\loo]^\ell)$ and $X_s\sim\fw{s\tbar0,t}{\widehat X_0,X_t}{}$, and use the correction of \citet[App.~D]{hoogeboom2026beyond}: for $r<s$,
\begin{align*}
    \mathcal L_{\mathrm{gen}}^{\loo}(\phi)
    &=\pE\!\left[\frac{1}{s-r}\sum_{\ell=1}^{\len}
        \left\langle
        \fw{r\tbar0,s}{\denoiser{t}{}{X_t}[\phi,\loo]^\ell,X_s^\ell}{}[\ell],
        \right.\right.\\
    &\hspace{7em}\left.\left.
        \log\fw{r\tbar0,s}{\denoiser{s}{}{X_s}[\psi,\loo]^\ell,X_s^\ell}{}[\ell]
        -\log\fw{r\tbar0,s}{\denoiser{s}{}{X_s}[\theta,\loo]^\ell,X_s^\ell}{}[\ell]
        \right\rangle\right],\\
    \mathcal L_{\mathrm{aux}}^{\loo}(\psi)
    &=-\pE\!\left[\frac{1}{s-r}\sum_{\ell=1}^{\len}
        \left\langle
        \fw{r\tbar0,s}{\widehat X_0^\ell,X_s^\ell}{}[\ell]
        +\fw{r\tbar0,s}{\denoiser{s}{}{X_s}[\theta,\loo]^\ell,X_s^\ell}{}[\ell],
        \right.\right.\\
    &\hspace{7em}\left.\left.
        \log\fw{r\tbar0,s}{\denoiser{s}{}{X_s}[\psi,\loo]^\ell,X_s^\ell}{}[\ell]
        \right\rangle\right].
\end{align*}

\subsection{DCD}
\label{sec:dcd-implementation}

For a fair comparison, every uniform-diffusion baseline starts from the same checkpoints trained with the UDM objective. We therefore implement DCD from \citet{sahoo2025diffusion} directly in the UDM state space by marginalizing its Gaussian construction. Sahoo et al. draw $G^{\ell j}\simiid\normpdf(0,1)$ and define
\begin{align*}
    X_u^\ell
    &\eqdef\argmax_{j\in\intset{1}{\vocab}}\left\{
        \widetilde\alpha_u\indic_{j=X_0^\ell}
        +\sqrt{1-\widetilde\alpha_u^2}\,G^{\ell j}
    \right\},\qquad u\in\{s,t\}.
\end{align*}
The same $G^{\ell j}$ is used at both times, and $\widetilde\alpha_u$ is chosen so that $X_u^\ell\mid X_0^\ell\sim\categorical(\alpha_u\oh{X}_0^\ell+(1-\alpha_u)\one/\vocab)$. The ordering of $G^{\ell j}$ for $j\neq X_0^\ell$ is fixed, and $\widetilde\alpha_s/\sqrt{1-\widetilde\alpha_s^2}\geq\widetilde\alpha_t/\sqrt{1-\widetilde\alpha_t^2}$. Hence $X_t^\ell=X_0^\ell$ implies $X_s^\ell=X_0^\ell$, while $X_t^\ell\neq X_0^\ell$ implies $X_s^\ell\in\{X_t^\ell,X_0^\ell\}$. In particular, $\{X_s^\ell=j\}\subseteq\{X_t^\ell=j\}$ for $j\neq X_0^\ell$, so
\begin{align*}
    \pP(X_s^\ell=j\mid X_t^\ell=j,X_0^\ell)
    &=
    \frac{\pP(X_s^\ell=j\mid X_0^\ell)}
         {\pP(X_t^\ell=j\mid X_0^\ell)}
    =\frac{1-\alpha_s}{1-\alpha_t}.
\end{align*}
The remaining probability goes to $X_0^\ell$, giving
\begin{align*}
    X_s^\ell\mid(X_0^\ell,X_t^\ell)
    \sim\categorical\!\left(
        \frac{\alpha_s-\alpha_t}{1-\alpha_t}\oh{X}_0^\ell
        +\frac{1-\alpha_s}{1-\alpha_t}\oh{X}_t^\ell
    \right).
\end{align*}
This is a maximal coupling of the two one-position UDM marginals conditional on $X_0^\ell$ \citep[Proposition~9]{gourevitch2026udm}. Therefore, this transition can be reproduced without the underlying Gaussian trajectory. We sample it directly, without Gaussian variables, by drawing independently across positions $U^\ell\sim\unif(0,1)$ and $Y^\ell\sim\categorical(\one/\vocab)$ once and setting
\begin{align*}
    X_u^\ell=
    \begin{cases}
        Y^\ell, & U^\ell<1-\alpha_u,\\
        X_0^\ell, & U^\ell\geq1-\alpha_u.
    \end{cases}
    \qquad u\in\{s,t\}.
\end{align*}
The coupling induces the transition
\begin{align*}
    \pdata{s\tbar t}{x_t}{x_s}[\mathrm{max}]
    &=\sum_{x_0\in\msx}\pdata{0\tbar t}{x_t}{x_0}
    \prod_{\ell=1}^{\len}
    \categorical\!\left(
        x_s^\ell;
        \frac{\alpha_s-\alpha_t}{1-\alpha_t}\oh{x}_0^\ell
        +\frac{1-\alpha_s}{1-\alpha_t}\oh{x}_t^\ell
    \right).
\end{align*}
At the beginning of each round, we freeze the current student parameters $\phi^-$ and double the time separation,
\begin{align*}
    \phi^-&\gets\phi,
    &s&=(t-\Delta_k)_+,
    &\Delta_{k+1}&=2\Delta_k.
\end{align*}
We convert both leave-one-out outputs to mean predictions using the UDM identity above and minimize the reverse KL
\begin{equation}
    \label{eq:dcd-loss}
    \mathcal L_{\mathrm{DCD}}(\phi;\phi^-)
    =\pE\!\left[\frac{1}{\len}\sum_{\ell=1}^{\len}
    \kldivergence{\denoiser{t}{}{X_t}[\phi]^\ell}
    {\denoiser{s}{}{X_s}[\phi^-]^\ell}\right].
\end{equation}
For fixed $\phi^-$ and $t$, if the target probabilities are strictly positive and their conditional logarithms are integrable, the pointwise minimizer of \eqref{eq:dcd-loss} is therefore, for $j\in\intset{1}{\vocab}$,
\begin{align*}
    \left\langle\oh{j},\denoiser{t}{}{x_t}[\phi^\star]^\ell\right\rangle
    =\frac{
        \exp\!\left\{
            \pE_{x_s\sim\pdata{s\tbar t}{x_t}{}[\mathrm{max}]}
            \!\left[
                \log\left\langle\oh{j},\denoiser{s}{}{x_s}[\phi^-]^\ell\right\rangle
            \right]
        \right\}
    }{
        \displaystyle\sum_{k=1}^{\vocab}
        \exp\!\left\{
            \pE_{x_s\sim\pdata{s\tbar t}{x_t}{}[\mathrm{max}]}
            \!\left[
                \log\left\langle\oh{k},\denoiser{s}{}{x_s}[\phi^-]^\ell\right\rangle
            \right]
        \right\}
    }.
\end{align*}
If a target probability can vanish, this expression instead requires a nonempty common support with a positive finite normalizing constant. In particular, the exact mean-prediction boundary at $s=0$ is a point mass at $X_0^\ell$. If distinct clean values remain possible conditionally on $X_t$, the expected reverse KL against that boundary is infinite for every student distribution. An implementation including $s=0$ must therefore specify positive target smoothing or a positive lower diffusion time; the softmax conversion above also requires an endpoint convention when $\alpha_s=1$. At inference, each prediction is re-noised to the next time on the sampling grid, $X_s^\ell\sim\fw{s\tbar0}{\denoiser{t}{}{X_t}[\phi]^\ell}{}[\ell]$.

\subsection{iDLM}
\label{sec:idlm-implementation}

We implement iDLM \citep{li2026idlm} with a frozen teacher, a student, and a trainable auxiliary denoiser, initialized from the same pretrained checkpoint used by the other methods in the corresponding diffusion family. Training alternates student and auxiliary updates. The student minimizes the difference between teacher and auxiliary denoising losses on its generated distribution, while the auxiliary denoiser is trained on that distribution. For masked diffusion, sampled categorical states are held fixed during differentiation and gradients act on the student probabilities appearing explicitly in the loss.

For uniform diffusion, we retain the Gaussian relaxation used by iDLM to obtain differentiable noisy inputs, with relaxation temperature $10^{-3}$. The teacher remains the shared pretrained UDM checkpoint; no separately pretrained Gaussian diffusion model is introduced. Thus the Gaussian relaxation is retained for iDLM training, whereas the Gaussian variables in DCD's coupled forward samples are marginalized as described in \Cref{sec:dcd-implementation}. At evaluation, iDLM generates discrete tokens and is subject to the same sampler and temperature controls as the other students.

%% file: appendix/experimental_details.tex
\section{Experimental details}
\label{sec:experimental-details}

\subsection{Sudoku}
\label{sec:sudoku-experimental-details}

We train one MDM and one UDM teacher per difficulty. Each example concatenates an $81$-digit quiz, a separator, its $81$-digit solution, and an end-of-sequence token; \texttt{0} denotes a blank quiz position. The quiz and separator remain fixed, while the model predicts the solution positions. The vocabulary includes ten digits and five special tokens for padding, separation, masking, end of sequence, and unknown inputs. \Cref{tab:sudoku-datasets,tab:sudoku-model,tab:sudoku-training} summarize the configuration. Exact-grid accuracy requires all $81$ predicted positions to match the reference solution; constraint violations are also computed.

\begin{table}[htbp]
    \centering
    \small
    \caption{Sudoku datasets, generated with seed $42$.}
    \label{tab:sudoku-datasets}
    \begin{tabular}{lrrrr}
        \toprule
        Difficulty & Clues & Training examples & Validation examples & Workers \\
        \midrule
        Easy & $40$ & $48{,}000$ & $2{,}000$ & $4$ \\
        Medium & $35$ & $48{,}000$ & $2{,}000$ & $4$ \\
        Hard & $30$ & $48{,}000$ & $2{,}000$ & $8$ \\
        \bottomrule
    \end{tabular}
\end{table}

\begin{table}[htbp]
    \centering
    \small
    \caption{Sudoku teacher architecture and diffusion configuration.}
    \label{tab:sudoku-model}
    \begin{tabular}{lll}
        \toprule
        Parameter & MDM & UDM \\
        \midrule
        Backbone & \multicolumn{2}{c}{DiT-style transformer} \\
        Transformer blocks & \multicolumn{2}{c}{$8$} \\
        Hidden dimension & \multicolumn{2}{c}{$512$} \\
        Attention heads & \multicolumn{2}{c}{$8$} \\
        Conditioning dimension & \multicolumn{2}{c}{$128$} \\
        Padded/truncated sequence length & \multicolumn{2}{c}{$164$} \\
        Vocabulary size & \multicolumn{2}{c}{$15$} \\
        Dropout & \multicolumn{2}{c}{$0.1$} \\
        Trainable parameters & \multicolumn{2}{c}{$28{,}562{,}191$} \\
        \bottomrule
    \end{tabular}
\end{table}

\begin{table}[htbp]
    \centering
    \small
    \caption{Shared Sudoku teacher training recipe, applied to all six runs.}
    \label{tab:sudoku-training}
    \begin{tabular}{ll}
        \toprule
        Parameter & Value \\
        \midrule
        Optimizer steps & $20{,}000$ \\
        Batch size & $256$ \\
        Numerical precision & bfloat16 \\
        Optimizer & AdamW \\
        Peak learning rate & $3\times10^{-4}$ \\
        Learning-rate schedule & $2{,}500$ warmup steps, then constant \\
        AdamW $(\beta_1,\beta_2)$ & $(0.9,0.999)$ \\
        Optimizer epsilon & $10^{-8}$ \\
        Weight decay & $0$ \\
        Gradient clipping & Maximum norm $1$ \\
        Diffusion loss & ELBO \\
        Antithetic time sampling & Enabled \\
        Sampling epsilon & $10^{-3}$ \\
        Training seed & $1$ \\
        EMA decay & $0.9999$ \\
        \bottomrule
    \end{tabular}
\end{table}

%% file: appendix/experiments.tex
\section{Additional experiments}
\label{sec:additional-experiments}


\subsection{Sampling quality across NFE budgets}
\label{sec:sampling-sweeps-appendix}

\begin{figure}[htbp]
    \centering
    \includegraphics[width=\linewidth]{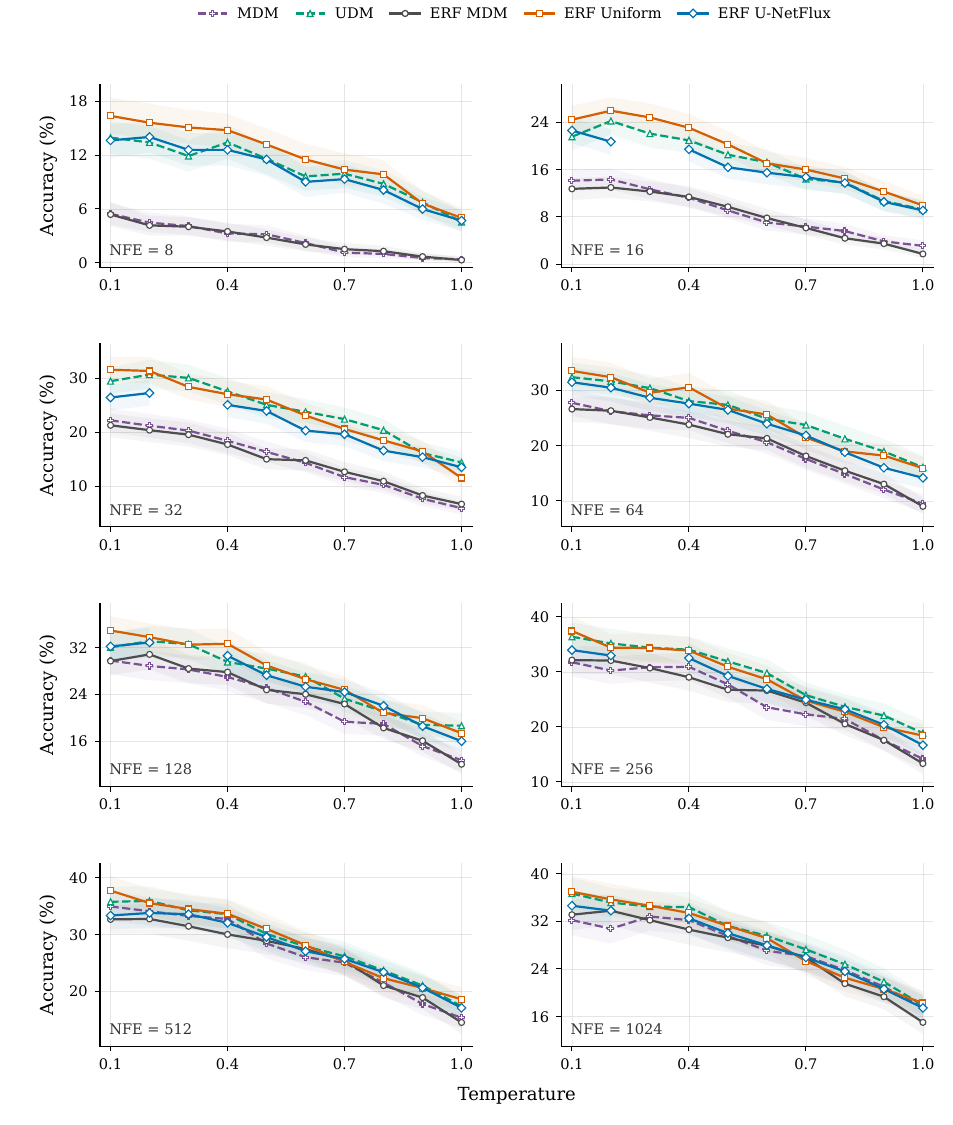}
    \caption{TinyGSM sampling across all evaluated NFE budgets. Each panel gives GSM8K accuracy across sampling temperatures, with pointwise $95\%$ bootstrap intervals. Dashed lines denote ancestral sampling and solid lines denote ERF.}
    \label{fig:appendix-sampler-tinygsm}
\end{figure}

\begin{figure}[htbp]
    \centering
    \includegraphics[width=\linewidth]{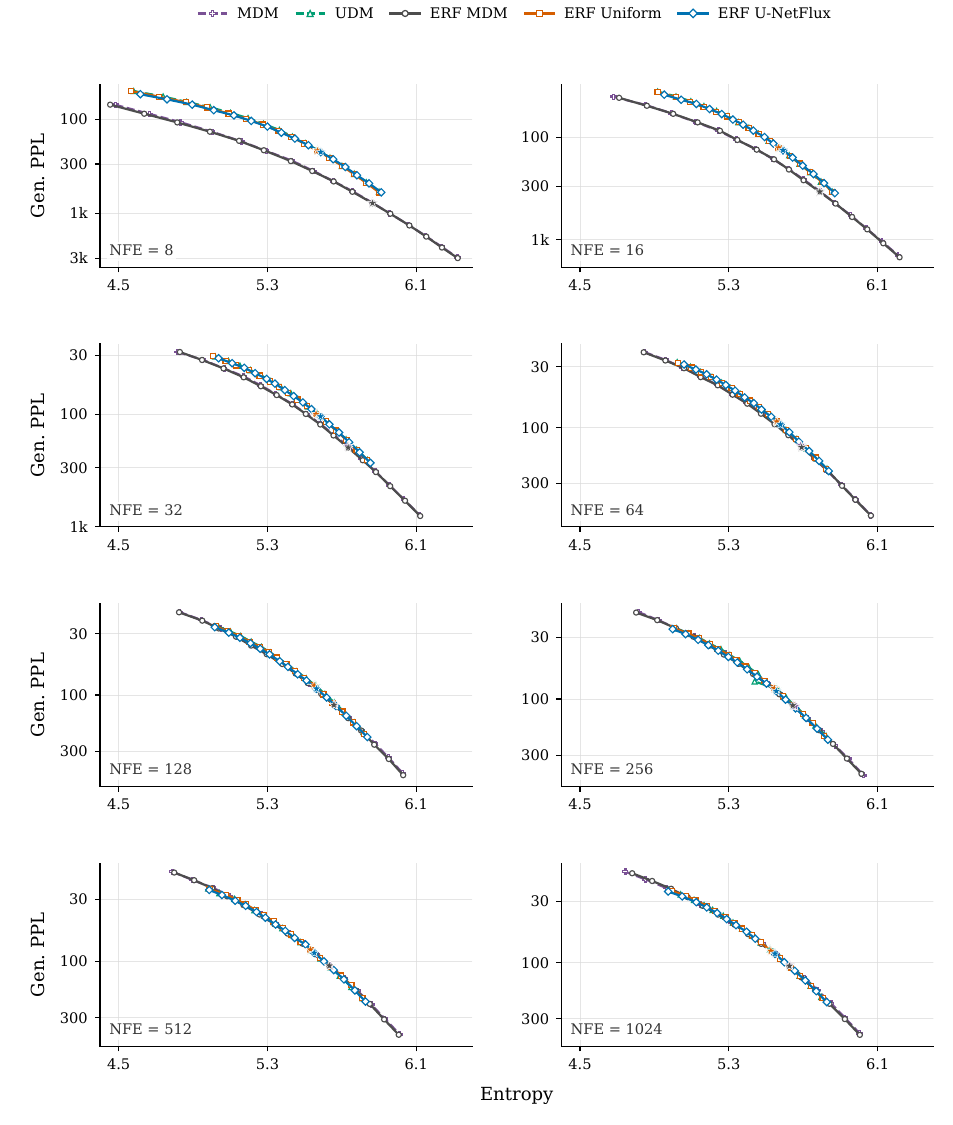}
    \caption{OpenWebText sampling across all evaluated NFE budgets. Markers follow the temperature grid from $0.80$ to $1.10$, with stars at $1$. Perplexity uses an inverted logarithmic axis, scaled independently in each panel.}
    \label{fig:appendix-sampler-owt}
\end{figure}

\Cref{fig:appendix-sampler-tinygsm,fig:appendix-sampler-owt} extend the main comparison to all eight evaluated NFE budgets, from $8$ to $1{,}024$. The samplers share a pretrained checkpoint within each diffusion family. TinyGSM evaluations use all $1{,}319$ GSM8K test questions and temperatures $\{0.1,0.2,\ldots,1\}$. Accuracy is the fraction of correct answers, with pointwise $95\%$ percentile-bootstrap intervals over questions. OpenWebText evaluations use temperatures $\{0.80,0.82,\ldots,1.10\}$ and report GPT-2 Large generative perplexity together with token entropy. The ancestral evaluations contain $960$ generated sequences per point and the ERF evaluations contain $480$.

\FloatBarrier
\subsection{Euler and categorical tau-leaping}
\label{sec:sampling-ctmc-comparison}

\begin{figure}[htbp]
    \centering
    \includegraphics[width=\linewidth]{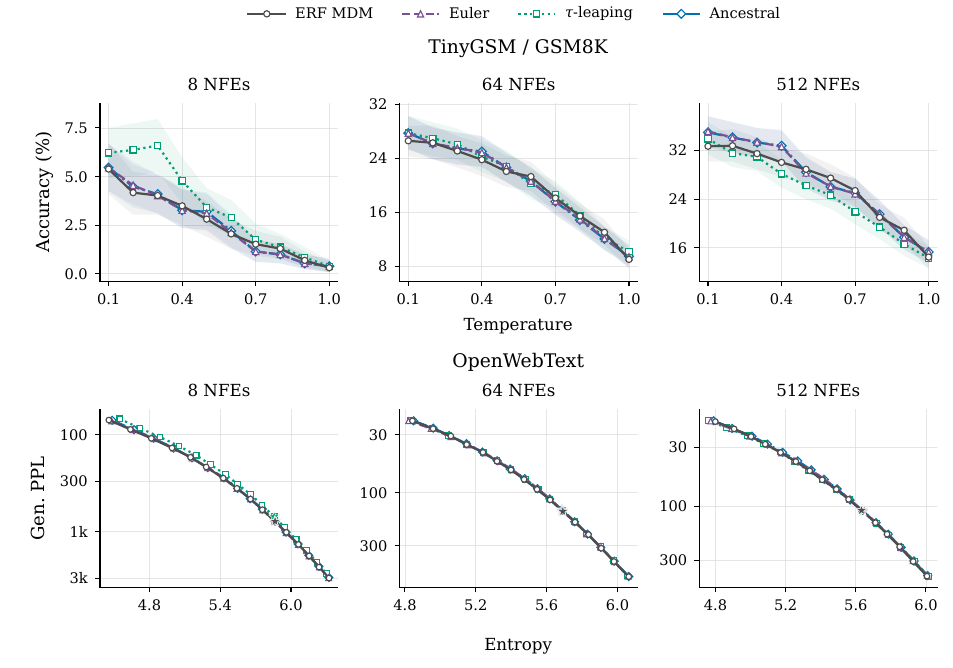}
    \caption{ERF, ancestral sampling, Euler, and categorical $\tau$-leaping for masked diffusion. The top row shows GSM8K accuracy across sampling temperatures, with pointwise $95\%$ bootstrap intervals; the bottom row shows OpenWebText entropy and GPT-2 Large generative perplexity. Stars mark temperature $1$. The perplexity axes are logarithmic and inverted, and panel limits follow the observed ranges.}
    \label{fig:ctmc-mdm}
\end{figure}

\begin{figure}[htbp]
    \centering
    \includegraphics[width=\linewidth]{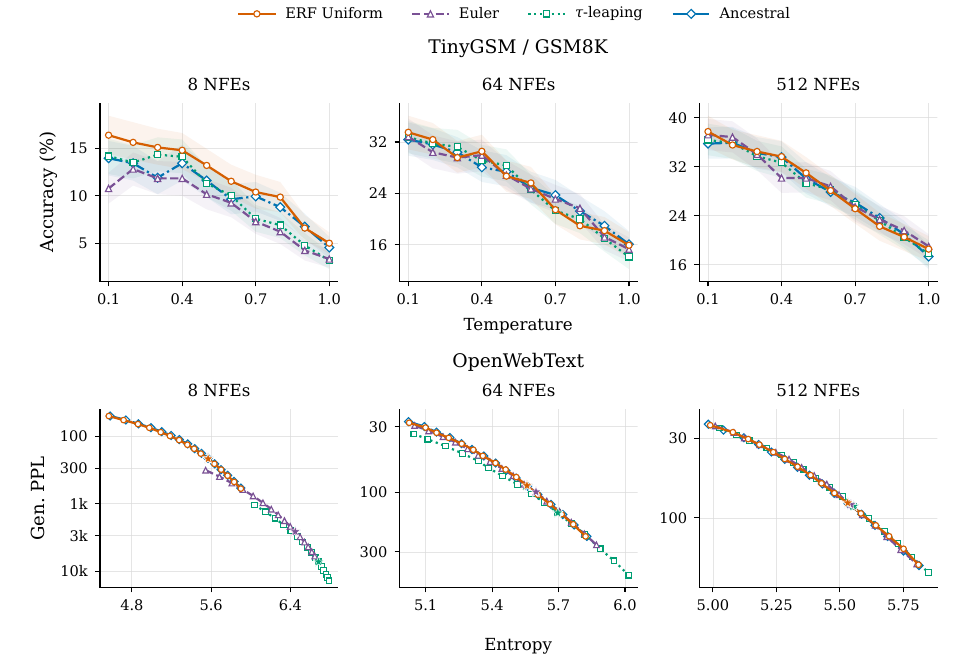}
    \caption{ERF Uniform, ancestral UDM, Euler, and categorical $\tau$-leaping for uniform diffusion, using the same plotting conventions as \Cref{fig:ctmc-mdm}. The ancestral reference uses temperature scaling before prediction conversion and $960$ OpenWebText sequences per point; the other methods use scaling after conversion and $480$ sequences. All four samplers share the same pretrained checkpoint.}
    \label{fig:ctmc-uniform}
\end{figure}

\begin{figure}[htbp]
    \centering
    \includegraphics[width=\linewidth]{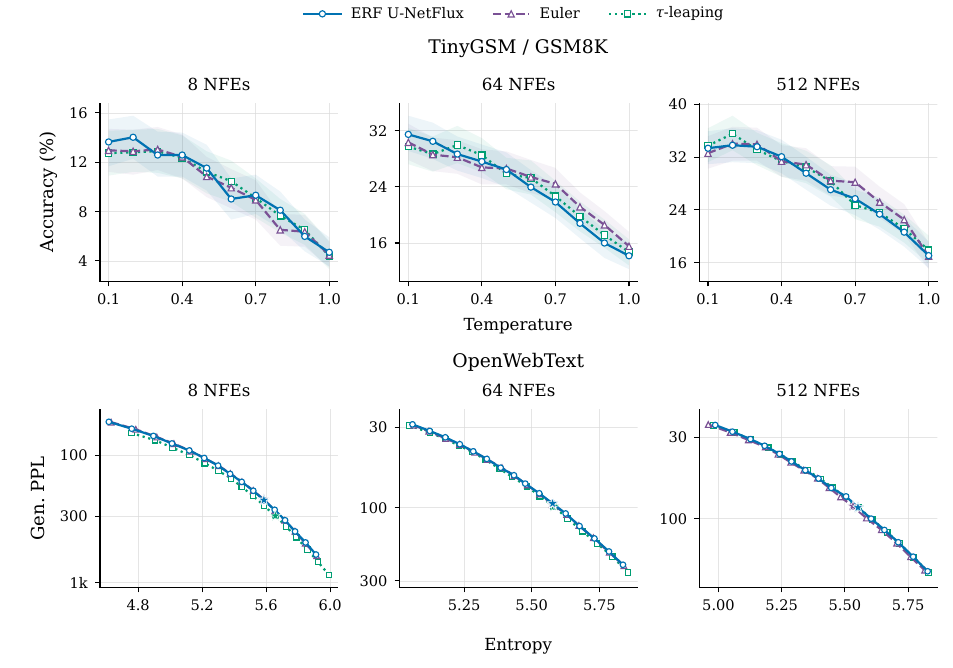}
    \caption{ERF U-NetFlux, Euler, and categorical $\tau$-leaping with U-NetFlux rates, using the same evaluation and plotting conventions as \Cref{fig:ctmc-mdm}. Each panel compares the three integration schemes with the same pretrained uniform-diffusion checkpoint.}
    \label{fig:ctmc-netflux}
\end{figure}

\Cref{fig:ctmc-mdm,fig:ctmc-uniform,fig:ctmc-netflux} compare ERF with the Euler and categorical $\tau$-leaping schemes described in \Cref{app:ctmc-euler-tau}, separately for masked, uniform, and U-NetFlux rates at $8$, $64$, and $512$ NFEs. Each rate family uses one pretrained checkpoint, and the masked and uniform comparisons also include the ancestral controls from the main sampling experiment. TinyGSM evaluations use all $1{,}319$ GSM8K test questions at temperatures $\{0.1,0.2,\ldots,1\}$. OpenWebText evaluations use temperatures $\{0.80,0.82,\ldots,1.10\}$, with $960$ sequences per point for the reused ancestral controls and $480$ for ERF, Euler, and categorical $\tau$-leaping; all runs use seed $1$. The ancestral UDM reference applies temperature to the raw model logits before the prediction conversion, whereas the other uniform samplers apply it to the converted prediction.

The Euler implementation clips the categorical weights and normalizes them when the frozen-rate step violates the probability constraint, while categorical $\tau$-leaping rejects multiple proposed jumps at a position. Both baselines replace the final scheduled update with a denoiser prediction.

\FloatBarrier
\subsection{Uniform substeps}
\label{sec:additional-sampling-experiments}

\begin{figure}[htbp]
    \centering
    \includegraphics[width=\linewidth]{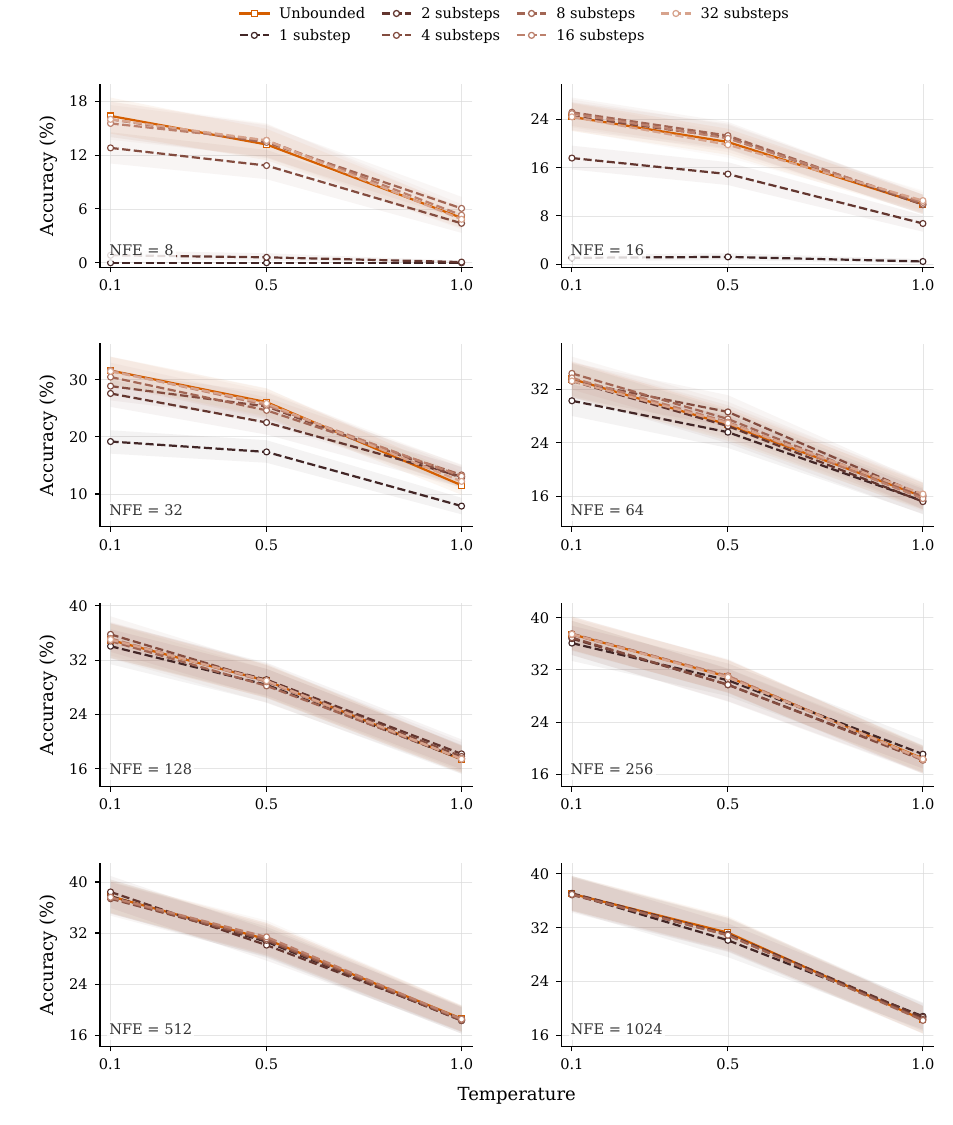}
    \caption{The full TinyGSM substep comparison. Curves show GSM8K accuracy at fixed NFE as the sampling temperature varies; bands are pointwise $95\%$ bootstrap intervals. Each legend entry specifies the number of substeps per frozen-model interval. Missing high-substep settings at larger NFE budgets are left absent.}
    \label{fig:erf-uniform-boundedness}
\end{figure}

\begin{figure}[htbp]
    \centering
    \includegraphics[width=\linewidth]{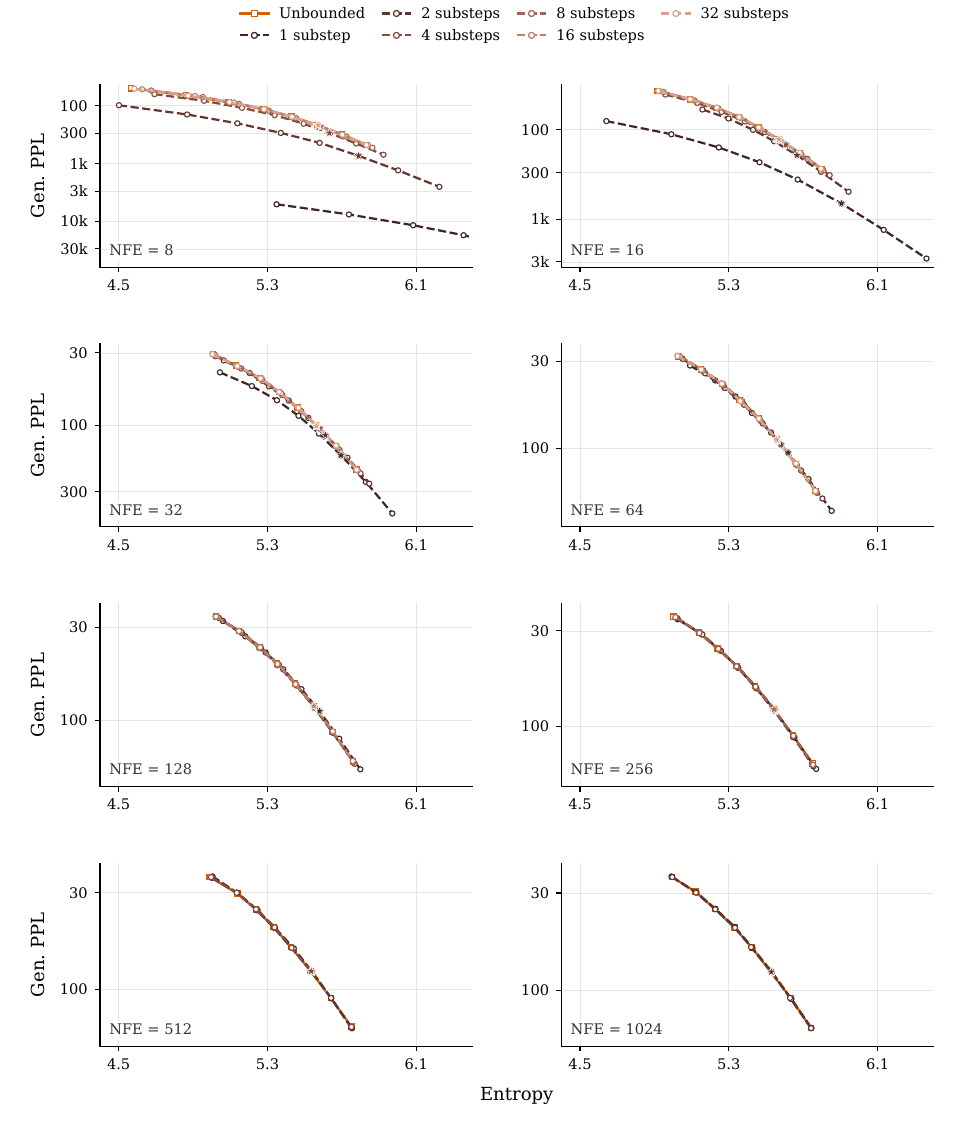}
    \caption{The full OpenWebText substep comparison. Curves show the entropy--perplexity trade-off over eight temperatures, with stars at $1$. The inverted logarithmic perplexity axis retains the observed range in each panel. Only completed evaluations are drawn, with lines interrupted at missing temperatures.}
    \label{fig:uniform-substeps-owt-full}
\end{figure}

The bounded variants divide each frozen-model interval into $1$, $2$, $4$, $8$, $16$, or $32$ substeps and permit at most one jump per position within a substep. The unbounded variant resolves every crossing. Substeps do not introduce additional model evaluations. \Cref{fig:erf-uniform-boundedness,fig:uniform-substeps-owt-full} display all completed settings across the eight NFE budgets. The TinyGSM grid uses temperatures $0.1$, $0.5$, and $1$; the OpenWebText grid uses $\{0.80,0.84,\ldots,1.08\}$.

\FloatBarrier
\subsection{Runtime and jump-count protocol}
\label{sec:sampler-profile-protocol}

The profiles use one AMD MI210, batches of four sequences, and generator seeds $1$, $2$, and $3$. For TinyGSM, the three batches contain the first twelve GSM8K test prompts, four successive prompts per seed. Prompt positions remain fixed in a $512$-token canvas. OpenWebText generation is unconditional with length $1{,}024$. Sampling temperatures are $0.1$ for TinyGSM and $1$ for OpenWebText. Both methods use the same checkpoint, bfloat16 model autocast, and float64 residual clocks.

Uniform explicitly uses $40$ bisection iterations per crossing-time solve. Its terminal time is $10^{-6}$, while U-NetFlux uses terminal time zero. Each method receives one warmup before measurement. Timings include sampling and exclude checkpoint loading and metric computation. Device synchronization brackets each timing interval.

\begin{figure}[htbp]
    \centering
    \includegraphics[width=\linewidth]{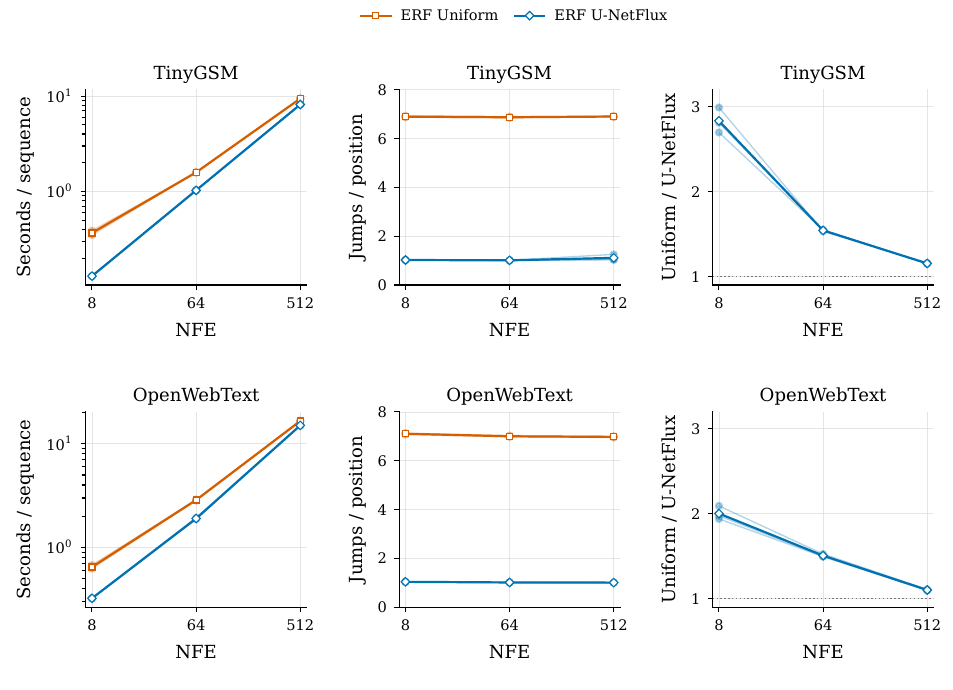}
    \caption{Individual runtime and jump-count measurements. Faint curves and points show the three seeds; the main curves show their means. In the right column, the main curve is the ratio of mean Uniform and U-NetFlux runtimes, while faint curves give the three paired ratios. The dotted line marks equal runtime.}
    \label{fig:sampler-efficiency-seeds}
\end{figure}

Jump counts come from a separate replay with the same generator seed and prompt batch. We check that its final tokens agree exactly with the uninstrumented run. Counts measure actual token changes and are divided by the number of generated positions, excluding fixed prompt positions. \Cref{fig:sampler-efficiency-seeds} exposes the three individual measurements; \Cref{tab:sampler-profile-detail} gives their means. These runs measure sampling cost on the stated batches.

\begin{table}[htbp]
    \centering
    \caption{Mean sampling time in seconds per sequence, speedup, and jumps per generated position across the three profile batches. Speedup is the ratio of mean Uniform and U-NetFlux runtimes.}
    \label{tab:sampler-profile-detail}
    \small
    \setlength{\tabcolsep}{3pt}
    \input{appendix/experiment_tables/sampler_profile}
\end{table}

\FloatBarrier
\subsection{Distillation evaluation}
\label{sec:distillation-evaluation-details}

The main distillation figures use the completed runs with budgets of $50$ GPU-hours per student on TinyGSM and $100$ GPU-hours on OpenWebText, all with global batch size $128$. The budget is the number of allocated GPUs multiplied by elapsed training-job time, including initialization, checkpointing and training validation, but excluding queue time and independent evaluation jobs. The number of student updates therefore varies by method. All methods share the original temperature-$1$ teacher checkpoint within each dataset and diffusion family, with top-$p$ truncation disabled. Uniform iDLM retains its Gaussian relaxation with the shared teacher, as detailed in \Cref{sec:idlm-implementation}.

The main figures use the final student EMA checkpoints from training seed $1$. All student and matching original-teacher evaluations use sampling seed $101$, so no across-seed uncertainty band is shown. ERC refreshes its frozen EMA target and rollout teacher at $25\%$, $50\%$ and $75\%$ of the training budget. We evaluate gaps $\delta\in\{1/8,1/16,1/32,1/64\}$ and select one gap per dataset and diffusion family, held fixed across NFEs. The selected gaps are $1/64$ for masked TinyGSM, $1/8$ for uniform TinyGSM, $1/64$ for masked OpenWebText and $1/16$ for uniform OpenWebText. UDM predicts mean LOO probabilities and converts them to mean predictions for ERC. The ERC objective is reverse KL on TinyGSM and forward KL on OpenWebText.

The main DCD curves use the final native consistency-distillation runs. The completed evaluation grid contains both bridge and refresh sampling for every student and teacher, with denoiser temperature placement for MDM and both denoiser and LOO placement for UDM. At temperature $1$, the two UDM temperature placements reuse the same evaluation within each sampler; these entries are aliases, not independent samples.

TinyGSM students and teachers in the main figure are evaluated on the same first $1{,}024$ GSM8K test questions, with a $512$-token canvas. The common sampling temperatures are $0.1$, $0.23$, $0.36$, $0.49$, $0.61$, $0.74$, $0.87$ and $1$, evaluated at $\{2,4,8,16,32,64\}$ NFEs. We maximize accuracy averaged with equal weight over these six NFE budgets to select one sampler, temperature placement and temperature per method; ERC additionally selects one gap. The selected configuration is fixed for the entire curve, and exact temperature ties favor the lower temperature. These are descriptive grid selections without a separate tuning set. The complete sampler-specific comparisons appear in \Cref{sec:distillation-sampler-ablations}.

The main OpenWebText figure uses bridge sampling and denoiser temperature placement for every method and teacher, at temperatures $0.8$, $0.85$, $0.9$, $0.95$, $1$, $1.05$, $1.1$ and $1.2$. Every point uses $2{,}048$ sequences of length $1{,}024$. To select the ERC gap within each family, we restrict each NFE comparison to the entropy interval shared by all four gaps, linearly interpolate log-perplexity as a function of entropy, and integrate over that interval, dividing by its width. We choose the gap with the lowest average of these normalized integrals over $\{2,4,8,16,32,64\}$ NFEs. This compares perplexity at matched entropy without extrapolation and fixes the same gap throughout the figure. The displayed curves contain measured points only. \Cref{sec:distillation-gap-ablations,sec:distillation-sampler-ablations} give the complete measured gap and sampler sweeps; their OpenWebText panels fit the visible points subject to entropy bounds $4$--$6$ and a generative-perplexity ceiling of $1{,}000$, clipping values outside these limits.

\FloatBarrier
\subsection{Sudoku distillation}
\label{sec:sudoku-distillation-results}

\Cref{tab:sudoku-mdm,tab:sudoku-udm} report exact-grid accuracy on the $2{,}000$ validation puzzles at each difficulty and NFE budget in $\{1,2,4,8,16,32,64\}$. We use EMA weights throughout and average sampling seeds $101$ and $303$ before selecting a configuration. For each method and NFE budget, the tables retain the highest two-seed mean over the ancestral bridge and categorical refresh samplers and over temperatures $\{0.1,0.25,0.5,0.7,0.8,0.9,1,1.1\}$. The ERC checkpoint is selected at $8$ NFEs with refresh sampling and then held fixed across NFE budgets. These selections are descriptive maxima over the evaluation grid rather than estimates from a separate validation sweep.

\begin{table*}[htbp]
    \centering
    \caption{Sudoku MDM exact-grid accuracy in percent. Subscripts report the selected sampler and temperature, with $\mathrm{R}$ denoting categorical refresh and $\mathrm{B}$ denoting the ancestral bridge. Bold denotes the highest accuracy within each difficulty and NFE budget.}
    \label{tab:sudoku-mdm}
    \scriptsize
    \setlength{\tabcolsep}{2pt}
    \input{appendix/experiment_tables/sudoku_mdm}
\end{table*}

\begin{table*}[htbp]
    \centering
    \caption{Sudoku UDM exact-grid accuracy in percent. Mean- and LOO-parameterized UDM use distinct teachers and method coverage. Selection, subscripts and boldface follow \Cref{tab:sudoku-mdm}.}
    \label{tab:sudoku-udm}
    \scriptsize
    \setlength{\tabcolsep}{2pt}
    \input{appendix/experiment_tables/sudoku_udm}
\end{table*}

\FloatBarrier
\input{appendix/distillation_ablations}
\FloatBarrier

%% file: appendix/experiment_tables/sampler_profile.tex
\begin{tabular}{@{}lrrrrrr@{}}
\toprule
Dataset & NFE & Uniform (s) & U-NetFlux (s) & Speedup & Uniform jumps & U-NetFlux jumps \\
\midrule
TinyGSM & $ 8 $ & $ 0.368 $ & $ 0.130 $ & $ 2.83 $ & $ 6.90 $ & $ 1.02 $ \\
TinyGSM & $ 64 $ & $ 1.586 $ & $ 1.028 $ & $ 1.54 $ & $ 6.87 $ & $ 1.01 $ \\
TinyGSM & $ 512 $ & $ 9.407 $ & $ 8.154 $ & $ 1.15 $ & $ 6.90 $ & $ 1.11 $ \\
OWT & $ 8 $ & $ 0.647 $ & $ 0.323 $ & $ 2.00 $ & $ 7.11 $ & $ 1.04 $ \\
OWT & $ 64 $ & $ 2.862 $ & $ 1.899 $ & $ 1.51 $ & $ 7.00 $ & $ 1.01 $ \\
OWT & $ 512 $ & $ 16.544 $ & $ 15.005 $ & $ 1.10 $ & $ 6.98 $ & $ 1.01 $ \\
\bottomrule
\end{tabular}

%% file: appendix/experiment_tables/sudoku_mdm.tex
\begin{tabular}{llccccccc}
\toprule
Difficulty & Method & 1 NFE & 2 NFE & 4 NFE & 8 NFE & 16 NFE & 32 NFE & 64 NFE \\
\midrule
Easy & Teacher & \sres{79.28}{R}{.1} & \sres{84.98}{R}{.1} & \sres{91.08}{R}{.1} & \sres{95.08}{R}{.25} & \sres{97.08}{R}{.1} & \sres{98.20}{R}{.1} & \sres{\mathbf{99.20}}{R}{.1} \\
 & ERC (ours) & \sres{\mathbf{81.08}}{B}{.1} & \sres{\mathbf{86.15}}{R}{.1} & \sres{\mathbf{92.88}}{R}{.1} & \sres{\mathbf{95.85}}{R}{.1} & \sres{\mathbf{97.45}}{R}{.25} & \sres{\mathbf{98.45}}{R}{.1} & \sres{98.90}{R}{.5} \\
 & D-MMD & \sres{78.43}{B}{.5} & \sres{84.18}{R}{.1} & \sres{91.98}{R}{.25} & \sres{95.03}{R}{.1} & \sres{97.13}{R}{1} & \sres{98.40}{R}{.1} & \sres{99.13}{R}{.5} \\
 & iDLM & \sres{79.70}{B}{.1} & \sres{85.03}{R}{.1} & \sres{91.25}{R}{.1} & \sres{95.08}{R}{.7} & \sres{97.10}{R}{.25} & \sres{98.23}{R}{.1} & \sres{99.10}{R}{.9} \\
 & SDTT & \sres{68.73}{R}{.1} & \sres{74.98}{R}{.1} & \sres{84.10}{R}{.1} & \sres{89.95}{R}{.1} & \sres{93.78}{R}{.25} & \sres{96.05}{R}{.25} & \sres{97.43}{R}{.9} \\
\cmidrule(lr){1-9}
Medium & Teacher & \sres{39.40}{R}{.1} & \sres{50.18}{R}{.1} & \sres{66.53}{R}{.25} & \sres{76.88}{R}{.1} & \sres{84.33}{R}{.1} & \sres{89.93}{R}{.1} & \sres{92.65}{R}{.1} \\
 & ERC (ours) & \sres{41.43}{B}{.1} & \sres{\mathbf{51.85}}{R}{.25} & \sres{\mathbf{68.28}}{R}{.25} & \sres{\mathbf{78.60}}{R}{.7} & \sres{\mathbf{84.88}}{R}{.1} & \sres{88.98}{R}{.5} & \sres{92.53}{R}{.25} \\
 & D-MMD & \sres{39.05}{B}{.1} & \sres{49.48}{B}{.7} & \sres{66.20}{R}{.25} & \sres{77.33}{R}{1.1} & \sres{83.58}{R}{1} & \sres{89.05}{R}{.5} & \sres{92.43}{R}{.1} \\
 & iDLM & \sres{\mathbf{41.58}}{B}{.1} & \sres{51.35}{R}{.1} & \sres{66.50}{R}{.7} & \sres{76.65}{R}{.25} & \sres{83.75}{R}{.25} & \sres{\mathbf{90.40}}{R}{.1} & \sres{\mathbf{93.05}}{R}{.1} \\
 & SDTT & \sres{12.45}{B}{.1} & \sres{21.75}{R}{.1} & \sres{40.53}{R}{.1} & \sres{55.15}{R}{.25} & \sres{66.85}{R}{.1} & \sres{76.05}{R}{.1} & \sres{82.50}{R}{.1} \\
\cmidrule(lr){1-9}
Hard & Teacher & \sres{2.68}{B}{.1} & \sres{7.23}{B}{.1} & \sres{20.75}{R}{.1} & \sres{36.10}{R}{.1} & \sres{\mathbf{47.88}}{R}{.1} & \sres{57.30}{R}{.1} & \sres{66.35}{R}{.25} \\
 & ERC (ours) & \sres{3.70}{B}{.25} & \sres{8.10}{R}{.25} & \sres{\mathbf{22.23}}{R}{.1} & \sres{\mathbf{36.58}}{R}{.1} & \sres{47.75}{R}{.1} & \sres{\mathbf{57.38}}{R}{.8} & \sres{\mathbf{66.43}}{R}{.8} \\
 & D-MMD & \sres{3.58}{B}{.25} & \sres{8.05}{B}{.25} & \sres{20.95}{R}{.25} & \sres{34.70}{R}{.25} & \sres{46.48}{R}{.25} & \sres{56.38}{R}{.8} & \sres{65.05}{R}{1.1} \\
 & iDLM & \sres{\mathbf{4.13}}{B}{.1} & \sres{\mathbf{8.83}}{B}{.1} & \sres{21.85}{R}{.1} & \sres{35.72}{R}{.5} & \sres{46.75}{R}{.25} & \sres{56.45}{R}{.1} & \sres{66.10}{R}{.7} \\
 & SDTT & \sres{0.55}{B}{.1} & \sres{2.65}{R}{.1} & \sres{11.45}{R}{.25} & \sres{22.60}{R}{.1} & \sres{31.78}{R}{.1} & \sres{41.25}{R}{.1} & \sres{51.43}{R}{.1} \\
\bottomrule
\end{tabular}

%% file: appendix/experiment_tables/sudoku_udm.tex
\begin{tabular}{llccccccc}
\toprule
Difficulty & Method & 1 NFE & 2 NFE & 4 NFE & 8 NFE & 16 NFE & 32 NFE & 64 NFE \\
\midrule
Easy & Teacher & \sres{74.50}{R}{.1} & \sres{92.85}{R}{.1} & \sres{96.83}{R}{.8} & \sres{98.05}{R}{.8} & \sres{98.70}{R}{.8} & \sres{99.35}{R}{.1} & \sres{\mathbf{99.63}}{R}{.5} \\
 & ERC (ours) & \sres{\mathbf{74.65}}{B}{.1} & \sres{\mathbf{93.28}}{R}{.1} & \sres{\mathbf{96.95}}{R}{1.1} & \sres{\mathbf{98.25}}{R}{1.1} & \sres{\mathbf{98.93}}{R}{.1} & \sres{\mathbf{99.55}}{R}{.1} & \sres{99.58}{R}{.7} \\
 & D-MMD & \sres{73.13}{B}{.1} & \sres{92.58}{R}{.25} & \sres{96.55}{R}{.1} & \sres{98.18}{R}{.8} & \sres{98.73}{R}{1.1} & \sres{99.30}{R}{.9} & \sres{99.48}{R}{.1} \\
 & DCD & \sres{74.50}{B}{.1} & \sres{92.30}{R}{.5} & \sres{96.55}{R}{.1} & \sres{97.80}{R}{.8} & \sres{98.50}{R}{.5} & \sres{99.13}{R}{.25} & \sres{99.28}{R}{.25} \\
\cmidrule(lr){1-9}
Medium & Teacher & \sres{40.03}{R}{.1} & \sres{74.38}{R}{.25} & \sres{85.95}{R}{.7} & \sres{90.78}{R}{1.1} & \sres{\mathbf{94.00}}{R}{1} & \sres{95.55}{R}{.1} & \sres{\mathbf{97.00}}{R}{.25} \\
 & ERC (ours) & \sres{40.95}{B}{.1} & \sres{\mathbf{75.78}}{R}{.1} & \sres{\mathbf{86.88}}{R}{.25} & \sres{\mathbf{91.23}}{R}{1} & \sres{93.93}{R}{.7} & \sres{\mathbf{95.85}}{R}{1} & \sres{96.98}{R}{1.1} \\
 & D-MMD & \sres{37.73}{B}{.1} & \sres{74.88}{R}{.25} & \sres{86.00}{R}{1.1} & \sres{90.40}{R}{.8} & \sres{93.48}{R}{.7} & \sres{95.55}{R}{.7} & \sres{96.60}{R}{.1} \\
 & DCD & \sres{\mathbf{41.25}}{B}{.1} & \sres{74.58}{R}{.8} & \sres{85.63}{R}{.5} & \sres{90.53}{R}{1.1} & \sres{93.28}{R}{.1} & \sres{95.20}{R}{.9} & \sres{96.65}{R}{.25} \\
\cmidrule(lr){1-9}
Hard & Teacher & \sres{7.80}{B}{.1} & \sres{39.78}{B}{.1} & \sres{57.55}{R}{.25} & \sres{\mathbf{68.35}}{R}{.8} & \sres{\mathbf{74.78}}{R}{.7} & \sres{80.25}{R}{.9} & \sres{85.73}{R}{1} \\
 & ERC (ours) & \sres{7.50}{B}{.1} & \sres{\mathbf{40.80}}{R}{.1} & \sres{\mathbf{58.88}}{R}{.1} & \sres{67.38}{R}{1} & \sres{74.63}{R}{.8} & \sres{\mathbf{80.40}}{R}{.8} & \sres{\mathbf{86.20}}{R}{1.1} \\
 & D-MMD & \sres{6.95}{B}{.1} & \sres{39.08}{B}{.25} & \sres{57.58}{R}{.9} & \sres{66.80}{R}{.5} & \sres{74.05}{R}{1.1} & \sres{79.55}{R}{.1} & \sres{85.00}{R}{.8} \\
 & DCD & \sres{\mathbf{7.93}}{B}{.25} & \sres{39.25}{B}{.25} & \sres{57.78}{R}{.25} & \sres{67.98}{R}{.5} & \sres{74.70}{R}{1} & \sres{79.55}{R}{.1} & \sres{84.68}{R}{1.1} \\
\bottomrule
\end{tabular}

%% file: appendix/distillation_ablations.tex
\subsection{Target and teacher updates}
\label{sec:distillation-policy-ablations}

The target is the stopped-gradient student on the right of the consistency loss; the transport teacher is the denoiser used to generate the paired ERF states. At a refresh, the target is replaced by the student EMA; target-and-teacher refresh also uses that copy for transport. Over $10{,}000$ updates, we compare a frozen target, per-step EMA, target refresh every $2{,}500$ updates, and target-and-teacher refresh. Each uses either a fixed gap $\delta=1/64$ or a gap increasing over four rounds from $1/64$ to $1/8$. At fixed gap, we also test $5{,}000$-update target refresh and EMA restricted to the target's conditioning projection. TinyGSM uses reverse KL and both inference samplers; OWT uses forward KL and bridge sampling. The figures use sampling seed $101$ and the measured temperature grid at $4$--$64$ NFEs.

\begin{figure}[htbp]
    \centering
    \includegraphics[width=0.88\linewidth]{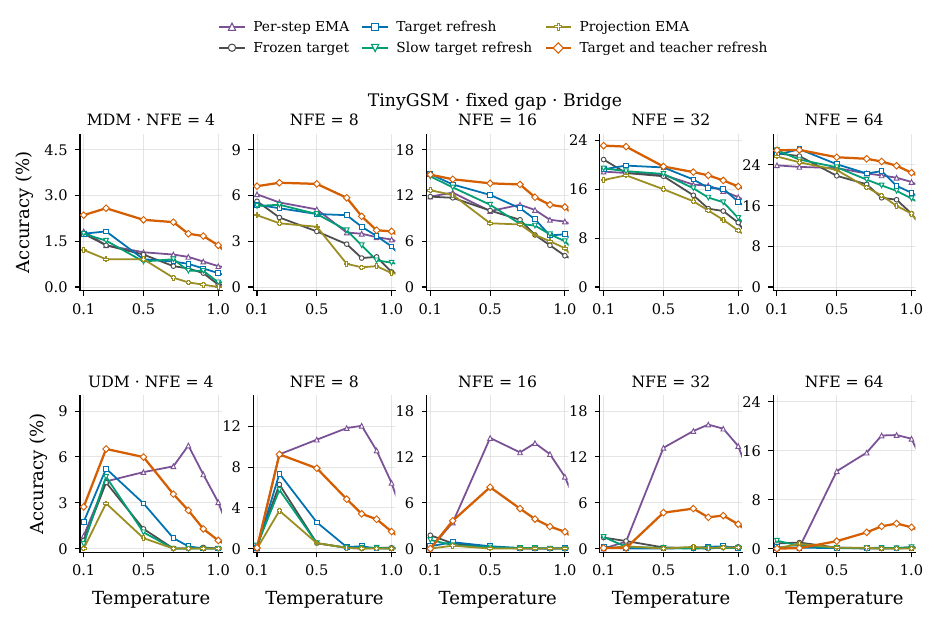}
    \caption{TinyGSM target-update policies with a fixed training gap, under bridge sampling.}
    \label{fig:ablation-policy-tinygsm-fixed-bridge}
    \medskip
    \includegraphics[width=0.88\linewidth]{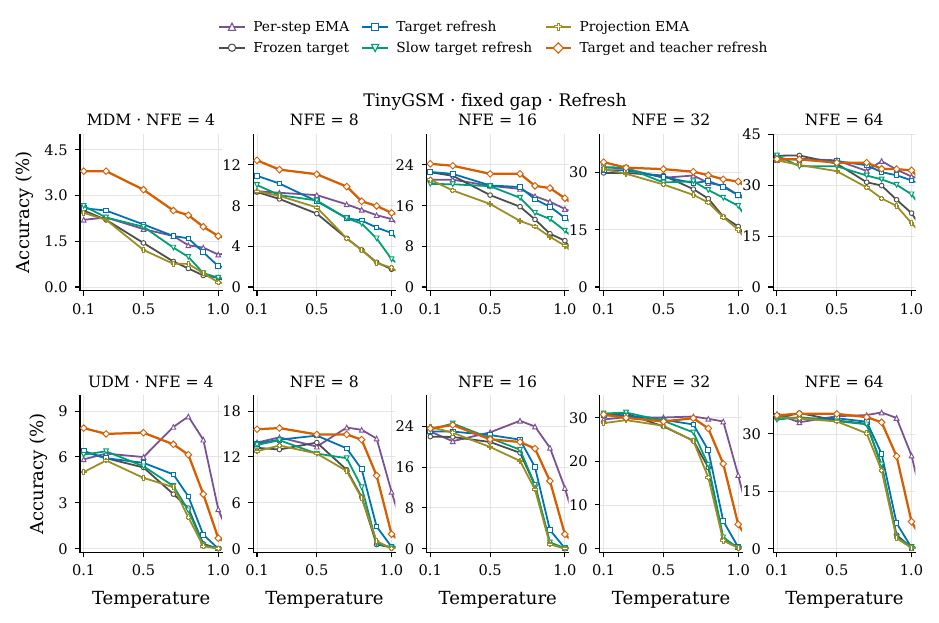}
    \caption{The same fixed-gap policies under refresh sampling.}
    \label{fig:ablation-policy-tinygsm-fixed-refresh}
\end{figure}

\begin{figure}[htbp]
    \centering
    \includegraphics[width=0.88\linewidth]{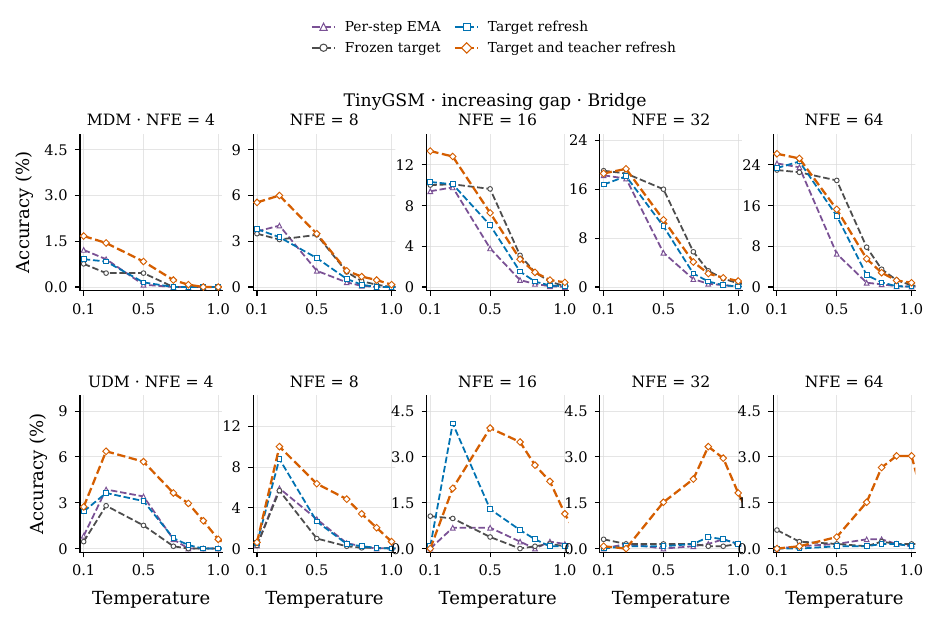}
    \caption{TinyGSM target-update policies with an increasing training gap, under bridge sampling.}
    \label{fig:ablation-policy-tinygsm-progressive-bridge}
    \medskip
    \includegraphics[width=0.88\linewidth]{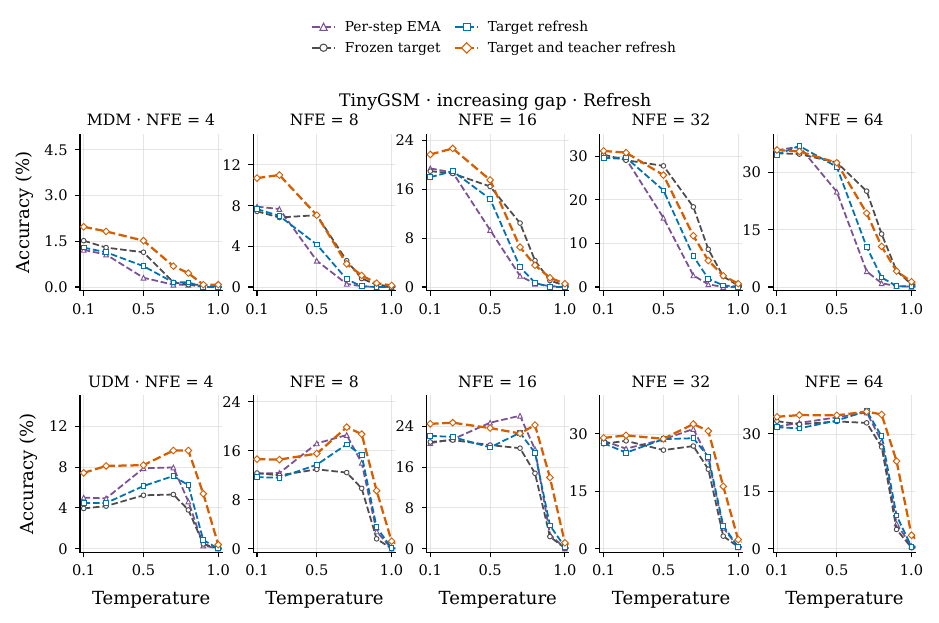}
    \caption{The same increasing-gap policies under refresh sampling.}
    \label{fig:ablation-policy-tinygsm-progressive-refresh}
\end{figure}

\begin{figure}[htbp]
    \centering
    \includegraphics[width=0.88\linewidth]{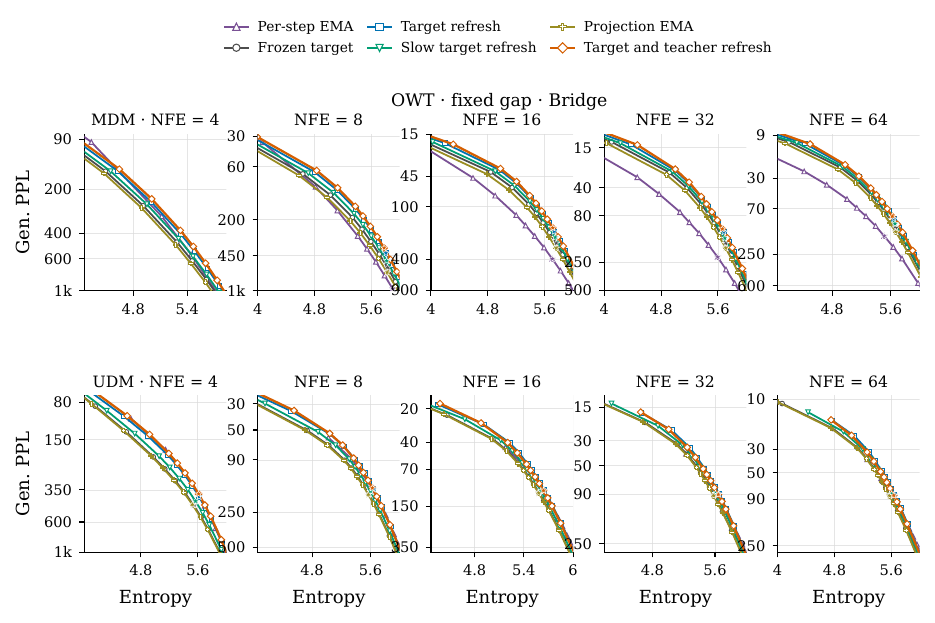}
    \caption{OWT target-update policies with a fixed training gap, under bridge sampling. Stars mark temperature $1$.}
    \label{fig:ablation-policy-owt-fixed}
    \medskip
    \includegraphics[width=0.88\linewidth]{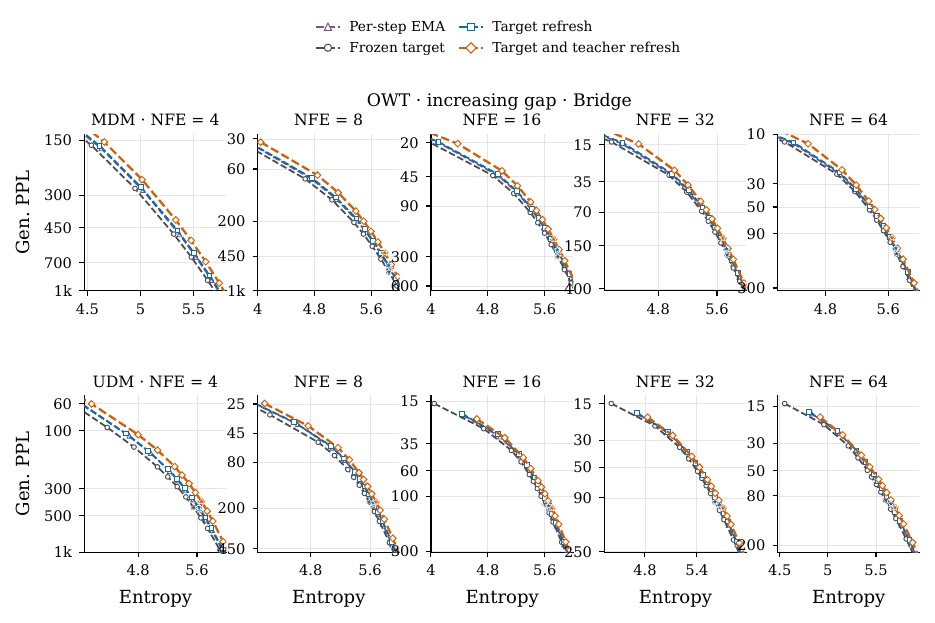}
    \caption{The same OWT policies with an increasing gap. Axes fit visible points within entropy $4$--$6$ and perplexity at most $1{,}000$.}
    \label{fig:ablation-policy-owt-progressive}
\end{figure}

\FloatBarrier
\subsection{Consistency gap}
\label{sec:distillation-gap-ablations}

\begin{figure}[htbp]
    \centering
    \includegraphics[width=0.94\linewidth]{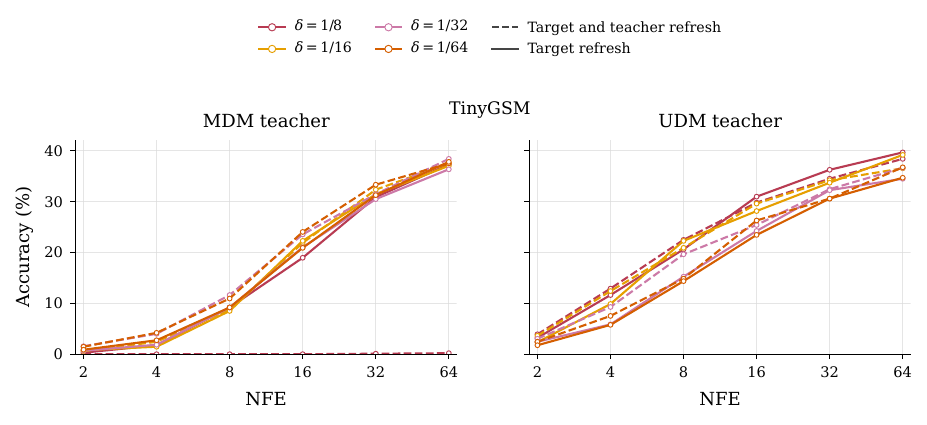}
    \caption{TinyGSM accuracy across training gaps and refresh policies. Solid curves refresh only the target; dashed curves also refresh the rollout teacher.}
    \label{fig:ablation-delta-tinygsm}
    \medskip
    \includegraphics[width=0.94\linewidth]{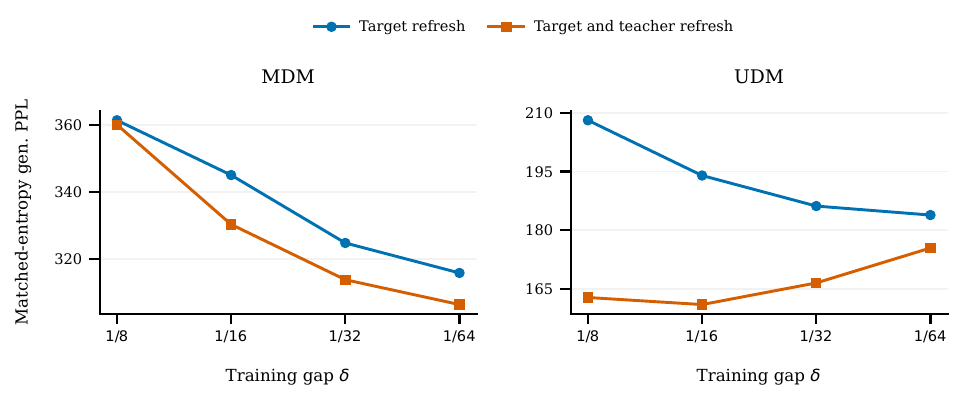}
    \caption{OWT matched-entropy generative perplexity, averaged across six NFEs, for each training gap and refresh policy.}
    \label{fig:ablation-delta-owt-summary}
\end{figure}

\begin{figure}[htbp]
    \centering
    \includegraphics[width=0.92\linewidth]{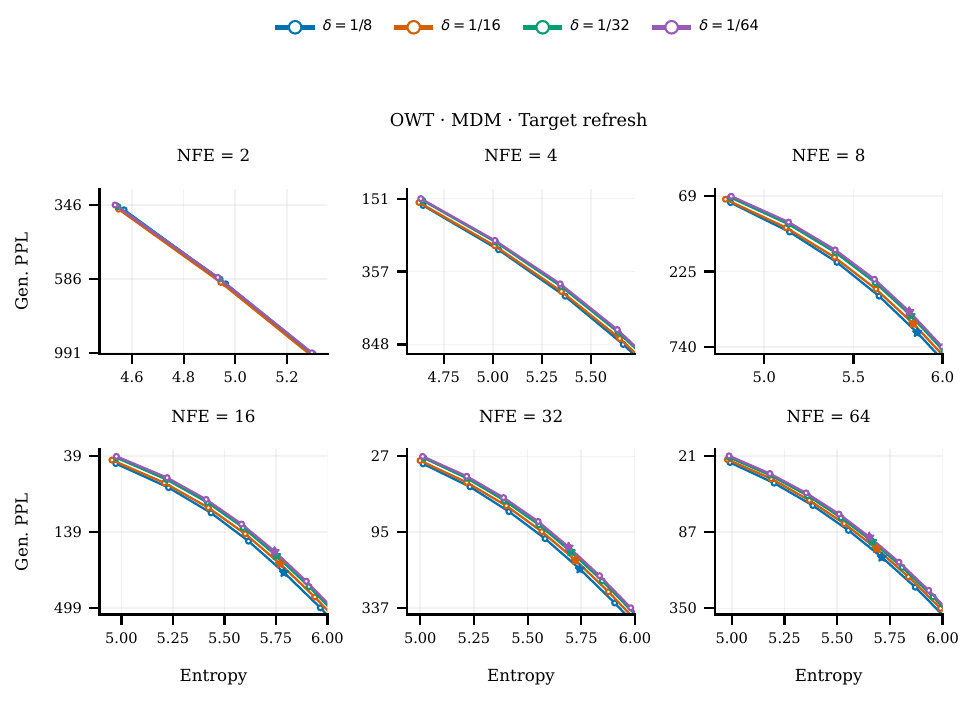}
    \caption{OWT MDM with target refresh: measured bridge-sampling temperature curves at each NFE. Stars mark temperature $1$.}
    \label{fig:ablation-delta-owt-mdm-target}
\end{figure}

\begin{figure}[htbp]
    \centering
    \includegraphics[width=0.92\linewidth]{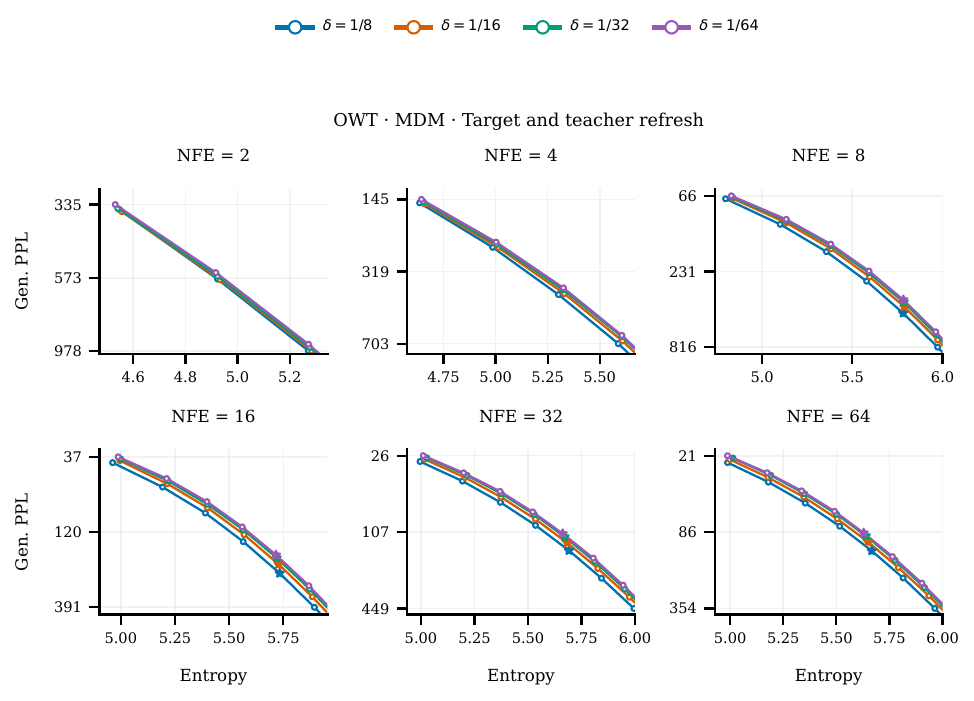}
    \caption{OWT MDM with joint target-and-teacher refresh, under the same evaluation grid as \Cref{fig:ablation-delta-owt-mdm-target}.}
    \label{fig:ablation-delta-owt-mdm-joint}
\end{figure}

\begin{figure}[htbp]
    \centering
    \includegraphics[width=0.92\linewidth]{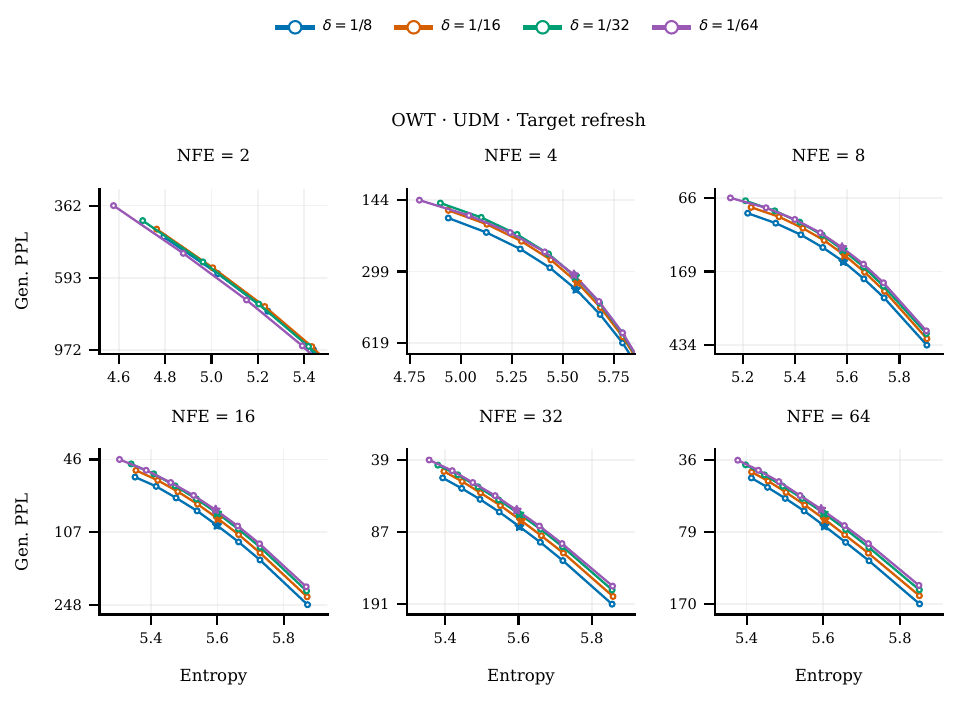}
    \caption{OWT UDM with target refresh. Axes show visible points within entropy $4$--$6$ and perplexity at most $1{,}000$.}
    \label{fig:ablation-delta-owt-udm-target}
\end{figure}

\begin{figure}[htbp]
    \centering
    \includegraphics[width=0.92\linewidth]{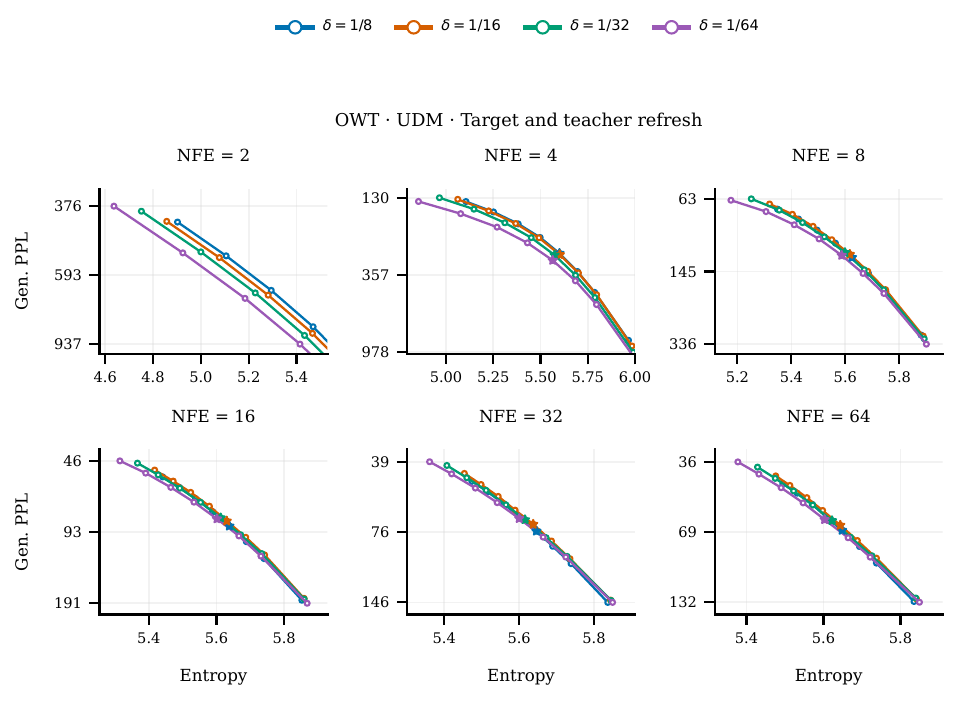}
    \caption{OWT UDM with joint target-and-teacher refresh, under the same evaluation grid as \Cref{fig:ablation-delta-owt-udm-target}.}
    \label{fig:ablation-delta-owt-udm-joint}
\end{figure}

We compare $\delta\in\{1/8,1/16,1/32,1/64\}$ with the target refreshed every round, either keeping the pretrained transport denoiser fixed or refreshing it from the same student EMA. For TinyGSM, each curve in \Cref{fig:ablation-delta-tinygsm} uses one sampler and temperature selected by mean accuracy across the six NFEs. For OWT, \Cref{fig:ablation-delta-owt-summary} averages log-perplexity over the entropy interval shared by all eight curves at each NFE, then across NFEs; lower scores correspond to lower matched-entropy perplexity. The measured OWT temperature curves appear in \Cref{fig:ablation-delta-owt-mdm-target,fig:ablation-delta-owt-mdm-joint,fig:ablation-delta-owt-udm-target,fig:ablation-delta-owt-udm-joint}.

\FloatBarrier
\subsection{Bridge and refresh sampling}
\label{sec:distillation-sampler-ablations}

We compare bridge sampling, which conditions the next state on the current state and the student's clean prediction, with refresh sampling, which re-noises that prediction and draws new residuals. For UDM, temperature is applied either to LOO probabilities before conversion or to denoising probabilities after conversion; MDM uses denoiser temperature. Each comparison includes the shared teacher, iDLM, D-MMD, SDTT for MDM or DCD for UDM, and ERC. ERC uses the selected gap in \Cref{sec:distillation-evaluation-details}, fixed for each dataset and diffusion family.

\begin{figure}[htbp]
    \centering
    \includegraphics[width=0.84\linewidth]{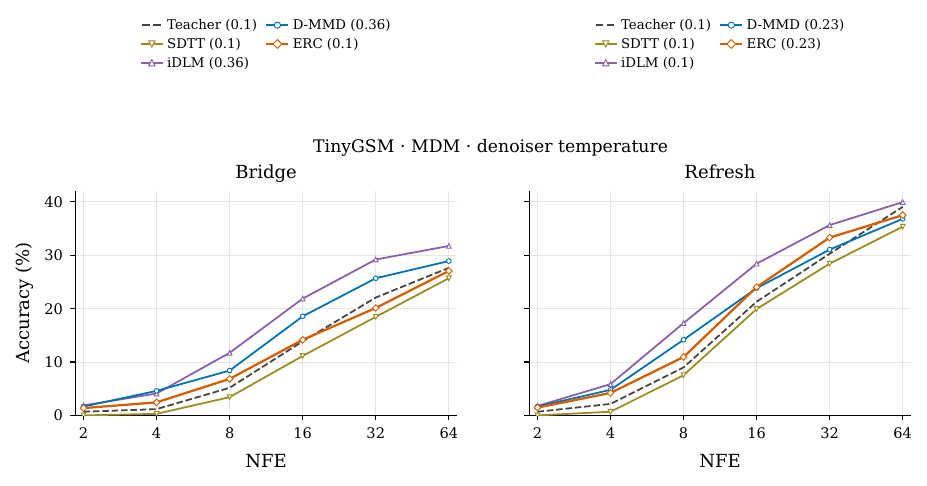}
    \caption{TinyGSM MDM accuracy versus NFE under bridge and refresh sampling. Each method retains the temperature selected by mean accuracy across six NFE budgets; the teacher, SDTT, iDLM, D-MMD, and ERC are included.}
    \label{fig:ablation-sampler-tinygsm-mdm-nfe}
    \medskip
    \includegraphics[width=0.84\linewidth]{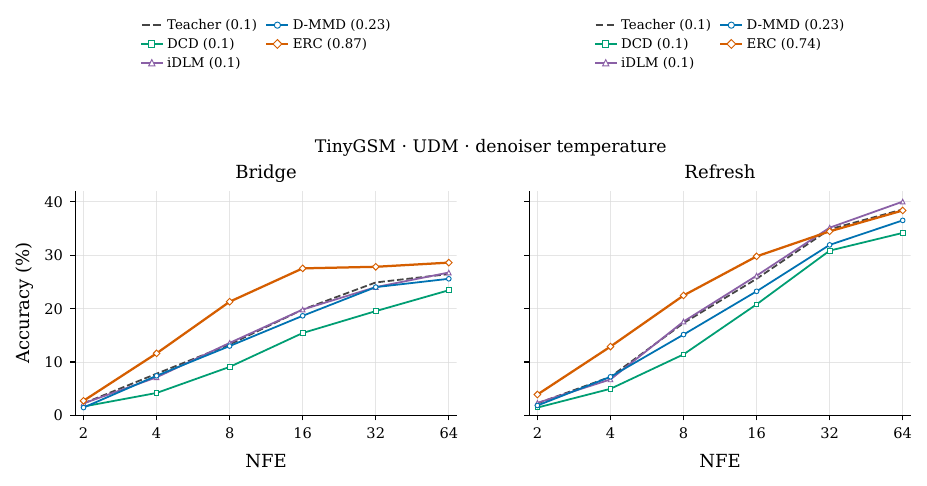}
    \caption{TinyGSM UDM accuracy versus NFE with temperature applied to denoiser predictions, comparing bridge and refresh sampling and all five methods.}
    \label{fig:ablation-sampler-tinygsm-udm-denoiser-nfe}
    \medskip
    \includegraphics[width=0.84\linewidth]{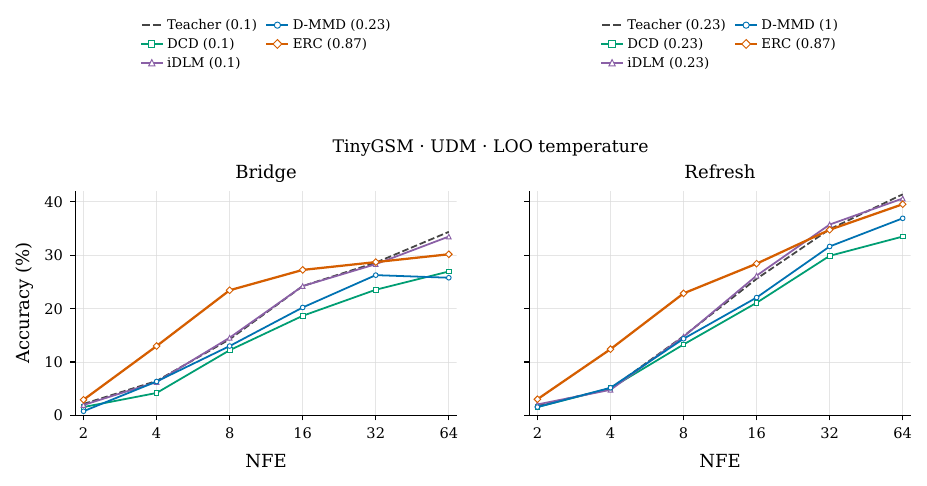}
    \caption{TinyGSM UDM accuracy versus NFE with temperature applied to LOO predictions. Bridge and refresh panels contain the teacher, iDLM, D-MMD, DCD, and ERC.}
    \label{fig:ablation-sampler-tinygsm-udm-loo-nfe}
\end{figure}

\begin{figure}[htbp]
    \centering
    \includegraphics[width=\linewidth,height=0.70\textheight,keepaspectratio]{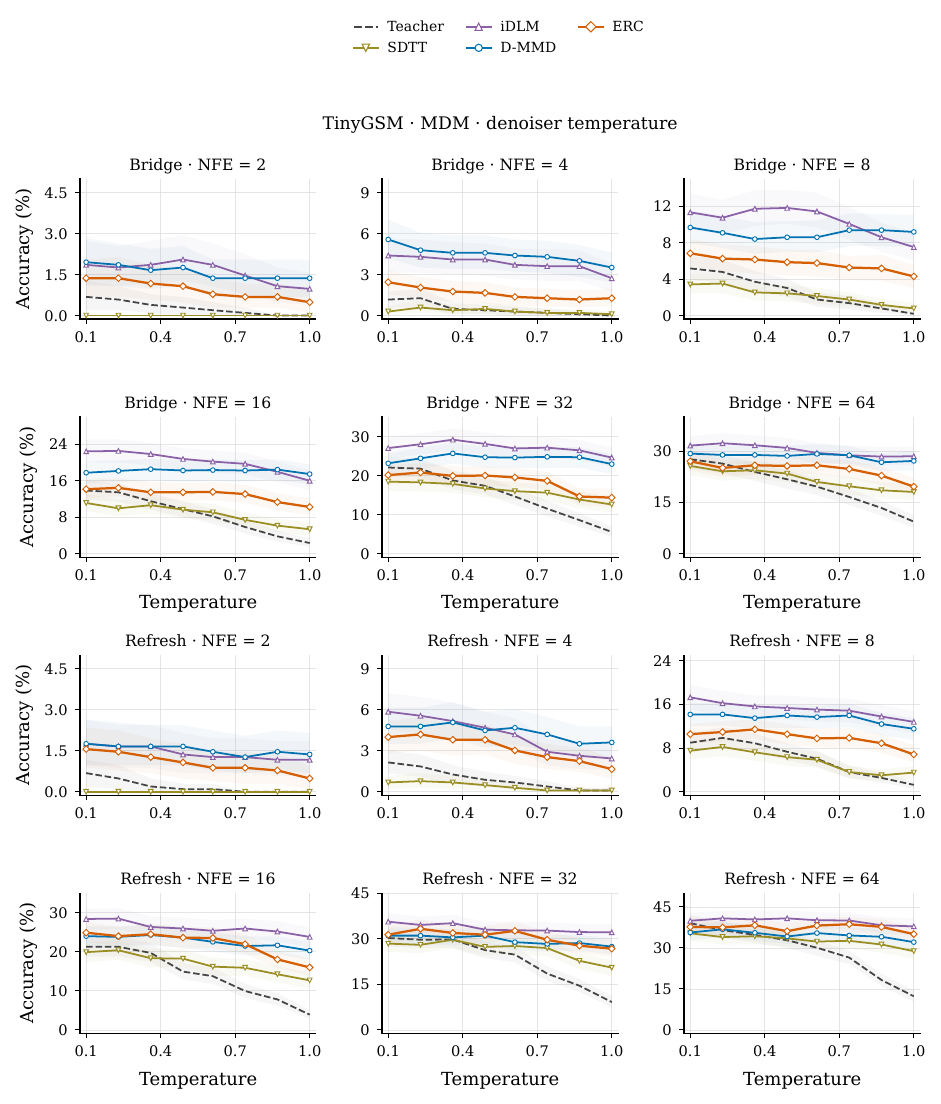}
    \caption{TinyGSM MDM accuracy across eight denoiser temperatures and six NFE budgets for bridge and refresh sampling. Shaded bands are pointwise $95\%$ bootstrap intervals over questions.}
    \label{fig:ablation-sampler-tinygsm-mdm-temperature}
\end{figure}

\begin{figure}[htbp]
    \centering
    \includegraphics[width=\linewidth,height=0.70\textheight,keepaspectratio]{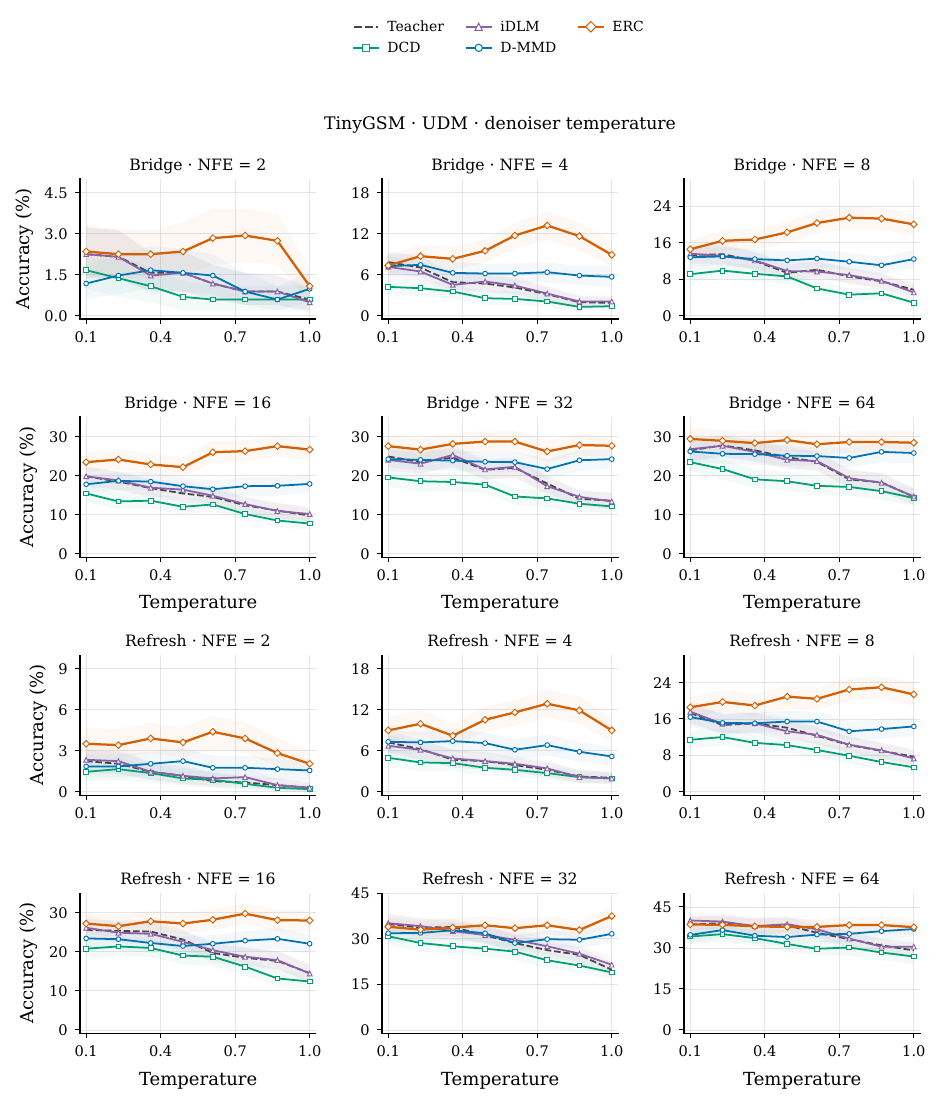}
    \caption{TinyGSM UDM denoiser-temperature sweeps for both samplers at all six NFE budgets. Bands give pointwise $95\%$ question-bootstrap intervals.}
    \label{fig:ablation-sampler-tinygsm-udm-denoiser-temperature}
\end{figure}

\begin{figure}[htbp]
    \centering
    \includegraphics[width=\linewidth,height=0.70\textheight,keepaspectratio]{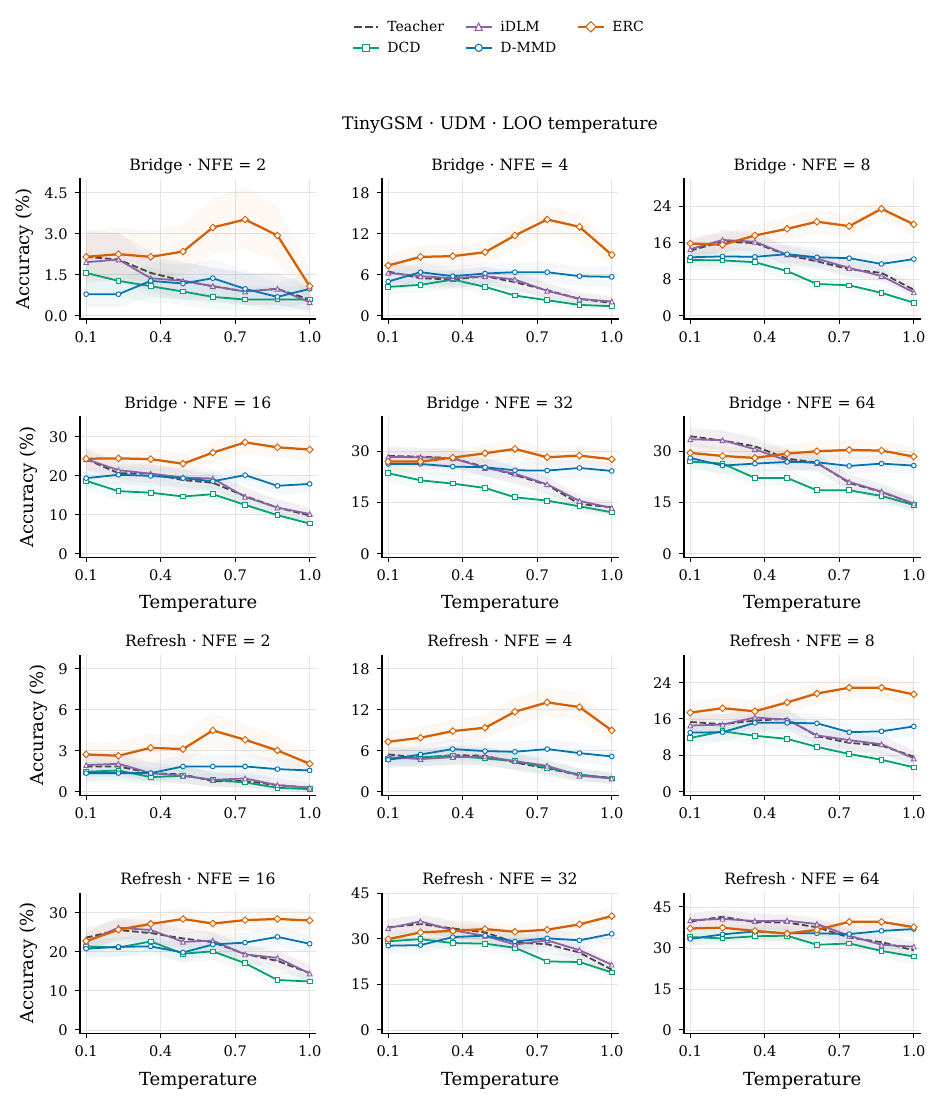}
    \caption{TinyGSM UDM LOO-temperature sweeps for bridge and refresh sampling. All eight temperatures and six NFE budgets are retained, with pointwise $95\%$ question-bootstrap intervals.}
    \label{fig:ablation-sampler-tinygsm-udm-loo-temperature}
\end{figure}

For TinyGSM, \Cref{fig:ablation-sampler-tinygsm-mdm-nfe,fig:ablation-sampler-tinygsm-udm-denoiser-nfe,fig:ablation-sampler-tinygsm-udm-loo-nfe} show accuracy against NFE with bridge and refresh in adjacent panels. Within each panel, one of the eight measured temperatures is selected separately for each method by its mean accuracy over the six NFE budgets; that temperature is printed in the legend and held fixed across NFEs. \Cref{fig:ablation-sampler-tinygsm-mdm-temperature,fig:ablation-sampler-tinygsm-udm-denoiser-temperature,fig:ablation-sampler-tinygsm-udm-loo-temperature} instead show all measured temperatures at every NFE for both samplers, with pointwise $95\%$ question-bootstrap bands. All points use the same $1{,}024$ GSM8K questions.

\begin{figure}[htbp]
    \centering
    \includegraphics[width=0.84\linewidth]{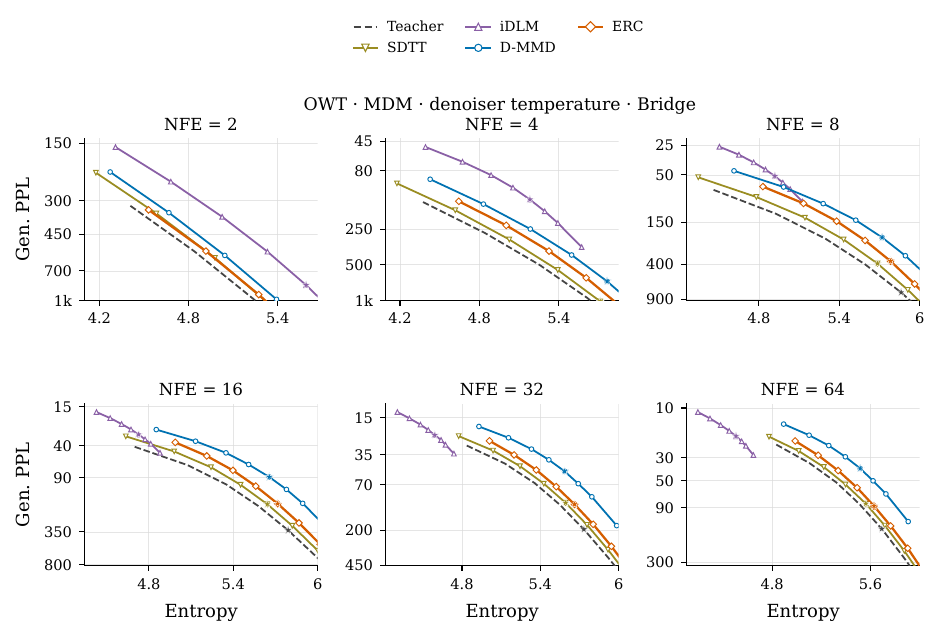}
    \caption{OWT MDM entropy--perplexity curves under bridge sampling.}
    \label{fig:ablation-sampler-owt-mdm-bridge}
    \medskip
    \includegraphics[width=0.84\linewidth]{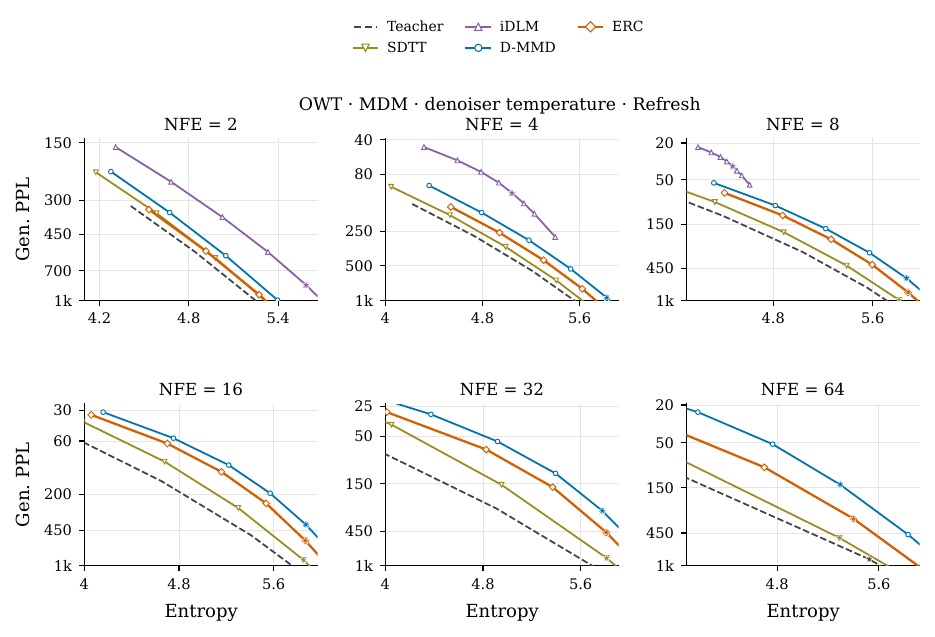}
    \caption{The same OWT MDM comparison under refresh sampling.}
    \label{fig:ablation-sampler-owt-mdm-refresh}
\end{figure}

\begin{figure}[htbp]
    \centering
    \includegraphics[width=0.84\linewidth]{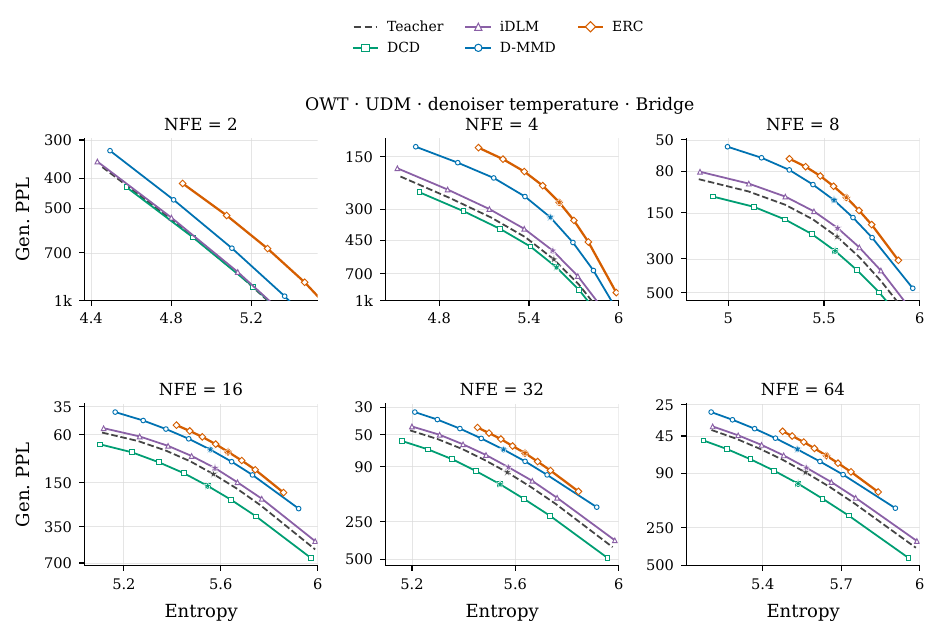}
    \caption{OWT UDM entropy--perplexity curves with denoiser temperature under bridge sampling.}
    \label{fig:ablation-sampler-owt-udm-denoiser-bridge}
    \medskip
    \includegraphics[width=0.84\linewidth]{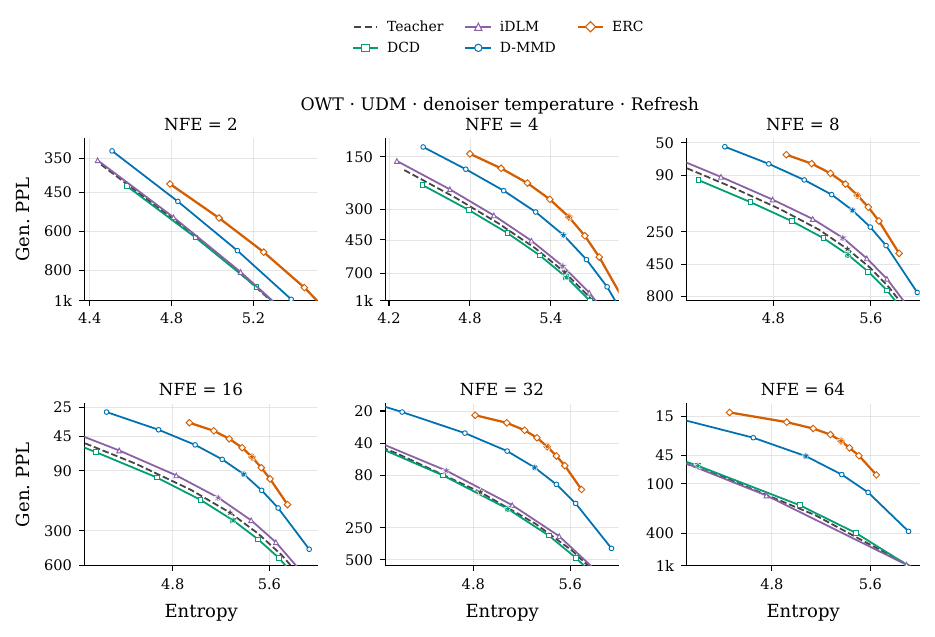}
    \caption{The same OWT UDM denoiser-temperature comparison under refresh sampling.}
    \label{fig:ablation-sampler-owt-udm-denoiser-refresh}
\end{figure}

\begin{figure}[htbp]
    \centering
    \includegraphics[width=0.84\linewidth]{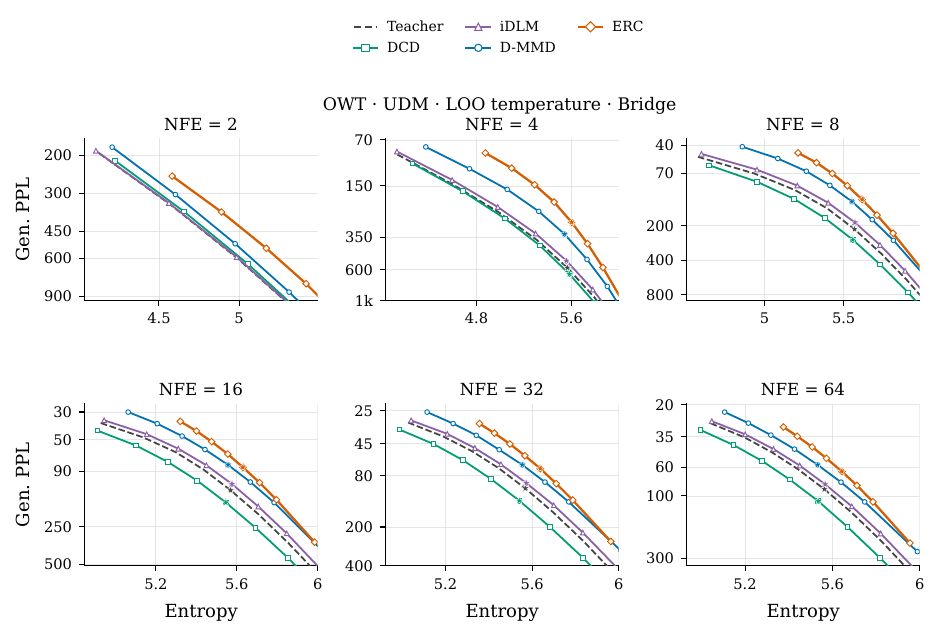}
    \caption{OWT UDM entropy--perplexity curves with LOO temperature under bridge sampling.}
    \label{fig:ablation-sampler-owt-udm-loo-bridge}
    \medskip
    \includegraphics[width=0.84\linewidth]{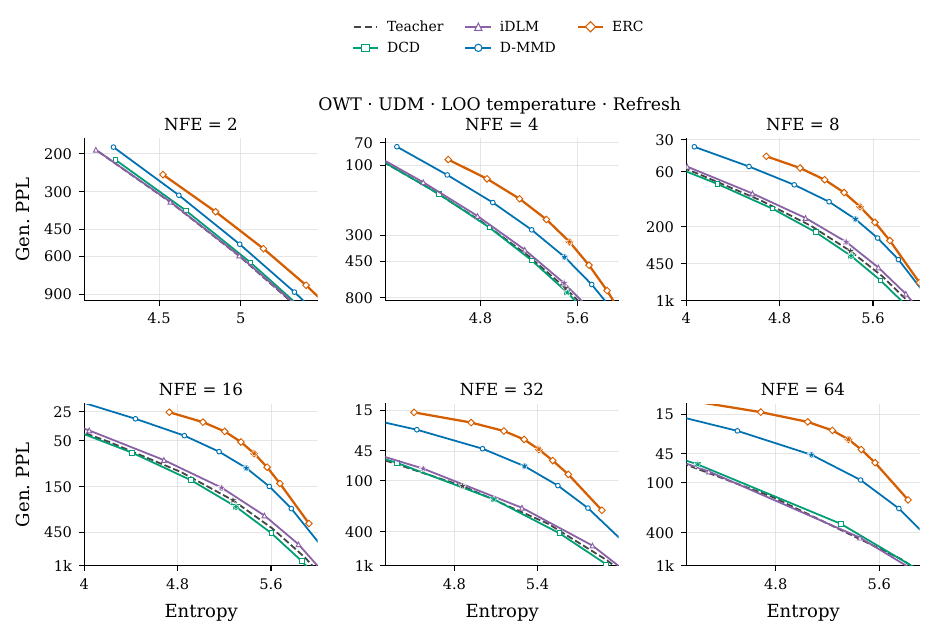}
    \caption{The same OWT UDM LOO-temperature comparison under refresh sampling.}
    \label{fig:ablation-sampler-owt-udm-loo-refresh}
\end{figure}

For OpenWebText, each panel of \Cref{fig:ablation-sampler-owt-mdm-bridge,fig:ablation-sampler-owt-mdm-refresh,fig:ablation-sampler-owt-udm-denoiser-bridge,fig:ablation-sampler-owt-udm-denoiser-refresh,fig:ablation-sampler-owt-udm-loo-bridge,fig:ablation-sampler-owt-udm-loo-refresh} fixes an NFE budget and sweeps the eight measured temperatures. Axes fit the visible points without extending below entropy $4$, above entropy $6$, or above generative perplexity $1{,}000$; points outside these bounds are not shown, and a star identifies temperature $1$ when visible. The perplexity axis is logarithmic and inverted. All points use $2{,}048$ generated sequences. Denoiser and LOO placement at temperature $1$ are aliases of one evaluation within each UDM sampler, not independent replicates.

\FloatBarrier
\subsection{Gumbel conditioning}
\label{sec:distillation-gumbel-ablations}

We compare ERC with and without conditioning on the Gumbel variables $-\log E_t$ after $10{,}000$ student updates. The matched MDM runs use a fixed training gap; the UDM runs increase the gap over four rounds. In both, the target and ERF transport denoiser are refreshed together. Removing Gumbel inputs does not change the ERF transport, including its net-flux rates for UDM. TinyGSM uses both samplers and averages sampling seeds $101$ and $303$; OWT uses bridge sampling with seed $101$. No matched OWT refresh comparison is available.

\begin{figure}[htbp]
    \centering
    \includegraphics[width=0.88\linewidth]{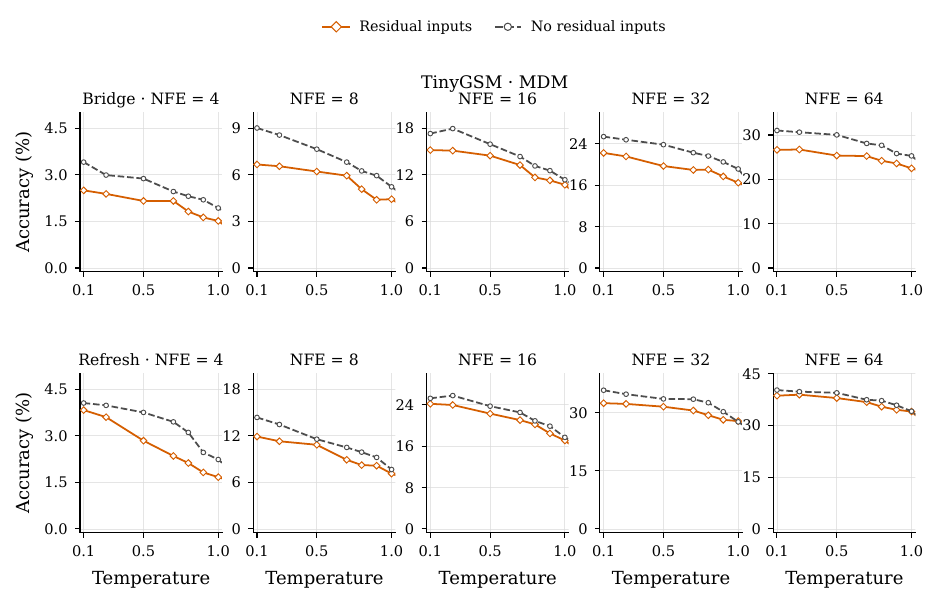}
    \caption{TinyGSM MDM with and without Gumbel inputs under bridge and refresh sampling. Points average two sampling seeds.}
    \label{fig:ablation-gumbel-tinygsm-mdm}
    \medskip
    \includegraphics[width=0.88\linewidth]{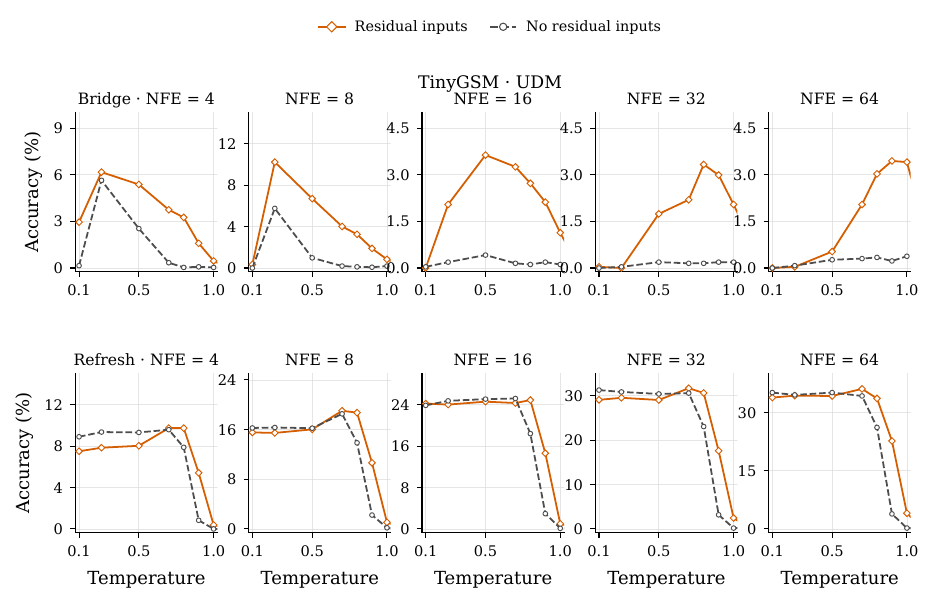}
    \caption{TinyGSM UDM under the same Gumbel-input comparison, with an increasing training gap.}
    \label{fig:ablation-gumbel-tinygsm-udm}
\end{figure}

\begin{figure}[htbp]
    \centering
    \includegraphics[width=\linewidth]{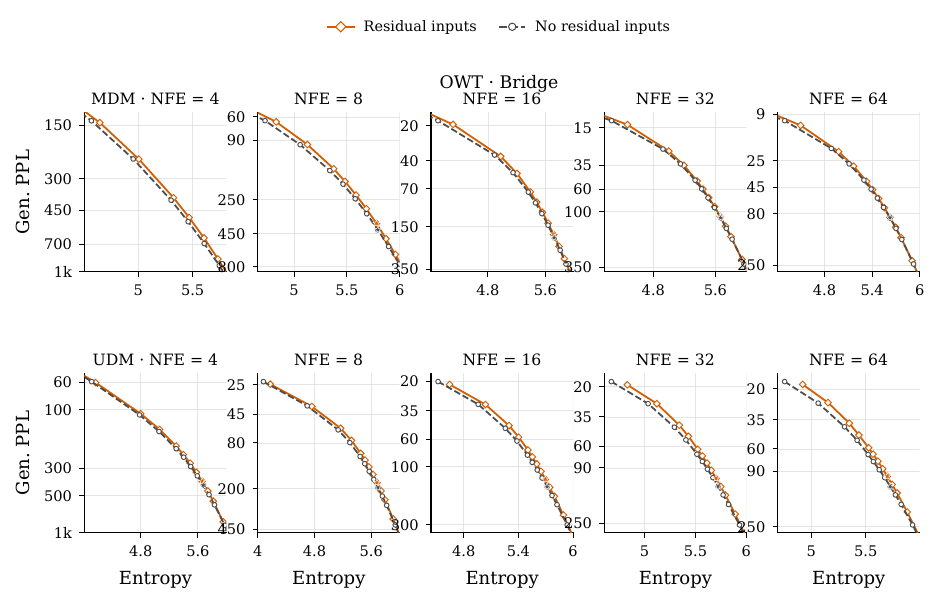}
    \caption{OWT Gumbel-input comparison for MDM and UDM under bridge sampling at five NFEs and fourteen temperatures.}
    \label{fig:ablation-gumbel-owt}
\end{figure}

%% file: appendix/diffusiongemma.tex
\section{DiffusionGemma experiment}
\label{app:diffusiongemma}

We evaluate the instruction-tuned $26$B-parameter, $4$B-active DiffusionGemma checkpoint \citep{gemma4diffusion2026blog} on the MMLU-Pro test split \citep{wang2024mmlu}. We select ten questions from each of its fourteen categories by a stable hash with seed $20260820$, yielding the same $140$ questions for run seeds $1$, $2$, and $3$. Each prompt requests a concise justification ending with \texttt{The answer is (X).}, and generation uses one canvas block of at most $256$ tokens. The strict evaluator accepts only this final-answer form and counts malformed or truncated completions as incorrect.

We compare the official native generation procedure with UDLM and the unbounded ERF Uniform and ERF U-NetFlux samplers. The custom samplers use $48$ NFEs, temperature $1$, and temperature scaling of the raw denoiser logits. The accuracy intervals in \Cref{tab:diffusiongemma-mmlu-pro} use $50{,}000$ category-stratified bootstrap replicates over question clusters while retaining all three paired seeds. The native procedure terminates adaptively after $11.1$ forward evaluations on average, whereas each custom sampler uses exactly $48$; the reported timings therefore describe the runs but do not constitute a compute-matched comparison. This MMLU-Pro evaluation is not directly comparable to evaluations on other benchmarks reported with the model release.

%% file: appendix/exponential_residual_flow_transcription.tex
\section{Proofs for the ERF sampler}
\label{sec:erf-transcription}

\subsection{Intuition on the ODEs \eqref{eq:exponential-residual-flow}}
\label{sec:erf-motivation}
To motivate the form of this ODE, we derive a continuity equation on the extended state space $\msx\times[0,\infty)^{\len\times\vocab}$ such that $X_t$ follows the Kolmogorov equation with $(\ratemat{t}{}{})_{t\in\ccint{0,1}}$. This will impose a velocity on the exponential clocks. Rewriting the Kolmogorov equation \eqref{eq:kolmogorov} and using \eqref{eq:reverse-rate}, we have for any $x$ and $t$:
\begin{equation}
    \label{eq:kolmogorov-split}
    \partial_t p_t(x)
    = p_t(x)\sum_{\ell=1}^{\len}
        \bigl(b_t^\ell(x)-\lambda_t^\ell(x^\ell)\bigr)\eqsp,
\end{equation}
where $b_t^\ell(x)\eqdef\sum_{j\in\msv\setminus\{x^\ell\}}\rratemat{t}{x}{j}[\ell]$ is the reverse exit rate at position $\ell$, and $\lambda_t^\ell(x^\ell)\eqdef\sum_{j\in\msv\setminus\{x^\ell\}}\ratemat{t}{x^\ell}{j}[\ell]$ is the forward exit rate.
Therefore, introducing the vector field (depending on the discrete state): $v_t(x) \in \rset^{\len \times \vocab}$ with components
\begin{equation}
    \label{eq:vector-field}
    v_t^{\ell j}(x) = \begin{cases}
        \rratemat{t}{x}{j}[\ell], & j \neq x^\ell,\\
        -\lambda_t^{\ell}(x^\ell), & j = x^\ell\eqsp,
    \end{cases}
\end{equation}
and the extended density $\mu_t(x,e) = p_t(x) \exp(-\sum_{\ell=1}^{\len} \sum_{j=1}^{\vocab} e^{\ell j})$ on $\msx \times \coint{0,\infty}^{\len\times\vocab}$, we can rewrite the Kolmogorov equation \eqref{eq:kolmogorov-split} as
\begin{equation}
    \label{eq:kolmogorov-continuity}
    \partial_t \mu_t(x,e) = - \nabla_{e} \cdot (v_t(x) \mu_t(x,e))\eqsp,
\end{equation}
where $\nabla_e \cdot$ denotes the divergence operator with respect to $e$.
This equation can be interpreted as a continuity equation on the extended state space, and suggests the dynamics we have imposed on $E_t$, namely \eqref{eq:exponential-residual-flow}.

\subsection{Setting}
\subsubsection{Notation and assumptions}
\textbf{Notation.}  By convention, we set $\prod_{i}^j = 1$ if $i>j$. In nonnegative integrands, $0\cdot(+\infty)$ is defined to be zero. Position indices are $\ell,h$, token indices are $j,k$, and $n,m$ count jumps or partial sums.

We recall that the we fix the state space $\msx = \msv^{\len}$ for some $\len\geq1$ and set the set of tokens here as $\msv=\{\representer{1},\ldots,\representer{\vocab}\}$, $\msx=\msv^{\len}$. We denote for ease of notation $D=\len\vocab$. Vectors in $\rset^D$ are indexed by $\intset{1}{\len}\times\intset{1}{\vocab}$ with components at top in lexicographic order. For any $x \in\msx$, $\ell \in \{1,\ldots,\len\}$ and $j \in\msv$, we denote by $x^{-\ell}$ the vector $(x^1,\ldots,x^{\ell-1},x^{\ell+1},\ldots,x^{\len})$ and write 
$x^{\ell \to j}$ as the element $y$ such that $y^{-\ell} = x^{-\ell}$ and $y^{\ell} = j$. 

\textbf{Setting and assumptions.}
We consider now an inhomogeneous rate matrix denoted by $(\ratemat{v}{}{})_{v\in[0,1]}$, \ie, satisfying for any $v \in \ccint{0,1}$, $\ratemat{v}{\bx}{\by}\geq0$ for any $\bx\neq\by$ and
\begin{equation}
	\ratemat{v}{\bx}{\bx} = - \lambda_v(\bx) = - \sum_{\by \neq \bx} \ratemat{v}{\bx}{\by}\eqsp.
\end{equation}
We denote by $\pdata{v}{}{}$ the marginal distribution of the corresponding continuous-time Markov chain (CTMC) at time $v$. 
In addition, we make the following assumption on the rates and the marginals of the corresponding CTMC.

\begin{assumption}
\label{ass:erf-tr-standing}
\begin{renumerate}
\item There exist $\len$ time-inhomogeneous rate matrices on $\msv$, $\{(\ratemat{v}{}{}[\ell])_{v\in[0,1]}\,: \, \ell \in \{1,\ldots,\len\}\}$ such that for any $\bx,\by \in \msx$
\begin{equation}
        \label{eq:erf-tr-rates}
	\ratemat{v}{\bx}{\by} = \sum_{\ell=1}^{\len} \ratemat{v}{\bx^\ell}{\by^\ell}[\ell] \indic_{\{\bx^{-\ell} = \by^{-\ell}\}} \eqsp.
\end{equation}
\item   
For every $\ell$ and distinct $x^{\ell},\tilde{x}^{\ell}\in\msv$, the map $v\mapsto\ratemat{v}{\tilde{x}^{\ell}}{x^{\ell}}[\ell]$ is nonnegative and continuous on $[0,1]$.
\item There is no explosion, \ie, $\sum_{\bx \in \msx} \pdata{v}{}{\bx} = 1$ for any $v\in\ccint{0,1}$.
\item For any $v\in[0,1]$ and $\bx\in\msx$, it holds $\pdata{v}{}{\bx}>0$.
\item  For every $\bx\in\msx$, the map $v\mapsto\pdata{v}{}{\bx}$ is continuously differentiable on $\ccint{0,1}$ and  it holds for any $v\in(0,1)$:
\begin{align}
    \partial_v\pdata{v}{}{\bx}
    =\sum_{\ell=1}^{\len}
      \sum_{\substack{\by^{-\ell}=\bx^{-\ell}\\\by^\ell\neq\bx^\ell}}
         \pdata{v}{}{\by}\ratemat{v}{\by^\ell}{\bx^\ell}[\ell]
      -\pdata{v}{}{\bx}\lambda_v(\bx).
    \label{eq:erf-tr-forward}
\end{align}
The derivatives at $0$ and $1$ are one-sided.
\end{renumerate}
\end{assumption}

Under \Cref{ass:erf-tr-standing}, denote for any $\ell \in \{1,\ldots,\len\}$ and $\bx^{\ell}\in\msv$:
\begin{align}
    \lambda_v^\ell(\bx^\ell)
    =\sum_{j:\representer{j}\neq\bx^\ell}\ratemat{v}{\bx^\ell}{\representer{j}}[\ell] \eqsp,
    \label{eq:erf-tr-lambda}
\end{align}
and note that with this notation, $\lambda_v(\bx)=\sum_{\ell=1}^{\len}\lambda_v^\ell(\bx^\ell)$ and $\ratemat{v}{\bx}{\bx}=-\lambda_v(\bx)$. 

For $v\in[0,1]$ and distinct $\bx,\by\in\msx$, define the reverse rates by
\begin{align}
    \rratemat{v}{\bx}{\by}
    =\frac{\pdata{v}{}{\by}}{\pdata{v}{}{\bx}}\ratemat{v}{\by}{\bx}.
    \label{eq:erf-tr-reverse-full}
\end{align}
Under \Cref{ass:erf-tr-standing}, the reverse rates are well-defined and satisfy, for distinct $\bx,\by$,
\begin{align}
    \rratemat{v}{\bx}{\by}
    &=\sum_{\ell \in \{1,\ldots,\len\}}
        \rratemat{v}{\bx}{\by^\ell}[\ell]
        \indic_{\{\bx^{-\ell}=\by^{-\ell}\}}\eqsp,\notag\\
    \rratemat{v}{\bx}{\representer{j}}[\ell]
    &=\frac{\pdata{v}{}{\by}}{\pdata{v}{}{\bx}}
        \ratemat{v}{\representer{j}}{\bx^\ell}[\ell]\eqsp.
    \label{eq:erf-tr-reverse}
\end{align}
In the second line, $\by^{-\ell}=\bx^{-\ell}$ and $\by^\ell=\representer{j}$.
Write $\bar\lambda_v(\bx)=\sum_{\ell=1}^{\len}\sum_{j:\representer{j}\neq\bx^\ell}\rratemat{v}{\bx}{\representer{j}}[\ell]$ and $\rratemat{v}{\bx}{\bx}=-\bar\lambda_v(\bx)$.

\subsubsection{ERF sampler revisited}
\label{sec:app:erf-tr-sampler}
We complete the description of the residual dynamics, described in \Cref{app:ctmc-gillespie}, mainly introducing new notations. 
For $\bx\in\msx$ and $v\in[0,1]$, define the vector field on $\msx \to \rset^D$ with components defined for any $\bx \in\msx$, $\ell\in\{1,\ldots,\len\}$ and $j\in\msv$ as 
\begin{equation}
       \label{eq:erf-tr-v}
     b_v^{\ell j}(\bx)= -\rratemat{v}{\bx}{\representer{j}}[\ell] \indic_{\{\representer{j}\neq\bx^\ell\}} + \lambda_v^\ell(\bx^\ell) \indic_{\{\representer{j}=\bx^\ell\}}  \eqsp.
 \end{equation}
In addition for $0\leq u\leq v\leq1$, define
\begin{align}
    \Delta_{s\tbar t}(\bx)=\int_s^t b_u(\bx)\,\rmd u\eqsp.
    \label{eq:erf-tr-D}
\end{align}
If the state remains $\bx$ from $v$ down to $u$, then
\begin{align}
    E_u=E_v+\Delta_{u\tbar v}(\bx).
    \label{eq:erf-tr-clocks}
\end{align}
For $v>0$, $\bx\in\msx$, and $\eta\in[0,\infty)^D$, define, when $\representer{j}\neq\bx^\ell$,
\begin{equation}
    \theta^{\ell j}(v,\bx,\eta)
    =\sup\left\{u\in[0,v]:
        \int_u^v\rratemat{\tau}{\bx}{\representer{j}}[\ell]\,\rmd\tau
        \geq\eta^{\ell j}\right\}\eqsp.
    \label{eq:erf-tr-candidate-time}
\end{equation}
Set $\sup\emptyset=-\infty$ and $\theta^{\ell j}(v,\bx,\eta)=-\infty$ when $\representer{j}=\bx^\ell$.

Set $T_0^\star=1$, draw $X_1\sim p_1$ and $E_1^{\ell j}\simiid\operatorname{Exp}(1)$ independently, and set $(X_{T_0^\star},E_{T_0^\star})=(X_1,E_1)$. Thus
\begin{align}
    \law(X_1,E_1)=\pdata{1}{}{}\otimes\operatorname{Exp}(1)^{\otimes D}.
    \label{eq:erf-tr-init}
\end{align}
Suppose that the first $n$ steps are defined and $(T_n^\star,X_{T_n^\star},E_{T_n^\star})=(v,\bx,\eta)$ with $v>0$. Let $\Theta_{n+1}=\max_{\ell,j}\theta^{\ell j}(v,\bx,\eta)$. If $\Theta_{n+1}=-\infty$, extend the path to $0$ without another jump. If
\begin{equation}
    0<\Theta_{n+1}<v,\qquad
    \argmax_{(\ell,j)}\theta^{\ell j}(v,\bx,\eta)=\{(h,k)\},\qquad
    \rratemat{\Theta_{n+1}}{\bx}{\representer{k}}[h]>0,
    \label{eq:erf-tr-contact-conditions}
\end{equation}
set
\begin{equation}
    \begin{split}
        T_{n+1}^\star&=\Theta_{n+1},\qquad C_{n+1}=(h,k),\qquad Y_{n+1}=\bx,\\
        E_{T_{n+1}^\star}&=\eta+\Delta_{T_{n+1}^\star\tbar v}(\bx),\qquad
        X_{T_{n+1}^\star}=\by,\quad
        \by^h=\representer{k},\quad \by^\ell=\bx^\ell\ \text{for }\ell\neq h.
    \end{split}
    \label{eq:erf-tr-recursive-path}
\end{equation}
Here $Y_{n+1}$ is the state before the jump and $C_{n+1}$ is its residual index. At any other finite candidate time, stop and record a failure, including a hit at time $0$. On the segment before the next candidate time, set
\begin{equation}
    X_u=X_{T_n^\star},\qquad
    E_u=E_{T_n^\star}+\Delta_{u\tbar T_n^\star}(X_{T_n^\star}),
    \qquad \Theta_{n+1}<u\leq T_n^\star.
    \label{eq:erf-tr-segment-path}
\end{equation}
If a successful step does not exist, set its time and all subsequent step times to $-\infty$; assign fixed values to the other variables after stopping. The proof below shows that failures and infinite sequences of jumps have probability zero.
Let $N_{\infty}=\sup\{n\geq0:T_n^\star\geq0\}$ denote the total number of jumps.

\subsection{Main result}
Let $\mathcal O=(0,\infty)^D$\alain{see if we will keep it...}. For $r\in[0,\infty)^D$ and $u\in[0,1]$, put
\begin{align}
    \gamma(r)=\exp\!\left(-\sum_{\ell,j}r^{\ell j}\right),
    \qquad F_u(\bx,r)=\pdata{u}{}{\bx}\gamma(r).
    \label{eq:erf-tr-F}
\end{align}
Lebesgue measure on $\mathcal O$ is denoted by $\rmd r$.

\begin{theorem}
\label{thm:erf-tr-main}
Assume \Cref{ass:erf-tr-standing}, define the reverse rates by \eqref{eq:erf-tr-reverse-full}--\eqref{eq:erf-tr-reverse}, and consider the EFR sampler described in \Cref{sec:app:erf-tr-sampler}.
Then, almost surely, $N_{\infty} < \infty$ and therefore the dynamics is well defined on $[0,1]$, with finitely many jumps. For every deterministic $u\in[0,1]$,
\begin{equation}
    \law(X_u,E_u)=\pdata{u}{}{}\otimes\operatorname{Exp}(1)^{\otimes D}.
    \label{eq:erf-tr-theorem}
\end{equation}
In addition, every $u \in \ccint{0,1}$, $\PP(T_n^{\star}=u \text{ for some } n)= 0$.
\end{theorem}

For $n\geq0$ and $u\in[0,1)$, define
\begin{align}
    H_u^{(n)}=\{
        \ T^{\star}_n>u,\ T^{\star}_{n+1}<u\}.
    \label{eq:erf-tr-Htk}
\end{align}
 For each Borel set $\mathcal{B}\subseteq[0,\infty)^D$, set
\begin{align}
     \mu_u^{(n)}(\{\bx\}\times \mathcal{B})
    =\pE\!\left[\indic_{H_u^{(n)}}\indic_{\{X_{T^{\star}_n}=\bx\}}
         \indic_{\{E_{T_n^{\star}}+\Delta_{u\tbar T^{\star}_n}(X_{T^{\star}_n})\in \mathcal{B}\}}\right].
    \label{eq:erf-tr-mu-k}
\end{align}
We prove the theorem by constructing densities $f_u^{(n)}$ of these measures and summing them. Initialization \eqref{eq:erf-tr-init} is not conditioning on a fixed state. The conclusion concerns deterministic times; it imposes no independence at jump times and does not assert that $X$ is Markovian.

\subsection{Preliminaries}
For $\bx\in\msx$, $\ell\in\intset{1}{\len}$, $\representer{j}\neq\bx^\ell$, and $0\leq u\leq v\leq1$, define
\begin{align}
    A_{u\tbar v}^\ell(\bx)=\int_u^v\lambda_\tau^\ell(\bx^\ell)\,\rmd\tau,
    \qquad B_{u\tbar v}^{\ell j}(\bx)
    =\int_u^v\rratemat{\tau}{\bx}{\representer{j}}[\ell]\,\rmd\tau.
    \label{eq:erf-tr-AB}
\end{align}
Consequently,
\begin{align}
    \Delta_{u\tbar v}^{\ell j}(\bx)=
    \begin{cases}
        A_{u\tbar v}^\ell(\bx),&\representer{j}=\bx^\ell,\\
        -B_{u\tbar v}^{\ell j}(\bx),&\representer{j}\neq\bx^\ell.
    \end{cases}
    \label{eq:erf-tr-D-components}
\end{align}

\begin{lemma}
\label{lem:erf-tr-balance}
Under \Cref{ass:erf-tr-standing}, all forward and reverse off-diagonal rates and all exit rates are continuous and bounded on $[0,1]$. For $v\in[0,1]$, $\bx\in\msx$, a position $\ell$, and $\by$ with $\by^{-\ell}=\bx^{-\ell}$ and $\by^\ell\neq\bx^\ell$,
\begin{align}
    \partial_v\log\pdata{v}{}{\bx}
       &=\bar\lambda_v(\bx)-\lambda_v(\bx),\label{eq:erf-tr-logp}\\
    \pdata{v}{}{\by}\rratemat{v}{\by}{\bx^\ell}[\ell]
       &=\pdata{v}{}{\bx}\ratemat{v}{\bx^\ell}{\by^\ell}[\ell],\label{eq:erf-tr-edge}\\
    \sum_{\substack{\by^{-\ell}=\bx^{-\ell}\\\by^\ell\neq\bx^\ell}}
       \pdata{v}{}{\by}\rratemat{v}{\by}{\bx^\ell}[\ell]
       &=\pdata{v}{}{\bx}\lambda_v^\ell(\bx^\ell).\label{eq:erf-tr-face}
\end{align}
In \eqref{eq:erf-tr-logp}, endpoint derivatives are one-sided. In particular,
\begin{align}
    \rratemat{v}{\by}{\bx^\ell}[\ell]>0
    \quad\Longrightarrow\quad\lambda_v^\ell(\bx^\ell)>0.
    \label{eq:erf-tr-positive-inward}
\end{align}
For $0\leq u\leq v\leq1$, $\bx\in\msx$, and $r,w\in[0,\infty)^D$ with $r=w+\Delta_{u\tbar v}(\bx)$,
\begin{align}
    \pdata{v}{}{\bx}\gamma(w)=\pdata{u}{}{\bx}\gamma(r).
    \label{eq:erf-tr-transport}
\end{align}
\end{lemma}
\begin{proof}
Since $\msx$ is finite and $\pdata{v}{}{\bx}$ is continuous and positive on $[0,1]$, the number $p_* =\min_{v\in[0,1],\bx\in\msx}\pdata{v}{}{\bx}$ is positive. Boundedness and continuity of the forward rates, of \eqref{eq:erf-tr-reverse}, and of their finite sums follow. Applying \eqref{eq:erf-tr-reverse} to the transition from $\by$ into $\bx$ gives
\begin{align*}
    \rratemat{v}{\by}{\bx^\ell}[\ell]
    =\frac{\pdata{v}{}{\bx}}{\pdata{v}{}{\by}}\ratemat{v}{\bx^\ell}{\by^\ell}[\ell].
\end{align*}
This proves \eqref{eq:erf-tr-edge}; summing over the predecessor states gives \eqref{eq:erf-tr-face}. Nonnegativity of the summands and positivity of $\pdata{v}{}{\bx}$ imply \eqref{eq:erf-tr-positive-inward}. Dividing \eqref{eq:erf-tr-forward} by $\pdata{v}{}{\bx}$ and using \eqref{eq:erf-tr-reverse} proves \eqref{eq:erf-tr-logp}.

For $0\leq u\leq v\leq1$, equations \eqref{eq:erf-tr-v} and \eqref{eq:erf-tr-logp} give
\begin{align}
    \begin{split}
        \sum_{\ell,j}\Delta_{u\tbar v}^{\ell j}(\bx)
        &=\int_u^v\bigl(\lambda_\tau(\bx)-\bar\lambda_\tau(\bx)\bigr)\,\rmd\tau\\
        &=-\int_u^v\partial_\tau\log\pdata{\tau}{}{\bx}\,\rmd\tau
         =\log\frac{\pdata{u}{}{\bx}}{\pdata{v}{}{\bx}}.
    \end{split}
    \label{eq:erf-tr-sum-D}
\end{align}
Since $w=r-\Delta_{u\tbar v}(\bx)$, substitution in \eqref{eq:erf-tr-F} yields
\begin{align*}
    \pdata{v}{}{\bx}\gamma(w)
    =\pdata{v}{}{\bx}\gamma(r)
       \exp\!\left(\sum_{\ell,j}\Delta_{u\tbar v}^{\ell j}(\bx)\right)
    =\pdata{u}{}{\bx}\gamma(r).
\end{align*}
\end{proof}

\begin{lemma}
\label{lem:erf-tr-calculus}
Let $\alpha<\beta$ be finite, let $\omega:[\alpha,\beta]\to[0,\infty)$ be continuous, and set $\Psi(v)=\int_\alpha^v\omega(\tau)\,\rmd\tau$ and $\rho(\zeta)=\inf\{v\in[\alpha,\beta]:\Psi(v)\geq\zeta\}$ for $0<\zeta<\Psi(\beta)$.

\textup{(i)} The set $\Psi(\{v\in[\alpha,\beta]:\omega(v)=0\})$ has one-dimensional Lebesgue measure zero. For almost every $\zeta\in(0,\Psi(\beta))$, the equation $\Psi(v)=\zeta$ has exactly one solution; it lies in $(\alpha,\beta)$ and has $\omega(v)>0$. More generally, for every $v_0\in(\alpha,\beta)$ with $\omega(v_0)>0$, one has $\Psi^{-1}(\{\Psi(v_0)\})=\{v_0\}$.

\textup{(ii)} For every nonnegative Borel $\chi$ on $[0,\Psi(\beta)]$,
\begin{align}
    \int_\alpha^\beta\chi(\Psi(v))\omega(v)\,\rmd v
    =\int_0^{\Psi(\beta)}\chi(\zeta)\,\rmd\zeta.
    \label{eq:erf-tr-substitution}
\end{align}
For every nonnegative Borel $\varphi$ on $[\alpha,\beta]$,
\begin{align}
    \int_\alpha^\beta\varphi(v)\omega(v)\,\rmd v
    =\int_0^{\Psi(\beta)}\varphi(\rho(\zeta))\,\rmd\zeta.
    \label{eq:erf-tr-inverse-substitution}
\end{align}
\textup{(iii)} If $\upsilon_1,\upsilon_2:[\alpha,\beta]\to\rset$ are Lipschitz, then $\{(\upsilon_1(v),\upsilon_2(v)):v\in[\alpha,\beta]\}$ has two-dimensional Lebesgue measure zero. An integral over $(0,\Psi(\beta))$ is zero when $\Psi(\beta)=0$.
\end{lemma}
\begin{proof}
Fix $\epsilon>0$. The relatively open set $O_\epsilon=\{v\in[\alpha,\beta]:\omega(v)<\epsilon\}$ is a countable disjoint union of intervals. On each component interval $J$, monotonicity and the integral expression for $\Psi$ imply $\operatorname{length}(\Psi(J))\leq\epsilon\operatorname{length}(J)$. Their images, together with at most their endpoint images, cover the image of the zero set of $\omega$. Countable subadditivity bounds its outer measure by $\epsilon(\beta-\alpha)$. Letting $\epsilon$ tend to zero proves nullity. Exclude these image values and $0,\Psi(\beta)$. The intermediate value theorem gives a solution $v\in(\alpha,\beta)$ of $\Psi(v)=\zeta$, and this solution has $\omega(v)>0$. If $v_1<v_2$ had the same image, monotonicity would make $\Psi$ constant on $[v_1,v_2]$; continuity and nonnegativity would give $\omega(v_1)=\omega(v_2)=0$, a contradiction. The same argument proves the last assertion whenever $\omega(v_0)>0$, without an exceptional set. This proves (i).

For continuous $\chi$, let $\Phi(\zeta)=\int_0^\zeta\chi(\zeta')\,\rmd\zeta'$. The chain rule gives $(\Phi\circ\Psi)'(v)=\chi(\Psi(v))\omega(v)$. Integration proves \eqref{eq:erf-tr-substitution}. Equality against continuous functions identifies the two finite measures, so the identity extends to nonnegative Borel $\chi$. For $\alpha<v<\beta$ with $\omega(v)>0$, monotonicity gives $\rho(\Psi(v))=v$. Applying \eqref{eq:erf-tr-substitution} to $\chi(\zeta)=\varphi(\rho(\zeta))$, with arbitrary values at the endpoints and exceptional $\zeta$, proves \eqref{eq:erf-tr-inverse-substitution}.

For (iii), let $\Lambda$ bound both Lipschitz constants. Divide $[\alpha,\beta]$ into $m$ intervals of length $(\beta-\alpha)/m$. Each image lies in a square of side at most $2\Lambda(\beta-\alpha)/m$. The total area of these $m$ squares is at most $4\Lambda^2(\beta-\alpha)^2/m$. Letting $m\to\infty$ proves the assertion.
\end{proof}

For a numerical pair $c\in\intset{1}{\len}\times\intset{1}{\vocab}$, define
\begin{align}
    \mathcal B_c=\{r\in[0,\infty)^D:r^c=0,\ r^d>0\text{ for every }d\neq c\}.
    \label{eq:erf-tr-face-set}
\end{align}
Let $\iota_c:(0,\infty)^{D-1}\to\mathcal B_c$ insert zero in coordinate $c$, using lexicographic order for the others. The measure $\sigma_c$ on $\mathcal B_c$ is defined by
\begin{align}
    \int_{\mathcal B_c}\varphi(w)\,\sigma_c(\rmd w)
    =\int_{(0,\infty)^{D-1}}\varphi(\iota_c(\zeta))\,\rmd\zeta
    \label{eq:erf-tr-face-measure}
\end{align}
for nonnegative Borel $\varphi$; equivalently $\sigma_c(\rmd w)=\delta_0(\rmd w^c)\prod_{d\neq c}\rmd w^d$ on $\mathcal B_c$. For $\bx\in\msx$, let
\begin{align}
    c_\ell(\bx)=\text{the unique pair }(\ell,j)\text{ with }\representer{j}=\bx^\ell,
    \qquad
    \mathcal O_{\bx}=\mathcal O\cup
       \bigcup_{\ell,j:\representer{j}\neq\bx^\ell}\mathcal B_{(\ell,j)}.
    \label{eq:erf-tr-Zx}
\end{align}
Thus $r^{c_\ell(\bx)}=\dotp{\bx^\ell}{r^\ell}$. For $u\in[0,1)$, $\bx\in\msx$, and $r\in\mathcal O_{\bx}$, define
\begin{align}
    \mathcal N_u(\bx)
    &=\{r\in\mathcal O_{\bx}:\dotp{\bx^h}{r^h}>A_{u\tbar1}^h(\bx)
        \text{ for every }h\},\label{eq:erf-tr-N}\\
    \rho_\ell(u,\bx,r)
    &=\inf\{v\in[u,1]:A_{u\tbar v}^\ell(\bx)\geq\dotp{\bx^\ell}{r^\ell}\},
    \qquad\inf\varnothing=+\infty.\label{eq:erf-tr-root}
\end{align}
Define also
\begin{align}
    \begin{split}
        \mathcal R_u^\ell(\bx)=\bigl\{r\in\mathcal O_{\bx}:\exists v\in(u,1)
        \text{ such that }\quad
        &\lambda_v^\ell(\bx^\ell)>0,\\
        &\dotp{\bx^\ell}{r^\ell}=A_{u\tbar v}^\ell(\bx),\\
        &\dotp{\bx^h}{r^h}>A_{u\tbar v}^h(\bx)\quad(h\neq\ell),\\
        &r^d-\Delta_{u\tbar v}^d(\bx)>0\quad(d\neq c_\ell(\bx))\bigr\},
    \end{split}
    \label{eq:erf-tr-R-ell-region}
\end{align}
and
\begin{align}
    \mathcal R_u(\bx)=\bigcup_{\ell=1}^{\len}\mathcal R_u^\ell(\bx).
    \label{eq:erf-tr-R-region}
\end{align}
For $r\in\mathcal R_u^\ell(\bx)$, use the notation
\begin{align}
    v=\rho_\ell(u,\bx,r),\qquad c=c_\ell(\bx),\qquad
    w=r-\Delta_{u\tbar v}(\bx)\in\mathcal B_c.
    \label{eq:erf-tr-reconstruction}
\end{align}

\begin{lemma}
\label{lem:erf-tr-reconstruction}
Assume \Cref{ass:erf-tr-standing}.

\textup{(i)} The maps in \eqref{eq:erf-tr-root} are Borel in $(u,r)$ for fixed $\bx$. The regions in \eqref{eq:erf-tr-N} and \eqref{eq:erf-tr-R-ell-region}, as sets of $(u,r)$, are Borel. The sets $\mathcal R_u^\ell(\bx)$ are pairwise disjoint and disjoint from $\mathcal N_u(\bx)$.

\textup{(ii)} For each fixed $u\in[0,1)$ and $\bx\in\msx$,
\begin{align}
    \operatorname{Leb}_D\bigl(\mathcal O\setminus(\mathcal N_u(\bx)\cup\mathcal R_u(\bx))\bigr)=0.
    \label{eq:erf-tr-null-interior}
\end{align}
For each such $u,\bx$ and $d=(h,j)$ with $\representer{j}\neq\bx^h$ and $\rratemat{u}{\bx}{\representer{j}}[h]>0$,
\begin{align}
    \sigma_d\bigl(\mathcal B_d\setminus(\mathcal N_u(\bx)\cup\mathcal R_u(\bx))\bigr)=0.
    \label{eq:erf-tr-null-face}
\end{align}
Equivalently, the exceptional set is null for Lebesgue measure in the $D-1$ coordinates other than $d$. In particular, for each fixed $\bx$ and active $d=(h,j)$,
\begin{align}
    \int_0^1\int_{\mathcal B_d}
       \indic_{\{r\notin\mathcal N_u(\bx)\cup\mathcal R_u(\bx)\}}
       \rratemat{u}{\bx}{\representer{j}}[h]\,\sigma_d(\rmd r)\,\rmd u=0.
    \label{eq:erf-tr-null-face-time}
\end{align}
\textup{(iii)} Fix $0\leq u<v<1$, $\bx\in\msx$, a position $\ell$, and $w\in\mathcal B_{c_\ell(\bx)}$ with $\lambda_v^\ell(\bx^\ell)>0$. Set $r=w+\Delta_{u\tbar v}(\bx)$. If $r\in\mathcal O$, there is no active hit on $[u,v)$, and $r\in\mathcal R_u^\ell(\bx)$ with $\rho_\ell(u,\bx,r)=v$. The same reconstruction conclusion holds if $r\in\mathcal B_d$ for an active $d=(h,j)$ with $\rratemat{u}{\bx}{\representer{j}}[h]>0$; in this case the first active hit after leaving $v$ is at $u$ through $d$.

\textup{(iv)} Conversely, if $r\in\mathcal R_u^\ell(\bx)$, then \eqref{eq:erf-tr-reconstruction} gives the only pair $(v,\ell)$ for which a segment in state $\bx$, starting with only residual $c_\ell(\bx)$ zero and positive initial passive speed, can end at $(u,\bx,r)$ without an intervening active hit. For $r\in\mathcal B_d$, this assertion is restricted to $\rratemat{u}{\bx}{\representer{j}}[h]>0$, where $d=(h,j)$.
\end{lemma}
\begin{proof}
For $\tau\in[0,1]$, on the domain $r\in\mathcal O_{\bx}$,
\begin{align*}
    \{(u,r):\rho_\ell(u,\bx,r)\leq\tau\}
    =\{(u,r):u\leq\tau,\ \dotp{\bx^\ell}{r^\ell}\leq A_{u\tbar\tau}^\ell(\bx)\}.
\end{align*}
The right-hand side is Borel because the integrated rate is continuous. If a witness $v$ exists in \eqref{eq:erf-tr-R-ell-region}, positivity of $\lambda_v^\ell(\bx^\ell)$ and \Cref{lem:erf-tr-calculus}(i) imply that it is unique and equals $\rho_\ell(u,\bx,r)$. The other passive inequalities imply $\rho_h(u,\bx,r)>v$ for $h\neq\ell$. Membership in the region is therefore tested on the Borel set $\{u<\rho_\ell(u,\bx,r)<1\}$ by substituting this Borel function into the remaining equalities and inequalities. This proves the Borel assertions and disjointness across positions. A finite root is incompatible with all the inequalities defining $\mathcal N_u(\bx)$.

Fix $u,\bx$. For each $\ell$, exclude the values
\begin{align*}
    \{0,A_{u\tbar1}^\ell(\bx)\}\cup
    \{A_{u\tbar v}^\ell(\bx):v\in[u,1],\ \lambda_v^\ell(\bx^\ell)=0\}
\end{align*}
for the coordinate $\dotp{\bx^\ell}{r^\ell}$. These sets have one-dimensional measure zero by \Cref{lem:erf-tr-calculus}. For $\ell\neq h$, also exclude pairs
\begin{align*}
    \bigl(\dotp{\bx^\ell}{r^\ell},\dotp{\bx^h}{r^h}\bigr)
    \in\{(A_{u\tbar v}^\ell(\bx),A_{u\tbar v}^h(\bx)):v\in[u,1]\}.
\end{align*}
Their planar measure is zero by \Cref{lem:erf-tr-calculus}(iii). Fubini and the finite number of indices show that the excluded set is null in $\mathcal O$. If no root is finite, the point belongs to $\mathcal N_u(\bx)$. Otherwise the least root $v$ has a unique position $\ell$, belongs to $(u,1)$, and has $\lambda_v^\ell(\bx^\ell)>0$. For $h\neq\ell$, strict minimality gives $\dotp{\bx^h}{r^h}-A_{u\tbar v}^h(\bx)>0$. For an active $d=(h,j)$,
\begin{align*}
    r^d-\Delta_{u\tbar v}^d(\bx)=r^d+B_{u\tbar v}^{hj}(\bx)>0.
\end{align*}
The remaining equality at $c_\ell(\bx)$ follows from continuity at the first root. Thus the point belongs to $\mathcal R_u^\ell(\bx)$, proving \eqref{eq:erf-tr-null-interior}.

On $\mathcal B_d$, with $d$ active at $\bx$, all $\len$ passive coordinates remain free under $\sigma_d$. The excluded sets are still null by Fubini in these coordinates. The only additional zero coordinate is $r^d=0$. If $d=(h,j)$ and $\rratemat{u}{\bx}{\representer{j}}[h]>0$, continuity yields $B_{u\tbar v}^{hj}(\bx)>0$ for every $v>u$. Consequently $w^d>0$, and the same argument proves \eqref{eq:erf-tr-null-face}. Joint Borel measurability and Tonelli, restricted first to bounded coordinate boxes if necessary, give \eqref{eq:erf-tr-null-face-time}.

For (iii), $\dotp{\bx^\ell}{r^\ell}=A_{u\tbar v}^\ell(\bx)$. Since $\lambda_v^\ell(\bx^\ell)>0$, this integral is strictly smaller for any upper limit below $v$, so $\rho_\ell(u,\bx,r)=v$. For $h\neq\ell$, positivity of $\dotp{\bx^h}{w^h}$ gives $\dotp{\bx^h}{r^h}>A_{u\tbar v}^h(\bx)$ and hence $\rho_h(u,\bx,r)>v$. If $r$ is interior, every active residual on $[u,v]$ is at least its positive final value. Residual $c_\ell(\bx)$ increases from zero and is positive at every time below $v$, since its speed is positive immediately below $v$. All other passive residuals increase from positive values. This proves absence of a hit. If $r^d=0$ and $\rratemat{u}{\bx}{\representer{j}}[h]>0$, residual $d$ has value $B_{u\tbar\tau}^{hj}(\bx)>0$ at every $\tau\in(u,v)$, while the other active residuals have positive final values. Its first hit is therefore exactly at $u$.

For (iv), the definitions give the stated $w$, its positivity, and the specified speed. The preceding monotonicity arguments give the required segment. By (iii), another such segment would have to produce the same unique minimum among the $\rho_h$. Its time, position, and initial residual vector are therefore those in \eqref{eq:erf-tr-reconstruction}.
\end{proof}

\begin{lemma}
\label{lem:erf-tr-construction}
Under \Cref{ass:erf-tr-standing}, every finite construction in \eqref{eq:erf-tr-candidate-time}--\eqref{eq:erf-tr-segment-path}, including its success and failure events, is Borel measurable in its initial condition. After a hit satisfying \eqref{eq:erf-tr-contact-conditions} at which only its triggering residual is zero, every residual is strictly positive for a sufficiently short interval in decreasing time after the update, and no immediate further hit occurs.
\end{lemma}
\begin{proof}
For fixed $\bx$, active $(\ell,j)$, and $u\in[0,1]$,
\begin{align*}
    \{(v,\eta):\theta^{\ell j}(v,\bx,\eta)\geq u\}
    =\left\{(v,\eta):v\geq u,\
       \int_u^v\rratemat{\tau}{\bx}{\representer{j}}[\ell]\,\rmd\tau\geq\eta^{\ell j}\right\}
\end{align*}
on $\eta^{\ell j}>0$. The right-hand side is Borel. Finite maxima, tests of uniqueness and positivity, and integrations of continuous rates are Borel operations. Induction over these finitely many operations proves the first assertion, including measurability of \eqref{eq:erf-tr-Htk}--\eqref{eq:erf-tr-mu-k}.

Suppose the state changes from $\by$ to $\bx$ at time $v$, with $\by^{-\ell}=\bx^{-\ell}$, $\by^\ell\neq\bx^\ell$, and $\rratemat{v}{\by}{\bx^\ell}[\ell]>0$. Equation \eqref{eq:erf-tr-positive-inward} gives $\lambda_v^\ell(\bx^\ell)>0$. By continuity, $\lambda_\tau^\ell(\bx^\ell)>0$ throughout a sufficiently short interval immediately below $v$, so the new passive residual increases strictly from zero. Every other residual is positive at the hit and has bounded speed, so all remain positive on a sufficiently short common interval. This proves the second assertion.
\end{proof}

\subsection{Change of variables}

\begin{lemma}\label{lem:erf-tr-initial}
Assume \Cref{ass:erf-tr-standing} and \eqref{eq:erf-tr-init}. Define, for every $u\in[0,1)$, $\bx\in\msx$, and $r\in\mathcal O_{\bx}$,
\begin{align}\label{eq:erf-tr-f0}
    f_u^{(0)}(\bx,r):=F_u(\bx,r)\indic_{\mathcal N_u(\bx)}(r).
\end{align}
Then, for every $u\in[0,1)$, every $\bx\in\msx$, and every Borel $B\subseteq\mathcal O$,
\begin{align}\label{eq:erf-tr-f0-law}
    \mu_u^{(0)}(\{\bx\}\times B)=\int_B f_u^{(0)}(\bx,r)\,\rmd r.
\end{align}
The measure $\mu_u^{(0)}$ gives zero mass to $\msx\times([0,\infty)^D\setminus\mathcal O)$.

For every $\bx\in\msx$, $d=(h,j)\in\intset{1}{\len}\times\intset{1}{\vocab}$ with $\representer{j}\neq\bx^h$, and nonnegative Borel $\psi:(0,1)\times\mathcal B_d\to[0,\infty]$,
\begin{align}\label{eq:erf-tr-first-flux}
    &\pE\!\left[\indic_{\{T^{\star}_1\in(0,1),\,Y_1=\bx,\,C_1=d\}}\psi(T^{\star}_1,E_{T_1^{\star}})\right]\notag\\
    &\qquad=\int_0^1\int_{\mathcal B_d}
       \psi(v,w)\rratemat{v}{\bx}{\representer{j}}[h]f_v^{(0)}(\bx,w)
       \,\sigma_d(\rmd w)\,\rmd v.
\end{align}
The first hit, when present on $[0,1]$, satisfies \eqref{eq:erf-tr-contact-conditions} almost surely and has only its firing residual equal to zero. For each fixed $s\in[0,1]$, the probability of a first hit at $s$ is zero.
\end{lemma}
\begin{proof}
Fix $u,\bx$. On $H_u^{(0)}\cap\{X_u=\bx\}$, the initial residuals corresponding to the endpoint $r$ are
\begin{align*}
    \eta=r-\Delta_{u\tbar1}(\bx).
\end{align*}
For a passive index $c_\ell(\bx)$, positivity of $\eta^{c_\ell(\bx)}$ is equivalent to
$\dotp{\bx^\ell}{r^\ell}>A_{u\tbar1}^\ell(\bx)$. For an active index $(\ell,j)$,
\begin{align*}
    \eta^{\ell j}=r^{\ell j}+B_{u\tbar1}^{\ell j}(\bx).
\end{align*}
If $r$ is interior, the active residual remains positive on the whole segment: at $\tau\in[u,1]$ it equals $r^{\ell j}+B_{u\tbar\tau}^{\ell j}(\bx)$. These conditions are also necessary. The endpoint map is a translation with derivative matrix $I_D$. Its density is therefore
\begin{align*}
    \pdata{1}{}{\bx}\gamma(r-\Delta_{u\tbar1}(\bx))\indic_{\mathcal N_u(\bx)}(r)
    =F_u(\bx,r)\indic_{\mathcal N_u(\bx)}(r),
\end{align*}
by \eqref{eq:erf-tr-transport}. Initial residuals are strictly positive, and no hit down to $u$ implies positive final active residuals and positive passive residuals. This proves \eqref{eq:erf-tr-f0-law} and the assertion about boundary mass.

We first exclude the exceptional first hits. Condition on $X_1=\bx$. For an active $d=(h,j)$, a finite candidate hitting time $v$ satisfies
\begin{align*}
    E_1^d=B_{v\tbar1}^{hj}(\bx).
\end{align*}
As a function of $v$, the right-hand side is $C^1$, nonincreasing, and has derivative $-\rratemat{v}{\bx}{\representer{j}}[h]$. \Cref{lem:erf-tr-calculus}, applied after reversing the time parameter, shows that the initial values yielding a zero-speed candidate form a one-dimensional null set. An exponential variable avoids that set almost surely. A candidate at a prescribed $s\in[0,1]$ requires $E_1^d=B_{s\tbar1}^{hj}(\bx)$, which is one fixed value and has probability zero. A tie for two distinct active indices $d=(h,j)$ and $a=(\ell,k)$ requires
\begin{align*}
    (E_1^d,E_1^a)\in
    \{(B_{v\tbar1}^{hj}(\bx),B_{v\tbar1}^{\ell k}(\bx)):v\in[0,1]\}.
\end{align*}
This set is planar-null by \Cref{lem:erf-tr-calculus}(iii), whereas the pair of independent exponentials has a Lebesgue density. A finite union over indices and states excludes all these cases. The passive residuals increase from positive values, so all residuals other than the unique firing residual are positive at this hit.

Now fix $\bx,d=(h,j)$ and write $w=\iota_d(\zeta)$. Consider the open set
\begin{align}\label{eq:erf-tr-first-map-domain}
    \begin{split}
    \mathcal U_{\bx,d}:=\{(v,\zeta)\in(0,1)\times(0,\infty)^{D-1}:\ &\rratemat{v}{\bx}{\representer{j}}[h]>0,\\
       &\dotp{\bx^\ell}{w^\ell}>A_{v\tbar1}^\ell(\bx)
          \text{ for every }\ell\in\intset{1}{\len}\}.
    \end{split}
\end{align}
Define $U:\mathcal U_{\bx,d}\to(0,\infty)^D$ by
\begin{align}\label{eq:erf-tr-first-map}
    U(v,\zeta)=w-\Delta_{v\tbar1}(\bx).
\end{align}
In particular, its $d$-th coordinate is $B_{v\tbar1}^{hj}(\bx)$. Its other active coordinates are $w^a+B_{v\tbar1}^{a}(\bx)$, where $B_{v\tbar1}^{a}=B_{v\tbar1}^{\ell k}$ for $a=(\ell,k)$, and its passive coordinates are $\dotp{\bx^\ell}{w^\ell}-A_{v\tbar1}^\ell(\bx)$. All of them are positive on \eqref{eq:erf-tr-first-map-domain}.

Order the target coordinates as $(d,-d)$ and the source coordinates as $(v,\zeta)$. Since $\partial_v\Delta_{v\tbar1}(\bx)=-b_v(\bx)$,
\begin{align}\label{eq:erf-tr-first-jacobian}
    DU(v,\zeta)=
    \begin{pmatrix}
        -\rratemat{v}{\bx}{\representer{j}}[h]&0_{1\times(D-1)}\\
        b_v^{-d}(\bx)&I_{D-1}
    \end{pmatrix},
    \qquad
    |\det DU(v,\zeta)|=\rratemat{v}{\bx}{\representer{j}}[h].
\end{align}
Here $b_v^{-d}$ is the vector with its $d$-th component removed. Thus every non-firing column is a column of the identity, and expansion of the determinant in the first row gives the displayed value.

The map $U$ is injective. Indeed, its $d$-th coordinate determines $v$ uniquely among points where $\rratemat{v}{\bx}{\representer{j}}[h]>0$, by monotonicity and \Cref{lem:erf-tr-calculus}(i). Once $v$ is known, $w=\eta+\Delta_{v\tbar1}(\bx)$ determines $\zeta$. For an input in its image, the firing residual first reaches zero at $v$; the other active residuals have positive value $w^a$ at $v$ and did not reach zero earlier. Conversely, every first hit satisfying \eqref{eq:erf-tr-contact-conditions} yields one point of \eqref{eq:erf-tr-first-map-domain}. Therefore the change-of-variables theorem applied to $U$ gives the left-hand side of \eqref{eq:erf-tr-first-flux} as
\begin{align*}
    \int_{\mathcal U_{\bx,d}}
       \psi(v,\iota_d\zeta)\,
       \pdata{1}{}{\bx}\gamma(U(v,\zeta))\rratemat{v}{\bx}{\representer{j}}[h]
       \,\rmd v\,\rmd\zeta.
\end{align*}
The map is $C^1$ and has nonzero determinant on its open domain, so this change of variables is applicable to all nonnegative Borel integrands. By \eqref{eq:erf-tr-transport},
\begin{align*}
    \pdata{1}{}{\bx}\gamma(w-\Delta_{v\tbar1}(\bx))\,
    \prod_{\ell=1}^{\len}\indic_{\{\dotp{\bx^\ell}{w^\ell}>A_{v\tbar1}^\ell(\bx)\}}
    =f_v^{(0)}(\bx,w).
\end{align*}
Inserting this identity, using \eqref{eq:erf-tr-face-measure}, and observing that the integrand is zero when $\rratemat{v}{\bx}{\representer{j}}[h]=0$ proves \eqref{eq:erf-tr-first-flux}. The exceptional hits have already been excluded.
\end{proof}

\begin{lemma}\label{lem:erf-tr-entrance}
Assume \Cref{ass:erf-tr-standing} and fix $\bx\in\msx$. For each $\ell\in\intset{1}{\len}$, let $(v,w)\mapsto g_v^\ell(\bx,w)$ be a nonnegative, finite-valued Borel function on $(0,1)\times\mathcal B_{c_\ell(\bx)}$. Suppose that
\begin{align}\label{eq:erf-tr-a-assumptions}
    \sum_{\ell=1}^{\len}\int_0^1\int_{\mathcal B_{c_\ell(\bx)}}
       g_v^\ell(\bx,w)\,\sigma_{c_\ell(\bx)}(\rmd w)\,\rmd v<\infty,
    \qquad
    g_v^\ell(\bx,w)=0\ \text{if }\lambda_v^\ell(\bx^\ell)=0.
\end{align}
For data $(v,\ell,w)$ distributed according to these finite measures, start the constant-state segment at physical time $v$ in state $\bx$, with residual vector $w$, and follow \eqref{eq:erf-tr-clocks} toward $0$ until its first active hit. Denote that hitting time by $\theta(v,\bx,w)$, its residual index by $C(v,\bx,w)$, and set $\theta=-\infty$ if there is no hit. Values of $C$ on exceptional hits may be fixed arbitrarily.

For every $u\in[0,1)$ and $r\in\mathcal O_{\bx}$, define
\begin{align}\label{eq:erf-tr-h-definition}
    h_u(\bx,r):=
    \begin{cases}
        \displaystyle\frac{g_v^\ell(\bx,w)}{\lambda_v^\ell(\bx^\ell)},
           &r\in\mathcal R_u^\ell(\bx),\quad v=\rho_\ell(u,\bx,r),\quad
             w=r-\Delta_{u\tbar v}(\bx),\\[3mm]
        0,&r\notin\mathcal R_u(\bx).
    \end{cases}
\end{align}
This defines its values on $\mathcal B_d$ separately from its interpretation as an interior density. The following statements hold.

(i) For every deterministic $u\in[0,1)$ and every Borel set $B\subseteq\mathcal O$,
\begin{align}\label{eq:erf-tr-h-interior-law}
    \int_B h_u(\bx,r)\,\rmd r
    =\sum_{\ell=1}^{\len}\int_u^1\int_{\mathcal B_{c_\ell(\bx)}}
       \indic_{\{w+\Delta_{u\tbar v}(\bx)\in B\}}g_v^\ell(\bx,w)
       \,\sigma_{c_\ell(\bx)}(\rmd w)\,\rmd v.
\end{align}
Only source times $v$ with $u<v<1$ occur in this formula. Its right-hand side is the measure of the data whose segment has no active hit down to $u$ and ends in $B$.

(ii) For every active index $d=(h,j)$ with $\representer{j}\neq\bx^h$ and every nonnegative Borel $\psi:(0,1)\times\mathcal B_d\to[0,\infty]$,
\begin{align}\label{eq:erf-tr-h-exit-law}
    &\sum_{\ell=1}^{\len}\int_0^1\int_{\mathcal B_{c_\ell(\bx)}}
       g_v^\ell(\bx,w)
       \indic_{\{0<\theta(v,\bx,w)<v,\ C(v,\bx,w)=d\}}\notag\\[-2pt]
    &\hspace{25mm}\times
       \psi\bigl(\theta(v,\bx,w),w+\Delta_{\theta(v,\bx,w)\tbar v}(\bx)\bigr)
       \,\sigma_{c_\ell(\bx)}(\rmd w)\,\rmd v\notag\\
    &\qquad=\int_0^1\int_{\mathcal B_d}
       \psi(u,r)\rratemat{u}{\bx}{\representer{j}}[h]h_u(\bx,r)
       \,\sigma_d(\rmd r)\,\rmd u.
\end{align}
Thus $u$ on the right is the hitting time, and its source time always satisfies $0<u<v<1$.

(iii) With respect to the finite source measures in \eqref{eq:erf-tr-a-assumptions}, a next hit, when present, is almost surely unique and has positive incoming speed. For every fixed $s\in[0,1]$, the set of data whose next hit occurs at $s$ has measure zero. At a hit, only the firing residual is zero. No singular part in the interior or in the hitting measure is omitted in \eqref{eq:erf-tr-h-interior-law}--\eqref{eq:erf-tr-h-exit-law}.
\end{lemma}
\begin{proof}
Fix an index $\ell$, let $c=c_\ell(\bx)$, and consider its source measure. By \eqref{eq:erf-tr-a-assumptions}, it is enough to use $v\in(0,1)$ with $\lambda_v^\ell(\bx^\ell)>0$.

For a fixed such $v$, all $D-1$ coordinates $w^{-c}$ have a density with respect to Lebesgue measure; independence is not needed. Residual $c$ is passive and increases from zero, with positive speed immediately below $v$. It is therefore positive at every later elapsed time before the next state change. Every other passive residual increases from a strictly positive value. Each active residual $d=(h,j)$ starts from a positive free coordinate and has a finite candidate hitting time $u$ only if
\begin{align}\label{eq:erf-tr-fixed-r-candidate}
    w^d=B_{u\tbar v}^{hj}(\bx),\qquad 0\leq u<v.
\end{align}
For fixed $v$, \Cref{lem:erf-tr-calculus} shows that values of $w^d$ yielding $\rratemat{u}{\bx}{\representer{j}}[h]=0$ form a null set. Two candidate hitting times coincide only if their two free coordinates lie on the Lipschitz curve of their integrated rates. This is null by \Cref{lem:erf-tr-calculus}(iii). A hit at a fixed $s<v$ imposes the single equality $w^d=B_{s\tbar v}^{hj}(\bx)$ and is null as well. At $s=v$, an active hit is impossible because all active coordinates are positive initially. These conclusions hold under any density in $w^{-c}$ by Fubini; integrating in $v$ proves the assertions of (iii) about exceptional hits and prescribed times, including $s=0$. Positivity of every non-firing residual at a unique hit follows from the same monotonicity and positivity of the passive residuals.

We prove (i). Fix $u\in[0,1)$, and write $w=\iota_c(\zeta)$ with $\zeta\in(0,\infty)^{D-1}$. On the open set
\begin{align}\label{eq:erf-tr-interior-map-domain}
    \{(v,\zeta):u<v<1,\ \lambda_v^\ell(\bx^\ell)>0,\quad
       \zeta>0,\ \iota_c(\zeta)+\Delta_{u\tbar v}(\bx)\in\mathcal O\},
\end{align}
consider
\begin{align*}
    V_u(v,\zeta)=\iota_c(\zeta)+\Delta_{u\tbar v}(\bx).
\end{align*}
Order the target coordinates as $(c,-c)$. Its derivative matrix is
\begin{align}\label{eq:erf-tr-interior-jacobian}
    DV_u(v,\zeta)=
    \begin{pmatrix}
        \lambda_v^\ell(\bx^\ell)&0_{1\times(D-1)}\\
        b_v^{-c}(\bx)&I_{D-1}
    \end{pmatrix},
    \qquad |\det DV_u|=\lambda_v^\ell(\bx^\ell).
\end{align}
In particular, the component $r^c=A_{u\tbar v}^\ell(\bx)$ has derivative $\lambda_v^\ell(\bx^\ell)$ in $v$ and no derivative in the free coordinates. Each other component has derivative $b_v^a(\bx)$ in $v$ and the identity derivatives in those free coordinates. This proves the determinant formula by expansion in the first row.

\Cref{lem:erf-tr-reconstruction}(iii)--(iv) shows that $V_u$ is injective on \eqref{eq:erf-tr-interior-map-domain}, with image $\mathcal R_u^\ell(\bx)\cap\mathcal O$ and inverse \eqref{eq:erf-tr-reconstruction}. Images for distinct $\ell$ are disjoint. Positivity of the final active coordinates is equivalent to absence of a hit down to $u$; all passive coordinates remain positive after leaving $v$. Applying the change-of-variables theorem, with determinant \eqref{eq:erf-tr-interior-jacobian}, and then summing over $\ell$ proves \eqref{eq:erf-tr-h-interior-law}. The source measure is zero where $\lambda_v^\ell(\bx^\ell)=0$, and the map has nonzero determinant everywhere else on its domain. Thus its interior pushforward has no omitted singular part.

We next prove (ii). Fix an active $d=(h,j)$, which is necessarily different from $c$. Use variables
\begin{align*}
    0<u<v<1,\qquad \zeta\in(0,\infty)^{D-2},
\end{align*}
indexed in the last factor by $(\intset{1}{\len}\times\intset{1}{\vocab})\setminus\{c,d\}$. Define $r$ and $w$ by
\begin{align}\label{eq:erf-tr-twoface-variables}
    \begin{split}
    r^d&=0,\qquad r^c=A_{u\tbar v}^\ell(\bx),\qquad
       r^a=\zeta^a\quad(a\notin\{c,d\}),\\
    w^c&=0,\qquad w^d=B_{u\tbar v}^{hj}(\bx),\qquad
       w^a=\zeta^a-\Delta_{u\tbar v}^a(\bx)\quad(a\notin\{c,d\}).
    \end{split}
\end{align}
Restrict these variables to the open set on which
\begin{align}\label{eq:erf-tr-twoface-domain}
    \lambda_v^\ell(\bx^\ell)>0,\qquad
    \rratemat{u}{\bx}{\representer{j}}[h]>0,\qquad
    w^a>0\quad(a\neq c).
\end{align}
On this set, $r\in\mathcal B_d$. Conversely, every next hit of the type under consideration, outside the zero-measure exceptions already excluded, determines exactly these variables. In fact, \eqref{eq:erf-tr-fixed-r-candidate} determines $u$, and each competing active residual has the positive final value $r^a$. The other passive residuals are positive at both ends of the segment.

There are two changes of variables on \eqref{eq:erf-tr-twoface-domain}. First define
\begin{align*}
    U_1(v,u,\zeta)=(v,w^{-c}),
\end{align*}
ordering the target as $(v,w^d,w^{-\{c,d\}})$. Since
\begin{align*}
    \partial_v B_{u\tbar v}^{hj}(\bx)=\rratemat{v}{\bx}{\representer{j}}[h],
    \qquad
    \partial_u B_{u\tbar v}^{hj}(\bx)=-\rratemat{u}{\bx}{\representer{j}}[h],
\end{align*}
and $\partial_v\Delta_{u\tbar v}^a=b_v^a$, $\partial_u\Delta_{u\tbar v}^a=-b_u^a$, its matrix is
\begin{align}\label{eq:erf-tr-twoface-jacobian-1}
    DU_1=
    \begin{pmatrix}
        1&0&0\\
        \rratemat{v}{\bx}{\representer{j}}[h]&-\rratemat{u}{\bx}{\representer{j}}[h]&0\\
        -b_v^{-\{c,d\}}(\bx)&b_u^{-\{c,d\}}(\bx)&I_{D-2}
    \end{pmatrix}.
\end{align}
Expansion in the first row and then in the first row of the remaining minor gives
\begin{align}\label{eq:erf-tr-twoface-det-1}
    |\det DU_1|=\rratemat{u}{\bx}{\representer{j}}[h].
\end{align}
For a fixed image $(v,w^{-c})$, the equation $w^d=B_{u\tbar v}^{hj}(\bx)$ and the positive derivative magnitude at $u$ give a unique $u$ by \Cref{lem:erf-tr-calculus}. The other equations then determine $\zeta$, so $U_1$ is injective.

Second define
\begin{align*}
    U_2(v,u,\zeta)=(u,r^{-d}),
\end{align*}
ordering the target as $(u,r^c,r^{-\{c,d\}})$. Its matrix is
\begin{align}\label{eq:erf-tr-twoface-jacobian-2}
    DU_2=
    \begin{pmatrix}
        0&1&0\\
        \lambda_v^\ell(\bx^\ell)&-\lambda_u^\ell(\bx^\ell)&0\\
        0&0&I_{D-2}
    \end{pmatrix},
    \qquad |\det DU_2|=\lambda_v^\ell(\bx^\ell).
\end{align}
Indeed, the determinant of the upper-left $2\times2$ block is $-\lambda_v^\ell(\bx^\ell)$. \Cref{lem:erf-tr-reconstruction} gives the unique inverse of $U_2$ on its image, by first determining $v$ from $(u,\bx,r)$ and then $w$. These calculations also apply when $D=2$, with $I_0$ and no $\zeta$ coordinates.

Applying these two substitutions in succession transforms
\begin{align*}
    g_v^\ell(\bx,w)\,\rmd v\,\sigma_c(\rmd w)
    \quad\text{on hits through }d
\end{align*}
into
\begin{align}\label{eq:erf-tr-twoface-density}
    \frac{\rratemat{u}{\bx}{\representer{j}}[h]}{\lambda_v^\ell(\bx^\ell)}g_v^\ell(\bx,w)
    \,\rmd u\,\sigma_d(\rmd r),\qquad
    r\in\mathcal R_u^\ell(\bx)\cap\mathcal B_d,\quad
    \rratemat{u}{\bx}{\representer{j}}[h]>0.
\end{align}
Both maps are $C^1$ on the stated open set, injective, and have nonzero
determinants there. The image of $U_2$ is exactly the set in
\eqref{eq:erf-tr-twoface-density} with its time coordinate. Images for distinct
$\ell$ are disjoint by \Cref{lem:erf-tr-reconstruction}. Summing
\eqref{eq:erf-tr-twoface-density} over $\ell$ gives \eqref{eq:erf-tr-h-exit-law}
with the pointwise definition \eqref{eq:erf-tr-h-definition}. Where
$\rratemat{u}{\bx}{\representer{j}}[h]=0$, its right-hand side is zero. The
fixed-$v$ argument at the start of the proof excludes any source mass of hits
with this speed or with ties. Hence there is no additional singular hitting
measure, which completes the proof of (ii)--(iii).
\end{proof}

\subsection{Induction on the number of jumps}

\begin{definition}
\label{def:erf-tr-operator}
Assume \Cref{ass:erf-tr-standing}. Let $f_u(\bx,r)$ be a nonnegative Borel function on
\begin{align*}
    \{(u,\bx,r):0\leq u<1,\ \bx\in\msx,\ r\in\mathcal O_{\bx}\}.
\end{align*}
For each such $(u,\bx,r)$, define
\begin{align}
    (\mathcal J f)_u(\bx,r)
    :=\begin{cases}
        \displaystyle\frac{1}{\lambda_v^\ell(\bx^\ell)}
        \sum_{\substack{\by^{-\ell}=\bx^{-\ell}\\\by^\ell\neq\bx^\ell}}
        \rratemat{v}{\by}{\bx^\ell}[\ell]f_v(\by,w),
        &r\in\mathcal R_u^\ell(\bx),\\[3mm]
        0,&r\notin\mathcal R_u(\bx).
    \end{cases}
    \label{eq:erf-tr-K}
\end{align}
In the first line, $\ell$ is the unique index for which $r\in\mathcal R_u^\ell(\bx)$, $v=\rho_\ell(u,\bx,r)$, and $w=r-\Delta_{u\tbar v}(\bx)$. If extended nonnegative functions are used, a summand with zero coefficient is defined to be zero. For each predecessor $\by$ in the sum, the index $c=c_\ell(\bx)$ is active at $\by$ and $w\in\mathcal B_c\subset\mathcal O_{\by}$; thus every evaluation in \eqref{eq:erf-tr-K} is on the specified domain. All denominators in its first line are strictly positive by \eqref{eq:erf-tr-R-ell-region}.
\end{definition}

\begin{proposition}
\label{prop:erf-tr-induction}
Assume \Cref{ass:erf-tr-standing} and \eqref{eq:erf-tr-init}. Define $f^{(0)}$ by \eqref{eq:erf-tr-f0}, with its values on all of $\mathcal O_{\bx}$, and recursively set
\begin{align}
    f^{(n+1)}:=\mathcal J f^{(n)},\qquad n\in\mathbb N\cup\{0\}.
    \label{eq:erf-tr-recursion}
\end{align}
Then every $f^{(n)}$ is a finite-valued nonnegative Borel function. For every $n\geq0$, every deterministic $u\in[0,1)$, every $\bx\in\msx$, and every Borel $B\subseteq\mathcal O$,
\begin{align}
    \mu_u^{(n)}(\{\bx\}\times B)
    &=\int_B f_u^{(n)}(\bx,r)\,\rmd r,
    &\mu_u^{(n)}\bigl(\msx\times([0,\infty)^D\setminus\mathcal O)\bigr)&=0.
    \label{eq:erf-tr-fk-law}
\end{align}
Moreover, for every active index $d=(h,j)$ at $\bx$ and every nonnegative Borel $\psi:(0,1)\times\mathcal B_d\to[0,\infty]$,
\begin{align}
    &\pE\!\left[\indic_{\{T^{\star}_{n+1}\in(0,1),\,Y_{n+1}=\bx,\,C_{n+1}=d\}}
        \psi(T^{\star}_{n+1},E_{T_{n+1}^{\star}})\right]\notag\\
    &\qquad=\int_0^1\int_{\mathcal B_d}
        \psi(v,w)\rratemat{v}{\bx}{\representer{j}}[h]
        f_v^{(n)}(\bx,w)\,\sigma_d(\rmd w)\,\rmd v.
    \label{eq:erf-tr-k-flux}
\end{align}
For each fixed finite index, a hit reached by the recursion is almost surely either absent or satisfies \eqref{eq:erf-tr-contact-conditions}; only its firing residual is zero. For every fixed $s\in[0,1]$ and $n\geq1$, the probability that the $n$th hit occurs at $s$ is zero. None of these assertions assumes nonexplosion.
\end{proposition}
\begin{proof}
The function $f^{(0)}$ is finite-valued and Borel by its explicit definition. \Cref{lem:erf-tr-reconstruction}(i), the finite sums in \eqref{eq:erf-tr-K}, and positivity of the denominator on its domain show inductively that every $f^{(n)}$ is Borel and finite-valued. We prove the probabilistic assertions simultaneously by induction.

For $n=0$, all assertions are \Cref{lem:erf-tr-initial}. Suppose that the assertion about $\mu^{(n)}$, the event formula \eqref{eq:erf-tr-k-flux} at level $n$, and regularity of the first $n+1$ hits have been proved. Fix a possible state $\bx$ after the $(n+1)$st hit and a position $\ell$. Write $c=c_\ell(\bx)$. The disjoint ways of entering $\bx$ through position $\ell$ have pre-jump states $\by$ with $\by^{-\ell}=\bx^{-\ell}$ and $\by^\ell\neq\bx^\ell$, and firing residual $c$. The residual vector is unchanged by the jump. Applying \eqref{eq:erf-tr-k-flux} to each such pre-jump state gives their sum as the measure with density
\begin{align}
    g_v^{(n+1),\ell}(\bx,w)
    &:=\sum_{\substack{\by^{-\ell}=\bx^{-\ell}\\\by^\ell\neq\bx^\ell}}
        \rratemat{v}{\by}{\bx^\ell}[\ell]f_v^{(n)}(\by,w),
    \qquad v\in(0,1),\quad w\in\mathcal B_c,
    \label{eq:erf-tr-a-k}
\end{align}
with respect to $\rmd v\,\sigma_c(\rmd w)$. More explicitly, for every nonnegative Borel $\varphi$ on $(0,1)\times\mathcal B_c$,
\begin{align}
    &\pE\!\left[\indic_{\{T^{\star}_{n+1}\in(0,1),\,X_{T^{\star}_{n+1}}=\bx,\,C_{n+1}=c\}}
        \varphi(T^{\star}_{n+1},E_{T_{n+1}^{\star}})\right]\notag\\
    &\qquad=\int_0^1\int_{\mathcal B_c}
        \varphi(v,w)g_v^{(n+1),\ell}(\bx,w)\,\sigma_c(\rmd w)\,\rmd v.
    \label{eq:erf-tr-a-law}
\end{align}
Summing \eqref{eq:erf-tr-a-law} over $\ell$ with $\varphi=1$ gives
\begin{align}
    \sum_{\ell=1}^{\len}\int_0^1\int_{\mathcal B_{c_\ell(\bx)}}
        g_v^{(n+1),\ell}(\bx,w)\,\sigma_{c_\ell(\bx)}(\rmd w)\,\rmd v
    =\pP(T^{\star}_{n+1}\in(0,1),X_{T^{\star}_{n+1}}=\bx)\leq1.
    \label{eq:erf-tr-a-finite}
\end{align}
If $\lambda_v^\ell(\bx^\ell)=0$, then $\ratemat{v}{\bx^\ell}{\by^\ell}[\ell]=0$ for every predecessor $\by$ in the sum. Equation \eqref{eq:erf-tr-edge} and strict positivity of $\pdata{v}{}{\by}$ imply $\rratemat{v}{\by}{\bx^\ell}[\ell]=0$. Therefore \eqref{eq:erf-tr-a-k} is zero at such a time. It satisfies all the hypotheses \eqref{eq:erf-tr-a-assumptions} of \Cref{lem:erf-tr-entrance}.

Given $(T^{\star}_{n+1},X_{T^{\star}_{n+1}},E_{T_{n+1}^{\star}})=(v,\bx,w)$, the segment until the next hit is exactly $w+\Delta_{\tau\tbar v}(\bx)$ in state $\bx$; it is a deterministic function of this triple. Consequently, after all preceding variables have been integrated out, its distribution is obtained by transporting the measures \eqref{eq:erf-tr-a-law}. \Cref{lem:erf-tr-entrance}(i) gives the density of its endpoint at any fixed $u<v$ without another hit as
\begin{align*}
    \frac{g_v^{(n+1),\ell}(\bx,w)}{\lambda_v^\ell(\bx^\ell)}
    \quad\text{when }r\in\mathcal R_u^\ell(\bx),\quad
    v=\rho_\ell(u,\bx,r),\quad w=r-\Delta_{u\tbar v}(\bx),
\end{align*}
and zero off the union of these sets. By \eqref{eq:erf-tr-a-k}, this is exactly \eqref{eq:erf-tr-K} applied to $f^{(n)}$. Its event description is $H_u^{(n+1)}$, so \eqref{eq:erf-tr-fk-law} follows at level $n+1$, including zero boundary mass.

\Cref{lem:erf-tr-entrance}(ii) applied to the same entrance measures gives the next-hit formula
\begin{align*}
    \rratemat{v}{\bx}{\representer{j}}[h]f_v^{(n+1)}(\bx,w)\,\rmd v\,\sigma_d(\rmd w),
    \qquad d=(h,j),\quad \representer{j}\neq\bx^h.
\end{align*}
This is \eqref{eq:erf-tr-k-flux} for the $(n+2)$nd hit. Part (iii) of that lemma proves its regularity, positivity of all other residuals, and absence of an atom at every prescribed time. \Cref{lem:erf-tr-construction} allows continuation after each successful hit. This completes the induction for every finite $n$.
\end{proof}

\begin{corollary}
\label{cor:erf-tr-recursion-forms}
Under the hypotheses of \Cref{prop:erf-tr-induction}, for every $n\geq0$, $u\in[0,1)$, $\bx\in\msx$, and Borel $B\subseteq\mathcal O$,
\begin{align}
    \mu_u^{(n+1)}(\{\bx\}\times B)
    ={}&\sum_{\ell=1}^{\len}
        \sum_{\substack{\by^{-\ell}=\bx^{-\ell}\\\by^\ell\neq\bx^\ell}}
        \int_u^1\int_{\mathcal B_{c_\ell(\bx)}}
        \indic_{\{w+\Delta_{u\tbar v}(\bx)\in B\}}\notag\\
    &\qquad\times\rratemat{v}{\by}{\bx^\ell}[\ell]f_v^{(n)}(\by,w)
        \,\sigma_{c_\ell(\bx)}(\rmd w)\,\rmd v.
    \label{eq:erf-tr-integral-recursion}
\end{align}
In this integral $v=T^{\star}_{n+1}$ and $u<v<1$. For every $r\in\mathcal R_u^\ell(\bx)$, with $(v,w)$ given by \eqref{eq:erf-tr-reconstruction},
\begin{align}
    \frac{f_u^{(n+1)}(\bx,r)}{F_u(\bx,r)}
    =\sum_{\substack{\by^{-\ell}=\bx^{-\ell}\\\by^\ell\neq\bx^\ell}}
        \frac{\ratemat{v}{\bx^\ell}{\by^\ell}[\ell]}{\lambda_v^\ell(\bx^\ell)}
        \frac{f_v^{(n)}(\by,w)}{F_v(\by,w)}.
    \label{eq:erf-tr-normalized-recursion}
\end{align}
In particular,
\begin{align}
    f_u^{(1)}(\bx,r)=F_u(\bx,r)
    \sum_{\substack{\by^{-\ell}=\bx^{-\ell}\\\by^\ell\neq\bx^\ell}}
        \frac{\ratemat{v}{\bx^\ell}{\by^\ell}[\ell]}{\lambda_v^\ell(\bx^\ell)}
        \prod_{h=1}^{\len}
        \indic_{\{\dotp{\by^h}{w^h}>A_{v\tbar1}^h(\by)\}}.
    \label{eq:erf-tr-explicit-f1}
\end{align}
Off $\mathcal R_u(\bx)$, $f_u^{(1)}(\bx,r)=0$ by definition.
\end{corollary}
\begin{proof}
Substitute \eqref{eq:erf-tr-a-k} into \eqref{eq:erf-tr-h-interior-law}. Its indicator is equivalent to no further hit before $u$ because $B\subseteq\mathcal O$ and the active residuals are decreasing along this segment. This gives \eqref{eq:erf-tr-integral-recursion}. For \eqref{eq:erf-tr-normalized-recursion}, the two identities \eqref{eq:erf-tr-edge} and \eqref{eq:erf-tr-transport} give, pointwise at the specified $(u,v,\bx,\by,r,w)$,
\begin{align}
    \rratemat{v}{\by}{\bx^\ell}[\ell]F_v(\by,w)
    &=\pdata{v}{}{\bx}\ratemat{v}{\bx^\ell}{\by^\ell}[\ell]\gamma(w)\notag\\
    &=\ratemat{v}{\bx^\ell}{\by^\ell}[\ell]F_u(\bx,r).
    \label{eq:erf-tr-weight-identity}
\end{align}
All factors $F$ are positive. Divide \eqref{eq:erf-tr-K} by $F_u(\bx,r)$ and use \eqref{eq:erf-tr-weight-identity} in each summand. Finally, substitute \eqref{eq:erf-tr-f0} at $(v,\by,w)$, where $w\in\mathcal O_{\by}$, to obtain \eqref{eq:erf-tr-explicit-f1}.
\end{proof}

\subsection{Bounds on the partial sums and number of jumps}

\begin{lemma}
\label{lem:erf-tr-domination}
Assume \Cref{ass:erf-tr-standing}. Define $F$ by \eqref{eq:erf-tr-F} on each $\mathcal O_{\bx}$, $f^{(0)}$ by \eqref{eq:erf-tr-f0}, and $f^{(n+1)}$ by \eqref{eq:erf-tr-recursion}. For every $u\in[0,1)$, $\bx\in\msx$, and $r\in\mathcal O_{\bx}$,
\begin{align}
    (\mathcal J F)_u(\bx,r)
    &=F_u(\bx,r)\indic_{\mathcal R_u(\bx)}(r),
    \label{eq:erf-tr-KF}\\
    f_u^{(0)}(\bx,r)+(\mathcal J F)_u(\bx,r)
    &=F_u(\bx,r)\indic_{\mathcal N_u(\bx)\cup\mathcal R_u(\bx)}(r)
      \leq F_u(\bx,r).
    \label{eq:erf-tr-candidate-identity}
\end{align}
For every integer $m\geq0$, at every such $(u,\bx,r)$,
\begin{align}
    0\leq S_u^{(m)}(\bx,r):=\sum_{n=0}^m f_u^{(n)}(\bx,r)\leq F_u(\bx,r).
    \label{eq:erf-tr-partial-bound}
\end{align}
These are pointwise statements about the specified functions, including their values at $r\in\mathcal B_d$ for every active index $d$ at $\bx$.
\end{lemma}
\begin{proof}
If $r\notin\mathcal R_u(\bx)$, the definition of $\mathcal J$ gives \eqref{eq:erf-tr-KF}. If $r\in\mathcal R_u^\ell(\bx)$, use \eqref{eq:erf-tr-reconstruction} and \eqref{eq:erf-tr-face} to calculate
\begin{align*}
    (\mathcal J F)_u(\bx,r)
    &=\frac{\gamma(w)}{\lambda_v^\ell(\bx^\ell)}
      \sum_{\substack{\by^{-\ell}=\bx^{-\ell}\\\by^\ell\neq\bx^\ell}}
      \pdata{v}{}{\by}\rratemat{v}{\by}{\bx^\ell}[\ell]\\
    &=\pdata{v}{}{\bx}\gamma(w)=F_u(\bx,r).
\end{align*}
The last step is \eqref{eq:erf-tr-transport}. Since $\mathcal N_u(\bx)$ and $\mathcal R_u(\bx)$ are disjoint, \eqref{eq:erf-tr-f0} yields \eqref{eq:erf-tr-candidate-identity} at every point of $\mathcal O_{\bx}$.

For $m=0$, \eqref{eq:erf-tr-partial-bound} follows from \eqref{eq:erf-tr-f0}. Suppose it holds for an integer $m$ at every time, state, and point of the specified domain. Linearity and positivity of the finite sums defining $\mathcal J$ imply
\begin{align*}
    S^{(m+1)}=f^{(0)}+\mathcal J S^{(m)}
    \leq f^{(0)}+\mathcal J F\leq F.
\end{align*}
Every evaluation of $S^{(m)}$ in this inequality is within its domain, as checked in \Cref{def:erf-tr-operator}. The induction is therefore valid pointwise, including on the faces, and proves the result.
\end{proof}

\begin{proposition}
\label{prop:erf-tr-nonexplosion}
Assume \Cref{ass:erf-tr-standing} and \eqref{eq:erf-tr-init}. For $u\in[0,1]$, set
\begin{align}
    M_u:=\int_u^1\sum_{\bx\in\msx}\pdata{v}{}{\bx}\bar\lambda_v(\bx)\,\rmd v.
    \label{eq:erf-tr-Ct}
\end{align}
Then $M_u<\infty$, and
\begin{align}
    M_u=\int_u^1\sum_{\bx\in\msx}\pdata{v}{}{\bx}\lambda_v(\bx)\,\rmd v.
    \label{eq:erf-tr-Ct-equality}
\end{align}
For every $u\in[0,1)$ and every integer $m\geq1$,
\begin{align}
    \sum_{q=1}^m\pP(T^{\star}_q\in(u,1))\leq M_u,
    \qquad \pP(T^{\star}_m\in(u,1))\leq\frac{M_u}{m}.
    \label{eq:erf-tr-event-tail}
\end{align}
Here $T^{\star}_q$ is a successful finite step time from \eqref{eq:erf-tr-recursive-path} and is set to $-\infty$ if that step does not exist. The construction almost surely has no failure and only finitely many hits on $[0,1]$. For every deterministic $u\in[0,1]$, its jump count
\begin{align}
    N_u:=\sum_{q\geq1}\indic_{\{T^{\star}_q\in(u,1)\}}
    \label{eq:erf-tr-Nt}
\end{align}
satisfies $\pE N_u\leq M_u$.
\end{proposition}
\begin{proof}
For each $q\in\{1,\ldots,m\}$, take $\psi(v,w)=\indic_{\{u<v<1\}}$ in \eqref{eq:erf-tr-k-flux} at level $q-1$ and sum over the pre-jump state and active residual. Since these choices partition the possible $q$th jumps, Tonelli's theorem gives
\begin{align}
    \sum_{q=1}^m\pP(T^{\star}_q\in(u,1))
    ={}&\int_u^1\sum_{\bx\in\msx}\sum_{\ell=1}^{\len}
      \sum_{j:\representer{j}\neq\bx^\ell}\rratemat{v}{\bx}{\representer{j}}[\ell]\notag\\
    &\qquad\times\int_{\mathcal B_{(\ell,j)}}
      S_v^{(m-1)}(\bx,w)\,\sigma_{(\ell,j)}(\rmd w)\,\rmd v.
    \label{eq:erf-tr-count-exact}
\end{align}
\Cref{lem:erf-tr-domination} applies pointwise to the functions on these particular faces. For each active $d=(\ell,j)$,
\begin{align}
    \int_{\mathcal B_d}F_v(\bx,w)\,\sigma_d(\rmd w)
    &=\pdata{v}{}{\bx}\int_{(0,\infty)^{D-1}}
      \exp\!\left(-\sum_{a\neq d}w^a\right)\,\rmd w^{-d}\notag\\
    &=\pdata{v}{}{\bx}\prod_{a\neq d}\int_0^\infty\exp(-\zeta)\,\rmd\zeta
      =\pdata{v}{}{\bx}.
    \label{eq:erf-tr-integrate-face}
\end{align}
Substitution in \eqref{eq:erf-tr-count-exact} proves the first inequality in \eqref{eq:erf-tr-event-tail}. \Cref{lem:erf-tr-balance} gives a common finite bound on $\bar\lambda_v(\bx)$, which proves $M_u<\infty$.

For each fixed $v$, the full time-reversal formula \eqref{eq:erf-tr-reverse-full} gives
\begin{align*}
    \sum_{\bx}\pdata{v}{}{\bx}\bar\lambda_v(\bx)
    &=\sum_{\bx}\sum_{\by\neq\bx}\pdata{v}{}{\bx}\rratemat{v}{\bx}{\by}\\
    &=\sum_{\bx}\sum_{\by\neq\bx}\pdata{v}{}{\by}\ratemat{v}{\by}{\bx}
      =\sum_{\by}\pdata{v}{}{\by}\lambda_v(\by).
\end{align*}
Integration proves \eqref{eq:erf-tr-Ct-equality}.

If $T^{\star}_m\in(u,1)$, then $T^{\star}_q\in(u,1)$ for all $1\leq q\leq m$, by their strict ordering. Hence
\begin{align*}
    m\pP(T^{\star}_m\in(u,1))
    \leq\sum_{q=1}^m\pP(T^{\star}_q\in(u,1))\leq M_u,
\end{align*}
which proves the second inequality.

For each finite index, \Cref{prop:erf-tr-induction} assigns probability zero to failure of the hitting conditions, including a hit at $0$. The union of these failure events over the positive integers is null. On its complement, an infinite number of hits in $[0,1]$ would imply $T^{\star}_n\in(0,1)$ for every $n$. These events decrease with $n$, and \eqref{eq:erf-tr-event-tail} at $u=0$ implies
\begin{align*}
    \pP\bigl(T^{\star}_n\in(0,1)\text{ for all }n\geq1\bigr)
    =\lim_{n\to\infty}\pP(T^{\star}_n\in(0,1))=0.
\end{align*}
Thus a successful finite construction eventually reaches a segment having no further hit down to $0$. This yields a path on all of $[0,1]$ with finitely many jumps. Every finite construction is measurable by \Cref{lem:erf-tr-construction}; the full path at each fixed time is defined on the countable union of these successful finite constructions and is consequently measurable. The probability of a hit at any given deterministic time is zero by \Cref{prop:erf-tr-induction} and a countable union over the index.

Finally, monotone convergence applied to the first inequality in \eqref{eq:erf-tr-event-tail} gives
\begin{align*}
    \pE N_u=\sum_{q\geq1}\pP(T^{\star}_q\in(u,1))\leq M_u.
\end{align*}
At $u=1$, both quantities are zero.
\end{proof}

\subsection{Proof of Theorem~\ref{thm:erf-tr-main}}

\begin{proof}[Proof of \Cref{thm:erf-tr-main}]
\Cref{prop:erf-tr-nonexplosion} gives the almost-sure construction, finiteness of the number of jumps, and absence of a hit at every prescribed deterministic time.

Fix $u\in[0,1)$. Except on the null set of failures, explosion, or a hit at $u$, the events $H_u^{(n)}$, $n\geq0$, are disjoint and cover the sample space. On this set $H_u^{(n)}=\{N_u=n\}$. By \eqref{eq:erf-tr-mu-k} and \Cref{prop:erf-tr-induction}, the law at time $u$ is therefore supported on $\msx\times\mathcal O$ and has density
\begin{align}
    f_u(\bx,r):=\sum_{n=0}^{\infty}f_u^{(n)}(\bx,r),
    \qquad\bx\in\msx,\quad r\in\mathcal O.
    \label{eq:erf-tr-sum-density}
\end{align}
\Cref{lem:erf-tr-domination} and monotone convergence give $0\leq f_u(\bx,r)\leq F_u(\bx,r)$ pointwise. Tonelli's theorem and the partition by $H_u^{(n)}$ give
\begin{align*}
    \sum_{\bx}\int_{\mathcal O}f_u(\bx,r)\,\rmd r
    &=\sum_{n=0}^{\infty}\sum_{\bx}\int_{\mathcal O}f_u^{(n)}(\bx,r)\,\rmd r\\
    &=\sum_{n=0}^{\infty}\pP(H_u^{(n)})=1.
\end{align*}
By the definition of $F_u$,
\begin{align*}
    \sum_{\bx}\int_{\mathcal O}F_u(\bx,r)\,\rmd r
    =\sum_{\bx}\pdata{u}{}{\bx}
      \prod_{c\in\intset{1}{\len}\times\intset{1}{\vocab}}
      \int_0^\infty\exp(-\zeta)\,\rmd\zeta=1.
\end{align*}
Consequently,
\begin{align*}
    \sum_{\bx}\int_{\mathcal O}\bigl(F_u(\bx,r)-f_u(\bx,r)\bigr)\,\rmd r=0.
\end{align*}
Its integrand is nonnegative, so $f_u(\bx,\cdot)=F_u(\bx,\cdot)$ for Lebesgue-almost every $r$, for each $\bx$. This proves \eqref{eq:erf-tr-theorem} for every deterministic $u<1$. At $u=1$ it is \eqref{eq:erf-tr-init}. Integrating the product density over the coordinate box $0<r^{\ell j}\leq \nu^{\ell j}$ gives
\begin{align*}
    \pdata{u}{}{\bx}\prod_{(\ell,j)\in\intset{1}{\len}\times\intset{1}{\vocab}}
      \int_0^{\nu^{\ell j}}\exp(-\zeta)\,\rmd\zeta
    =\pdata{u}{}{\bx}\prod_{(\ell,j)\in\intset{1}{\len}\times\intset{1}{\vocab}}
      \bigl(1-\exp(-\nu^{\ell j})\bigr),
\end{align*}
which gives the corresponding distribution function.
\end{proof}

\begin{corollary}
\label{cor:erf-tr-truncation}
Under the hypotheses of \Cref{thm:erf-tr-main}, for every deterministic $u\in[0,1)$ and integer $m\geq0$,
\begin{align}
    \sum_{\bx}\int_{\mathcal O}
      \left(F_u(\bx,r)-\sum_{n=0}^m f_u^{(n)}(\bx,r)\right)\,\rmd r
    =\pP(N_u\geq m+1)\leq\frac{M_u}{m+1}.
    \label{eq:erf-tr-truncation}
\end{align}
\end{corollary}
\begin{proof}
The integrand is nonnegative by \eqref{eq:erf-tr-partial-bound}. Subtract the masses of the events $H_u^{(0)},\ldots,H_u^{(m)}$ from the total mass one in \eqref{eq:erf-tr-sum-density}. This gives the equality. Except on a null set, $\{N_u\geq m+1\}=\{T^{\star}_{m+1}\in(u,1)\}$; apply \eqref{eq:erf-tr-event-tail}.
\end{proof}

\subsection{Last-jump distribution}

\begin{proposition}
\label{prop:erf-tr-last}
Assume the hypotheses of \Cref{thm:erf-tr-main}. Fix $u\in[0,1)$. On $\{N_u\geq1\}$ define
\begin{align*}
    R_u:=T^{\star}_{N_u},\qquad I_u:=\text{the first coordinate of }C_{N_u}.
\end{align*}
For every $\bx\in\msx$ and $\ell\in\intset{1}{\len}$, put $c=c_\ell(\bx)$. The joint measure of $(R_u,E_u^{-c})$, restricted to $\{N_u\geq1,X_u=\bx,I_u=\ell\}$, is
\begin{align}
    &\pP(N_u\geq1,X_u=\bx,I_u=\ell,R_u\in\rmd v,E_u^{-c}\in\rmd r^{-c})\notag\\
    &\quad=\pdata{u}{}{\bx}\lambda_v^\ell(\bx^\ell)
      \exp\!\left[-A_{u\tbar v}^\ell(\bx)-\sum_{a\neq c}r^a\right]
      \prod_{h\neq\ell}\indic_{\{\dotp{\bx^h}{r^h}>A_{u\tbar v}^h(\bx)\}}
      \,\rmd v\,\rmd r^{-c},
    \label{eq:erf-tr-last}
\end{align}
on $(u,1)\times(0,\infty)^{D-1}$. On this event, the remaining component satisfies $E_u^c=A_{u\tbar R_u}^\ell(\bx)$. The no-jump subdensity is \eqref{eq:erf-tr-f0}.
\end{proposition}
\begin{proof}
Almost surely there are finitely many successful jumps, no jump at $u$, and only the firing residual is zero at a jump. On $\{N_u\geq1,X_u=\bx\}$, apply \Cref{lem:erf-tr-reconstruction}(iii)--(iv) to the segment from its last jump $v=R_u$ to $u$. It shows that $E_u\in\mathcal R_u^{I_u}(\bx)$, that $R_u=\rho_{I_u}(u,\bx,E_u)$, and that $E_u^{c_{I_u}(\bx)}=A_{u\tbar R_u}^{I_u}(\bx)$. If there is no jump, the endpoint instead belongs to $\mathcal N_u(\bx)$ by the proof of \Cref{lem:erf-tr-initial}.

\Cref{thm:erf-tr-main} supplies the interior density $\pdata{u}{}{\bx}\gamma(r)$. On $\mathcal R_u^\ell(\bx)\cap\mathcal O$, replace the coordinate $r^c$ by $v$ through
\begin{align*}
    r^c=A_{u\tbar v}^\ell(\bx),\qquad r^a=r^a\quad(a\neq c).
\end{align*}
The determinant in these coordinates is $\lambda_v^\ell(\bx^\ell)$. The conditions for $\ell$ to be the least reconstruction index are precisely
\begin{align*}
    \dotp{\bx^h}{r^h}>A_{u\tbar v}^h(\bx)\quad(h\neq\ell).
\end{align*}
All active coordinates are already positive. On the set $\lambda_v^\ell(\bx^\ell)>0$, the change is injective, as in \Cref{lem:erf-tr-reconstruction}. \Cref{lem:erf-tr-calculus} shows that the remaining values of $r^c$ form a null set; the corresponding right-hand side of \eqref{eq:erf-tr-last} is zero where $\lambda_v^\ell(\bx^\ell)=0$. Substituting $r^c=A_{u\tbar v}^\ell(\bx)$ into the exponential density and multiplying by the determinant gives exactly \eqref{eq:erf-tr-last}.
\end{proof}

\begin{corollary}
\label{cor:erf-tr-L1}
Under the hypotheses of \Cref{thm:erf-tr-main}, suppose $\len=1$ and identify the state space with $\msv$. For every $u\in[0,1)$, $j\in\intset{1}{\vocab}$, and $r\in\mathcal O_{\representer{j}}$,
\begin{align*}
    f_u^{(0)}(\representer{j},r)
    =\pdata{u}{}{\representer{j}}\gamma(r)
      \indic_{\{r^{1j}>\int_u^1\lambda_\tau^1(\representer{j})\,\rmd\tau\}}.
\end{align*}
If $r\in\mathcal R_u(\representer{j})$, let $v\in(u,1)$ be the unique solution with $\lambda_v^1(\representer{j})>0$ of
\begin{align*}
    r^{1j}=\int_u^v\lambda_\tau^1(\representer{j})\,\rmd\tau,
\end{align*}
and set $w^{1j}=0$, $w^{1k}=r^{1k}+\int_u^v\rratemat{\tau}{\representer{j}}{\representer{k}}[1]\,\rmd\tau$ for $k\neq j$. Then, for every integer $n\geq0$,
\begin{align}
    f_u^{(n+1)}(\representer{j},r)
    =\frac{1}{\lambda_v^1(\representer{j})}
      \sum_{k\in\intset{1}{\vocab}\setminus\{j\}}
      \rratemat{v}{\representer{k}}{\representer{j}}[1]f_v^{(n)}(\representer{k},w).
    \label{eq:erf-tr-L1-recursion}
\end{align}
It is zero when $r\notin\mathcal R_u(\representer{j})$.
\end{corollary}
\begin{proof}
There is one passive residual, namely $c_1(\representer{j})=(1,j)$. Formula \eqref{eq:erf-tr-N} reduces to the single inequality in the no-jump formula. In \eqref{eq:erf-tr-K}, the only possible value of $\ell$ is $1$, the predecessor state is $\representer{k}$, and the reversed rate from this predecessor into $\representer{j}$ is $\rratemat{v}{\representer{k}}{\representer{j}}[1]$. Equations \eqref{eq:erf-tr-D-components} and \eqref{eq:erf-tr-reconstruction} give the stated values of $w$. This proves \eqref{eq:erf-tr-L1-recursion} and the zero convention directly from the general recursion.
\end{proof}